\pdfoutput=1

\documentclass[11pt,table]{article}

\usepackage[preprint]{acl}

\usepackage{times}
\usepackage{latexsym}

\usepackage[T1]{fontenc}

\usepackage[utf8]{inputenc}

\usepackage{microtype}

\usepackage{inconsolata}

\usepackage{graphicx}

\usepackage{graphicx}
\usepackage{caption,subcaption}
\usepackage{wrapfig}
\usepackage{enumitem}
\usepackage{url}
\newcommand{\tinycitep}[1]{\tiny\citep{#1}\normalsize}
\usepackage[section]{placeins} 

\usepackage{multirow}
\usepackage{adjustbox}
\usepackage{booktabs}
\usepackage{array}
\usepackage{amsmath}
\usepackage{enumitem}
\usepackage{tablefootnote}
\usepackage{xcolor}
\newcommand\todo[1]{\textcolor{red}{TODO: #1}}

\title{\textsc{Laser}: Stratified Selective Sampling for Instruction Tuning \\ with Dedicated Scoring Strategy}

\author{
 \textbf{Paramita Mirza\textsuperscript{1}$^*$},
 \textbf{Lucas Weber\textsuperscript{2}$^*$},
 \textbf{Fabian Küch\textsuperscript{2}}
\\
\\
 \textsuperscript{1}ScaDS.AI \& TU Dresden, Germany,\\
 \textsuperscript{2}Fraunhofer IIS, Germany
 \\
 \small{\texttt{paramita.mirza@tu-dresden.de
 \hspace{20pt}
 \{lucas.weber, fabian.kuech\}@iis.fraunhofer.de}}
}

\begin{document}
\maketitle

\renewcommand{\thefootnote}{}
\footnotetext{Accepted at \emph{Findings of the Association for Computational Linguistics: EMNLP 2025}.}
\renewcommand{\thefootnote}{\arabic{footnote}}

\def\thefootnote{*}\footnotetext{These authors contributed equally to this work.}\def\thefootnote{\arabic{footnote}}

\begin{abstract}
Recent work shows that post-training datasets for LLMs can be substantially downsampled without noticeably deteriorating performance. However, data selection often incurs high computational costs or is limited to narrow domains. In this paper, we demonstrate that data selection can be both---efficient and universal---by using a multi-step pipeline in which we efficiently bin data points into groups, estimate quality using specialized models, and score difficulty with a robust, lightweight method. Task-based categorization allows us to control the composition of our final data---crucial for finetuning multi-purpose models. To guarantee diversity, we improve upon previous work using embedding models and a clustering algorithm. This integrated strategy enables high-performance fine-tuning with minimal overhead.
\end{abstract}

\section{Introduction}
\label{sec:intro}

Large language models (LLMs) can perform a wide range of text-based tasks through chat interfaces. Their generalist abilities stem from an extensive post-training phase, during which they are optimized to generate useful responses to user queries \citep{sanh2022multitask,mishra2022cross,flant5,ouyang_training_2022}. 
The choice of training data in this phase has a major impact on model performance \citep[e.g.][]{zhou2023lima,xia2024less,liu2024what,chen2024alpagasus,ge-etal-2024-clustering}. 

Prior work identifies three key properties of effective instruction-tuning (IT) data \citep{zhou2023lima}:
\emph{i)} \textbf{difficulty}/\textbf{relevance}, reflecting how much a query contributes to learning \citep{li-etal-2024-quantity,liu2024selectit,liu2024what,zhao2024tree};
\emph{ii)} \textbf{quality}, the usefulness and accuracy of responses \citep{zhao_long_2024,chen2024alpagasus,liu2024what}; and
\emph{iii)} \textbf{diversity}, the scope of within-domain variability and cross-domain coverage of the data \citep{ge-etal-2024-clustering,lu_instag_2023}.
Although many approaches address one or more of these aspects, they often face limitations that make them difficult to apply in practice.

\begin{figure}[t!]
    \centering
    \includegraphics[width=1.0\linewidth]{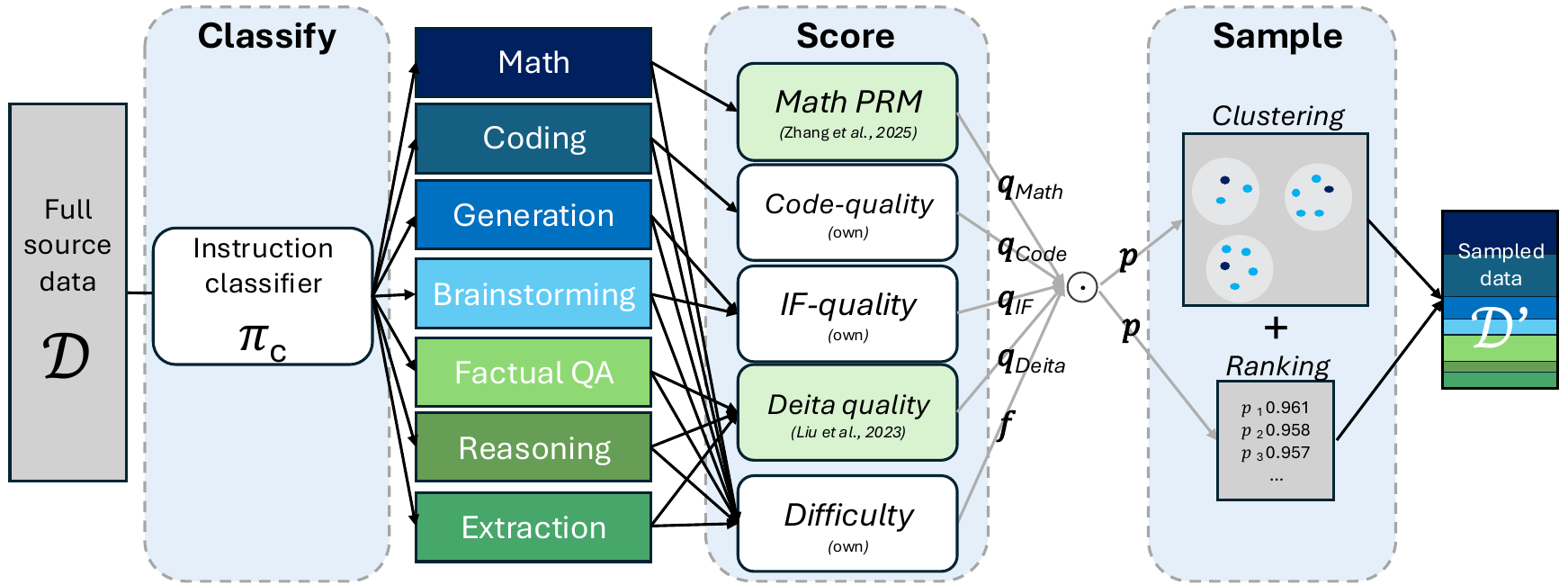}
    \caption{Overview of our three-stage pipeline, \textsc{Laser}. Instructions are first classified into seven types, then routed and scored for difficulty ($f$) and quality ($q$). Their combined scores ($p$) are ranked, and top examples are sampled both within embedding clusters and overall, while preserving fixed category proportions.}
    \label{fig:pipeline_illustration}
\end{figure}

Most existing methods cover only a subset of criteria (e.g., \emph{diversity} in \citealt{chen2024alpagasus}; \emph{quality} and \emph{diversity} in \citealt{ge-etal-2024-clustering}). 
To our knowledge, \textsc{Deita} \citep{liu2024what} is the only \textbf{complete} pipeline that integrates all three simultaneously. 
Beyond completeness, prior work can be further refined in terms of \textbf{generality} and \textbf{robustness}. 
Many heuristic-based methods cannot be applied to different tasks \citep{muennighoff2025s1} or domains \citep{zhou2023lima}. 
Broader approaches---that do not limit themselves explicitly to a single domain---are usually evaluated only on narrow domains or task-types (e.g., open-ended language generation), fine-tuned on a small set of models (typically from the Llama or Mistral family with 3B–8B parameters), or limited to a single source dataset. 
Outside these settings, they often prove brittle, failing to consistently outperform random sampling or simple heuristics such as a \emph{longest} baseline \citep{diddee-ippolito-2025-chasing,zhao_long_2024}.
Ultimately, many methods are \textbf{expensive} as they rely on large or closed-weight LLM judges \citep{chen2024alpagasus,lu_instag_2023}.

We address these limitations with \textsc{Laser} (Label-Aware Scoring and clustERing), a robust, complete, and efficient data selection pipeline.\footnote{Code, model access, and replication details will be released upon publication at \url{https://github.com/Fraunhofer-IIS/LASER}}

Our \textbf{key contributions} are the following:
\begin{enumerate}[label=\emph{(\roman*)},leftmargin=*,topsep=0pt]
    \item A novel, \emph{complete} and \emph{robust} data sampling strategy for instruction-tuning that integrates \emph{efficient} scoring of difficulty and quality via routing while maintaining both inter- and intra-class diversity through stratified sampling and clustering;
    \item A new and general approach to estimate the difficulty of instructions; 
    \item Task-specific quality scoring strategies, including custom-designed scorers, particularly for constrained generation and coding tasks focusing on responses' correctness; and 
    \item A large-scale evaluation spanning diverse model families and sizes, on a wide range of benchmarks covering various task types, with comparisons to strong baselines.
\end{enumerate}

\section{Related work}
\label{sec:related}

\paragraph{IT data selection.} 
Prior work on IT data selection generally falls into two categories based on how they assess sample difficulty, quality, and diversity: \emph{external scoring} and \emph{model-inherent criteria}. \textbf{External scoring} methods rely on: \emph{(i)} handcrafted features such as coherence, grammaticality, naturalness and understandability \citep{cao2024instruction}; \emph{(ii)} heuristics based on length or formatting \citep{zhao_long_2024,muennighoff2025s1}; and \emph{(iii)} scores derived from LLMs of varying scales (\emph{LLM-scorers}). Notable approaches leveraging LLM-scorers are: \textsc{AlpaGasus} \citep{chen2024alpagasus}, which prompts ChatGPT to score each data point; \textsc{InsTag} \citep{lu_instag_2023}, which uses ChatGPT to generate open-ended tags for deriving complexity/diversity; \textsc{Deita} \citep{liu2024what}, which fine-tunes LLaMA-1-13B on ChatGPT-annotated complexity/quality labels as \textsc{Deita}-scorers; and \textsc{CaR} \citep{ge-etal-2024-clustering}, which trains a 550M reward model to rank instruction-response pairs by quality.

On the other hand, \textbf{model-inherent criteria} form another group of approaches that 
rely on signals from the target model itself,
including: \textsc{LESS} \citep{xia2024less}, which 
estimates data influence using gradient information;
\textsc{SelectIT} \citep{liu2024selectit}, 
which measures uncertainty via
token probability, prompt variation and multiple models' assessments; \textsc{SHED} \citep{he2024shed}, which clusters data 
and estimates impact using Shapley values; \emph{instruction-following difficulty} \citep[\textsc{IFD};][]{li-etal-2024-quantity}, measuring discrepancies between the model’s intrinsic generation capability and its desired response; and the approach by \citet{li-etal-2024-one}, which 
evaluates sample utility based on how much it reduces loss when used as an in-context example.

Our approach embraces external and model-inherent perspectives simultaneously: we use domain-specific LLM-scorers to assess quality, and leverage various models' performances across benchmarks to estimate difficulty---that is, how likely an average model is to fail on a given instruction. For diversity, our approach closely follows \textsc{CaR} \citep{ge-etal-2024-clustering}, which clusters data points and samples one representative from each cluster.

\vspace{-5pt}
\paragraph{IT data categorisation.} 
Previous work has proposed various task types (e.g., \emph{open QA}, \emph{brainstorming}, \emph{creative writing}), to guide IT data collection from human annotators \citep{ouyang_training_2022, DatabricksBlog2023DollyV2}. \citet{grattafiori_llama_2024_reduced} finetuned Llama 3 8B for coarse-grained (e.g., \emph{mathematical reasoning}) and fine-grained (e.g., \emph{geometry and trigonometry}) topic classification to help filter low-quality samples, though they provide little detail on the methodology or intended downstream use. Other efforts impose tags or domain taxonomy on IT data to ensure diversity \citep{lu_instag_2023,muennighoff2025s1}.
\citet{dong-etal-2024-abilities} study how data composition across tasks affects model performance, but assume that ShareGPT only contains general alignment tasks, while our analysis shows that nearly 30\% of it actually consists of coding tasks. To our knowledge, no prior work leverages IT data categorisation to apply task-specific scoring strategies during selection.

\vspace{-5pt}
\paragraph{Utilities of IT data selection methods.}
Several papers criticized the effectiveness and cost-efficiency of existing IT data selection strategies.
\citet{zhao_long_2024} show that simply selecting the longest responses can outperform more complex methods while being significantly cheaper and easier to implement. 
Similarly, \citet{diddee-ippolito-2025-chasing} find that many sophisticated methods barely outperform random sampling under realistic conditions, and emphasize the cost-performance trade-off.
However, most of these comparisons are limited to a single-source sampling setup, whereas a more practical scenario involves selecting from a pool of IT data sources. Additionally, prior evaluations focus on LLaMA models with a few exceptions using Mistral \cite{liu-etal-2025-take}, leaving the generalizability of selection strategies across model families and sizes largely unexplored.

\section{Methods}
\label{sec:methods}

Our goal is to determine a subset of an arbitrarily large instruction source dataset that is effective when finetuning a given pre-trained base model.
We define an effective dataset as one which achieves 
high performance on a large and general set of evaluation tasks, while requiring relatively few model parameter updates.
Formally, let $\mathcal{D} = \{d_i\}_{i=1}^N$ with $d_i = (x_i, y_i)$ be the full source dataset of size $N$, where \(x_i\) represents an input sequence (i.e. an instruction) and \(y_i\) represents the corresponding output (i.e. the desirable model response).
We wish to select a subset  $\mathcal{D}' \subset \mathcal{D}$ of size $m$ (with $m \ll N$) that is most effective for instruction tuning.

Previous research has shown that \emph{i)} \textit{instruction difficulty} \citep[see \textup{e.g.}][]{li-etal-2024-quantity,liu2024selectit,liu2024what,zhao2024tree} \emph{ii)} \textit{response quality} \citep[see \textup{e.g.}][]{zhao_long_2024,chen2024alpagasus,liu2024what} and \emph{iii)} \textit{diversity}/\textit{composition} \citep[see \textup{e.g.}][]{ge-etal-2024-clustering,lu_instag_2023} are crucial for effective data selection for LLM finetuning. 
We address these three aspects in the three-step \textsc{Laser} pipeline as illustrated in Figure~\ref{fig:pipeline_illustration}:

\begin{enumerate}[leftmargin=*,topsep=0pt,noitemsep]
    \item \textbf{Classification}. We train a lightweight classifier $\pi_{c}$ to categorise all inputs $x_i$ in $\mathcal{D}$ into one out of seven categories, which we denote as $l \in \mathcal{L}$. 
    \item \textbf{Scoring}. Each input–output pair $(x_i, y_i)$ is scored using a general-purpose difficulty scorer (yielding $f_i$) and a category-specific quality scorer (yielding $q_i$); the results are combined into an overall preference score $p_i$.
    \item \textbf{Clustering + Ranking}. Ultimately, we select the samples with the highest $p_i$ while using a clustering approach to maintain diversity and minimise redundancies in $\mathcal{D}'$.
\end{enumerate}
We detail each step in the remainder of this section.





\subsection{Instruction Classification}
\label{subsec:setfit}
Inspired by the use-case categories defined in \citet{ouyang_training_2022}, we establish the following task categories $\mathcal{L}$ for samples in instruction tuning datasets: 
\emph{i)} \textbf{Math}, from simple calculation to problems requiring multi-step reasoning; \emph{ii)} \textbf{Coding}, code generation tasks or programming-related question answering; \emph{iii)} \textbf{Generation}, textual generation tasks including roleplaying, summarizing and rewriting passages; \emph{iv)} \textbf{Reasoning}, questions requiring deductive/logical reasoning; \emph{v)} \textbf{Brainstorming}, information-seeking and recommendation questions that require inductive reasoning, including classification tasks; \emph{vi)} \textbf{Factual QA}, factual questions with simple facts as answers; and \emph{vii)} \textbf{Extraction}, tasks requiring structured/answer extraction from textual contexts. 
We let two human annotators classify 80 samples from the popular MT-Bench dataset \citep{zheng2023judging} and achieve high inter-annotator agreement (Cohen’s Kappa = 0.8635), demonstrating the discriminability of our categories.
We compare two different approaches to build a classification model: \emph{i) LLM annotator} and \emph{ii) SetFit  classifier}. 


With the LLM-annotator approach \citep{flant5}, we prompt instruction categorization by listing categories with brief explanations, followed by \emph{``What is the category of the following task?''} (see Figure~\ref{fig:zero-shot-prompt}, Appendix~\ref{app:instr_category}).
Meanwhile, SetFit \citep{tunstall_efficient_2022} is a few-shot learning method that tunes Sentence Transformers \citep{reimers-gurevych-2019-sentence} on labelled input pairs in a contrastive, Siamese manner. 
We manually identify approximately 250 samples 
strongly associated with each category (see Table \ref{tab:setfit-training-data}, Appendix \ref{app:instr_category})
to train the SetFit classifier; 
hyperparameters are detailed in Appendix \ref{app:instr_category}.

For evaluation, we use the manually annotated MT-Bench dataset as the test set, assessing accuracy, macro F1-score, and Cohen's Kappa agreement with human judgment (see Table \ref{tab:setfit-evaluation}). While zero-shot prompting with GPT-4o performs best, we choose the SetFit classifier
for our pipeline due to its comparable performance and higher efficiency than larger LLMs like GPT-4o and Falcon3-10B-Instruct \citep{Falcon3}.

\begin{table}[!t] 
\centering 
\scriptsize
\begin{adjustbox}{width=0.48\textwidth}
\begin{tabular}{@{}llccc@{}}
\toprule
\textbf{Approach} & \textbf{LLM annotator/embedding model} & \textbf{Acc} & \textbf{F1} & \textbf{Cohen's} \\
 &  &  &  & \textbf{Kappa} \\
\midrule
Zero-shot & GPT-4o & \textbf{0.88} & \textbf{0.86} & \textbf{0.85} \\
 prompting & tiiuae/Falcon3-10B-Instruct & 0.85 & 0.84 & 0.82 \\
  & meta-llama/Llama-3.1-8B-Instruct & 0.76 & 0.65 & 0.71 \\
\arrayrulecolor{black!30}\midrule
SetFit & NovaSearch/stella\_en\_400M\_v5 $\dagger$ & 0.85 & 0.81 & 0.82 \\
 classifier & Lajavaness/bilingual-embedding-large & 0.82 & 0.78 & 0.79 \\
 & NovaSearch/stella\_en\_1.5B\_v5 & 0.66 & 0.60 & 0.60 \\
\arrayrulecolor{black}\bottomrule
\end{tabular}
\end{adjustbox}
\caption{Evaluation results on instruction categorization. $\dagger$ denotes the chosen approach and model for our data selection pipeline.}
\label{tab:setfit-evaluation}
\end{table}

\subsection{Scoring}
\label{subsec:scoring}

\subsubsection{Difficulty scorer}
\label{subsubsec:difficulty_scoring}

Previous research suggests that \emph{difficulty} (sometimes referred to as \emph{complexity}) matters for data selection \citep[see e.g.][]{liu2024what,cao2024instruction,zhao2024tree,muennighoff2025s1}, with more challenging data generally resulting in better model performance.
However, existing difficulty metrics either lack generality across domains \citep[e.g. length of reasoning trace in response in][or the \textsc{Deita} \textit{complexity scorer}\footnote{\href{https://huggingface.co/hkust-nlp/deita-complexity-scorer}{\nolinkurl{hkust-nlp/deita-complexity-scorer}}} as shown in Figure~\ref{fig:scorer_eval_complexity}]{muennighoff2025s1} or are strongly influenced by spurious features 
\citep[e.g. the widely used \textsc{Deita}-complexity is strongly biased towards long sequences; see Figure~\ref{fig:corr_length_x_complexity};][]{liu2024what}.

Our goal is to train a general and robust difficulty scorer that predicts how likely it is for an average model to solve a data point incorrectly, independent of its category $l$.
We denote this difficulty score as $f$.

To source the training set $\mathcal{D}_{\text{diff}}$ for such a scorer, we collect 20k instruction-response pairs, approximately equally distributed across categories $l_i\in\mathcal{L}$. We list the data sources in Table~\ref{tab:diff_scorer_training_datasets} and show the category proportions in Figure~\ref{fig:diff_scorer_setfit_proportions}.
Then, we evaluate every item with a heterogeneous pool of 18 instruction‑tuned LLMs (see Table~\ref{tab:diff_scorer_evaluated_models}).
We continue by applying multiple preprocessing steps to the model scores:
First, we normalise the item scores to the interval $[0,1]$ and remove items with a score of 0 across all models, as they are likely to contain noise or annotation errors and will not yield any valuable learning signal.
We then go on to convert the absolute scores into \emph{relative} deviations with respect to the model’s mean performance on the corresponding dataset, by subtracting the average from the absolute score on each item (illustrated in Figure~\ref{fig:relative_performance_illustration}).
This step has the effect of weighing wrong responses of strong models more than wrong responses of weaker models (and vice versa). Further, it mitigates potential skews in the model performance distribution (e.g. if many models perform weakly on a specific dataset).
Ultimately, we obtain difficulty targets $f$ by averaging over the model pool for every item.

We fine‑tune a Qwen‑3‑8B backbone \citep{qwen3} with a single‑layer regression head to minimise the mean‑squared error on this dataset (training details can be found in Appendix~\ref{app:diff_scorer:training_details}).

\vspace{-5pt}
\paragraph{Difficulty scorer evaluation.}
We evaluate the effectiveness and the validity of the difficulty scoring.
For effectiveness, we sample 25k data points from $\mathcal{D}$ (as it is described in Section~\ref{sec:exp_setup}) either at random or following the difficulty scores. 
We then finetune Mistral-7B-v0.3 \citep{jiang2023mistral7b} on the resulting datasets.
Appendix~\ref{app:diff_scorer:effectiveness} shows how the difficulty scorer helps improve performance on benchmarks independently of their domain.
For validity (i.e. whether we are really measuring some notion of difficulty), we score the CodeForces section of DeepMind's code-contests dataset \citep{li2022competition} and find high correlations with its human-annotated difficulty scores (details in Appendix~\ref{app:diff_scorer:validity}).

\subsubsection{Quality scorer}
\label{subsubsec:quality_scorer}

Prior work shows that sample selection strategies based on response length or quality, as judged by external models, lead to better instruction tuning datasets \citep[see e.g.][]{zhao_long_2024,chen2024alpagasus,liu2024what}. However, these strategies underestimate the diverse problem types within instruction tuning data, which may require different evaluation criteria. For example, the quality of responses to math and coding problems is heavily dependent on solution correctness, while constrained generation requires evaluation of adhered constraints. In this work, we designate a dedicated quality scorer for each category $l\in\mathcal{L}$ defined in Section \ref{subsec:setfit}. 

\vspace{-5pt}
\paragraph{\textsc{Deita} quality scorer.} 
For \emph{Reasoning}, \emph{Factual QA} and \emph{Extraction} samples, we employ the \emph{\textsc{Deita} quality scorer}\footnote{\href{https://huggingface.co/hkust-nlp/deita-quality-scorer}{\nolinkurl{hkust-nlp/deita-quality-scorer}}} \citep{liu2024what}, a finetuned LLaMA-13B for quality assessment, which yields the $q_{\mathrm{deita}}$ score.

\vspace{-5pt}
\paragraph{Process reward model.} 
For \emph{Math} data, we assess the soundness of mathematical reasoning using a specialised process reward model \citep[Qwen2.5-Math-PRM-7B; ][]{zhang2025lessons}. 
 Process reward models \citep
[PRMs;][]
{lightman2023let,uesato2022solving} are trained to verify steps in reasoning traces as they are common in mathematical reasoning. 
We find double linebreaks and---if no double linebreaks present---single linebreaks as a delimiter to be a good heuristic to separate reasoning steps.
As a reasoning trace breaks with a single erroneous step, we aggregate scores by taking the minimal score out of all steps within each trace, as the $q_{\mathrm{math}}$ score.

\vspace{-5pt}
\paragraph{Code quality scorer.}
We design a quality scoring framework for \emph{Coding} samples by drawing inspiration from \citet{wadhwa_core_2024}. For each data point $(x_i, y_i)$, we leverage code-oriented LLMs to: \emph{(i)} assess the functional correctness of the code snippet in $y_i$ with respect to the problem $x_i$, and \emph{(ii)} produce a revised version that improves or fixes the original code (see Figure~\ref{fig:zero-shot-prompt-code-score}, Appendix~\ref{app:quality_scorer:code}).
The resulting score, $q_{\mathrm{code}}$, is based on the \emph{normalized Levenshtein similarity} between lines of the original ($lo_0, ..., lo_n$) and revised ($lr_0, ..., lr_m$) code:
$nls = (\max(n, m) - \mathrm{lev}(lo, lr)) / \max(n, m)$,
where $\mathrm{lev}(lo, lr)$ is the line-level Levenshtein distance. If the original code is functionally correct, we set $q_{\mathrm{code}} = nls$; otherwise, 
$q_{\mathrm{code}} = nls / 2$. If no code snippet is present, we assign $q_{\mathrm{code}} = 0.5$ if $y_i$ is judged correct, and 0.0 otherwise.

To evaluate this scoring method, we use a 1K-sample test set from LiveCodeBench \citep{jain_livecodebench_2024}, containing coding problems and LLM-generated responses.\footnote{\href{https://huggingface.co/spaces/livecodebench/code_generation_samples}{\nolinkurl{livecodebench/code_generation_samples}}} Using Qwen/Qwen2.5-Coder-14B-Instruct as the reviewing model, our framework achieves 70\% accuracy and a 0.412 Pearson correlation with binary correctness labels (see Appendix~\ref{app:quality_scorer:code} for details). 

\vspace{-5pt}
\paragraph{Instruction-following scorer.} 
For \emph{Generation} and \emph{Brainstorming} data, we implement an if-quality scorer.
Influenced by the IFEval benchmark \citep{zhou2023instructionfollowingevaluationlargelanguage} which defines ``verifiable constraints'' such as length (\emph{``400 or more words''}) and keyword (\emph{``without using the word sleep''}) constraints, we design a response quality scorer based on the fraction of expressed constraints ($\mathcal{C}_{exp}$) adhered by the response ($\mathcal{C}_{true}$). 
First, we use an LLM annotator to identify $\mathcal{C}_{exp}$, which comprises (span, constraint type) pairs $\{(s_i,c_i)\}^{n_{exp}}_{i=1}$, with $s_i$ represents the textual span found and $c_i$ is the corresponding constraint label.
For example, given the prompt \emph{``Write a funny blog post with 400 or more words about the benefits of sleeping in a hammock, without using the word sleep.''}, $\mathcal{C}_{exp}=$ \{(\emph{400 or more words}, length), (\emph{without using the word ``sleep''}, keyword avoided), (\emph{funny blog post}, writing type)\}. 
A list of considered constraint types is shown in Figure \ref{fig:few-shot-prompt-constraint-identification} (Appendix \ref{app:quality_scorer:if}).
Next, $\mathcal{C}_{exp}$ is passed to a constraint checker module, which performs two steps:
\begin{enumerate}[leftmargin=*,topsep=0pt,noitemsep]
\item Heuristic verification: We verify length, letter case, punctuation and keyword constraints, by adapting the IFEval verification script.
\item LLM-judge verification: We ask an LLM judge to assess constraints that cannot be verified heuristically (e.g., \emph{``Does the following text follow the [writing type] constraint of [funny blog post]?''}).
\end{enumerate}
This yields $\mathcal{C}_{true}=\{(s_j,c_j)\}^{n_{true}}_{j=1}$, with which we compute the quality score as $q_{\mathrm{if}}=n_{true} * (n_{true} / n_{exp})$, giving more incentives to responses adhering to more constraints. If $\mathcal{C}_{exp}$ is empty, we ask an LLM judge to evaluate whether the response \emph{i)} addresses the user's intent, while \emph{ii)} respecting any constraints expressed in the prompt, and to provide a final $score$ (1--10), which we use to compute the score as $q_{\mathrm{if}}=score/10$.

Our analysis with the IFEval benchmark dataset containing sample responses from ten models as our test bed (see Table~\ref{tab:models-ifeval}, Appendix~\ref{app:quality_scorer:if}), shows that Qwen3-14B \citep{qwen3} outperformed other medium-sized Instruct-LLMs as both LLM annotator and judge (see Table~\ref{tab:if-scoring-evaluation}, Appendix~\ref{app:quality_scorer:if}). It achieved a macro F1-score of 0.86 for identifying expressed constraints, a Pearson correlation coefficient of 0.523 at the instance-level, and 0.995 at the model-level, where it effectively replicated the IFEval model ranking.

\vspace{-5pt}
\paragraph{Quality scorer evaluation.}
Similar to the difficulty scorer, we demonstrate the effectiveness of our task-specific quality scorers for ranking-based data selection, particularly when compared to the general-purpose \textsc{Deita} quality scorer and random sampling. 
The corresponding results are presented in Appendix~\ref{app:quality_scorer:effectiveness}.

\begin{table*}[ht]
\begin{minipage}[b]{0.40\linewidth} 
\centering
\adjustbox{max width=1.0\textwidth}{%
    \begin{tabular}{@{}l@{}rr@{}}
        \toprule
        \textbf{Dataset} & \textbf{\#samples} \\
         & (\emph{\#turns}) \\
        \midrule
        HuggingFaceH4/ifeval-like-data   & 5K \\ 
        vicgalle/alpaca-gpt4 \tinycitep{taori2023stanford} & 52K \\
        nvidia/OpenMathInstruct-2 {\small\emph{(w/o augmented problems)}} \tinycitep{toshniwal2024openmath2} & 52K \\
        ai2-adapt-dev/flan\_v2\_converted \tinycitep{longpre2023flan} & 90K \\
        openbmb/UltraInteract\_sft (Coding) \tinycitep{yuan2024advancing} & 115K \\
        WizardLMTeam/WizardLM\_evol\_instruct\_V2\_196K \tinycitep{xu2024wizardlm} & 143K \\
        theblackcat102/sharegpt-english & 50K (392K) \\
        microsoft/orca-agentinstruct-1M-v1 {\small\emph{(random subset)}} \tinycitep{mitra_agentinstruct_2024} & 200K (903K) \\
        \arrayrulecolor{black!30}\midrule
        \emph{all} & 707K (1.75M) \\ 
        \arrayrulecolor{black}\bottomrule
    \end{tabular}}
    \caption{Source dataset $\mathcal{D}$}
    \label{tab:datasets}
\end{minipage}\hfill
\begin{minipage}[b]{0.58\linewidth} 
\centering
\raisebox{0pt}{\includegraphics[width=1.0\linewidth]{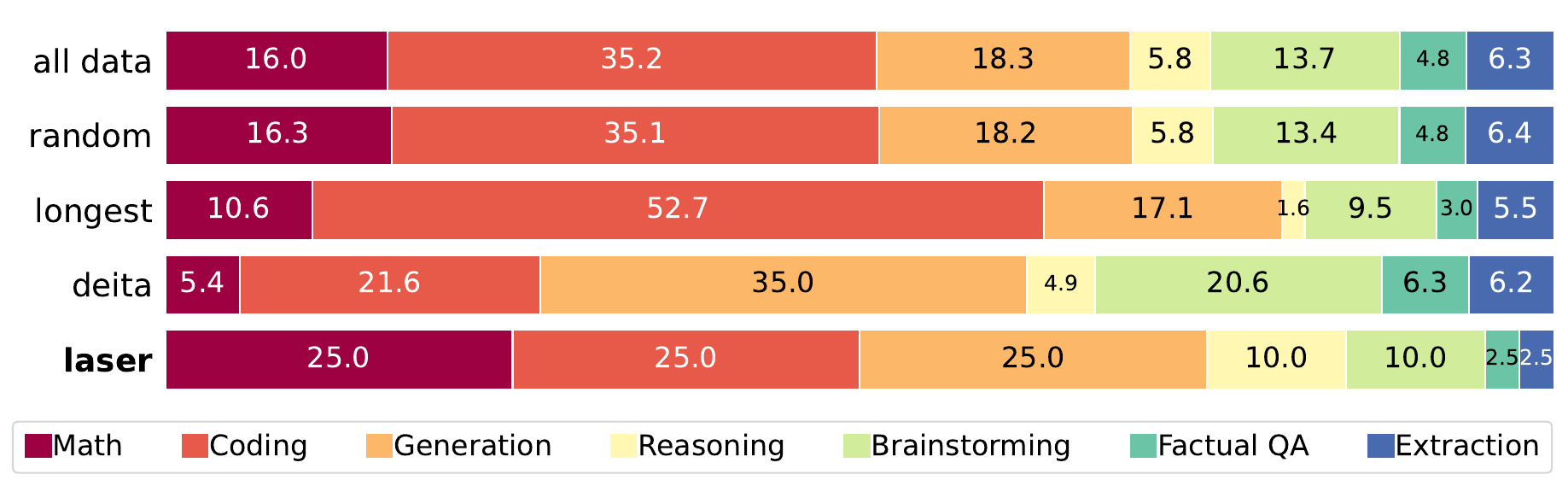}}
\captionof{figure}{Category distribution in IT datasets with varied sampling.}
\label{fig:sampling_categories}
\end{minipage}
\end{table*}

\vspace{-5pt}
\subsection{Sampling}
\label{subsec:Sampling} 

\paragraph{Overall preference scores.}
For each $(x_i, y_i) \in \mathcal{D}$, we compute the preference score $p_i=f_i \cdot q_i$, where $f_i$ and $q_i$ are the difficulty and quality scores, respectively. $f_i$ and $q_i$ are normalized per scorer as they have differing ranges, using min-max scaling, with the 1st and 99th percentiles as the minimum and maximum values across all samples in $\mathcal{D}$. 
For multi-turn conversations, where each data point $d_i$ consists of a sequence of turns $\{(x_0, y_0),..., (x_T, y_T)\}$, these scores are averaged across all turns to yield the overall conversation-level scores $f_i$ and $q_i$.


\vspace{-5pt}
\paragraph{Clustering + Ranking.}
Greedily choosing the highest-scoring samples often leads to redundancy in some domains and under-representation in others. 
Thus, maintaining diversity in the final dataset $\mathcal{D}'$ is essential.
Since diversity is a property of the dataset as a whole---not of individual samples---selection should consider the dataset globally \citep[e.g.,][]{ge-etal-2024-clustering}, rather than relying on iterative, sample-by-sample strategies \citep[e.g.,][]{liu2024what,bukharin-etal-2024-data}. 
Diversity can be promoted \emph{top-down} by balancing category proportions \citep[see e.g.][]{grattafiori_llama_2024_reduced,dong-etal-2024-abilities}, or bottom-up by ensuring sufficient semantic dissimilarity among selected samples \citep[e.g.,][]{liu2024what,ge-etal-2024-clustering,lu_instag_2023}.

As they are complementary, we propose a combination of both approaches: First, we determine the number of samples per category $l \in \mathcal{L}$ via a fixed quota, denoted $m_l$, ensuring a balanced proportion of \emph{Math}, \emph{Coding}, \emph{Generation} and so on.
Next, we embed all candidate samples within each category using a state-of-the-art sentence encoder \citep{reimers-gurevych-2019-sentence,zhang2025jasperstelladistillationsota}, and
cluster them into \(J\) groups using k-means \citep{lloyd1982least}, 
with $J=m_l$.
Let $\mathcal{K} = \{K_1, K_2, \dots, K_J\}$ denote the resulting clusters.
From each cluster $K$ in $\mathcal{K}$, we select the sample with the highest preference score $p_{\max}(K) = \max_{i \in K}p_i$. To improve robustness to clusters with very low $p_{\max}(K)$ values, we discard clusters whose best sample falls below a predetermined threshold, which is set to be the $\gamma^{\mathrm{th}}$-percentile\footnote{$\gamma$ is a hyperparameter and set to 80.} of $\{p_i\}_{i=1}^{N_l}$ where 
$N_l$ is the number of samples within the category $l$.
To reach the target of $m_l$ samples per category, we select the highest-scoring samples from the remaining candidates within that category.

Given the large value of $m_l$ and the high-dimensional nature of the input embeddings, we employ \emph{MiniBatchKMeans} from \emph{scikit-learn} with the \texttt{k-means++} initialization method and a batch size of 2048.

\vspace{-5pt}
\paragraph{Sampling evaluation.}
We evaluate the effectiveness of our stratified sampling and clustering as we did for the difficulty and quality scorer evaluation in Sections~\ref{subsubsec:difficulty_scoring} and \ref{subsubsec:quality_scorer}. In addition, we demonstrate the efficiency and robustness of \textsc{Laser}'s sampling technique, particularly in comparison with \textsc{Deita} \citep{liu2024what} and \textsc{CaR} \citep{ge-etal-2024-clustering}.
The respective results are reported in Appendix~\ref{app:sampling:effectiveness}.

\section{Experiments}
\label{sec:experiments}

\subsection{Experimental Setup}
\label{sec:exp_setup}

\paragraph{Datasets.}
We use the aggregation of \emph{all} listed datasets in Table \ref{tab:datasets} as the source dataset $\mathcal{D}$ in all experiments, if not specified otherwise.

\vspace{-5pt}
\paragraph{Models.}
While in most experiments we finetune Mistral-7B-v0.3, we also demonstrate generalisation across models of varying sizes and families:
\emph{i)} tiiuae/Falcon3-\textbf{10B}-Base, \emph{ii)} meta-llama/Llama-3.1-\textbf{8B}, \emph{iii)} mistralai/Mistral-\textbf{7B}-v0.3, \emph{iv)} Qwen/Qwen2.5-\textbf{3B} and \emph{v)} HuggingFaceTB/SmolLM2-\textbf{1.7B}.

\begin{figure*}[th!]
    \centering
    \begin{subfigure}[b]{0.87\textwidth}
        \hspace{6pt}
        \includegraphics[width=\textwidth]{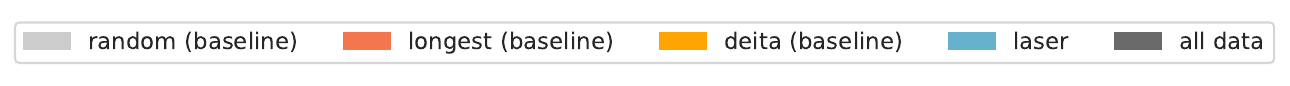}
        \label{fig:legend_strategies}
    \end{subfigure}\\[-4ex] 
    \begin{subfigure}[b]{0.56\textwidth}
        \centering
        \includegraphics[width=\textwidth]{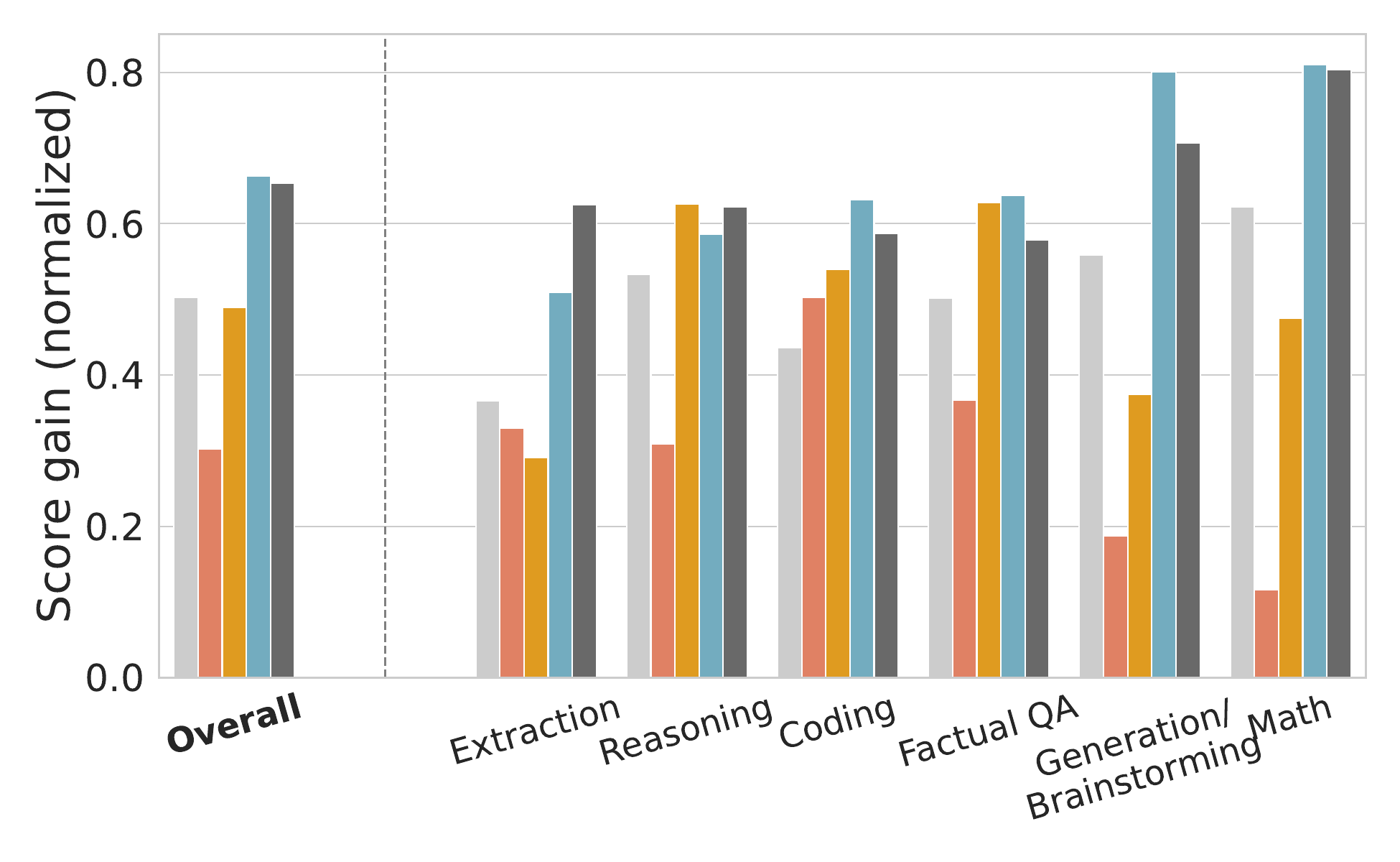}
        \vspace{-2em}
        \caption{Score gains per benchmark category}
        \label{fig:eval_aggr}
    \end{subfigure}
    \begin{subfigure}[b]{0.4\textwidth}
        \centering
        \raisebox{8pt}{\includegraphics[width=\textwidth]{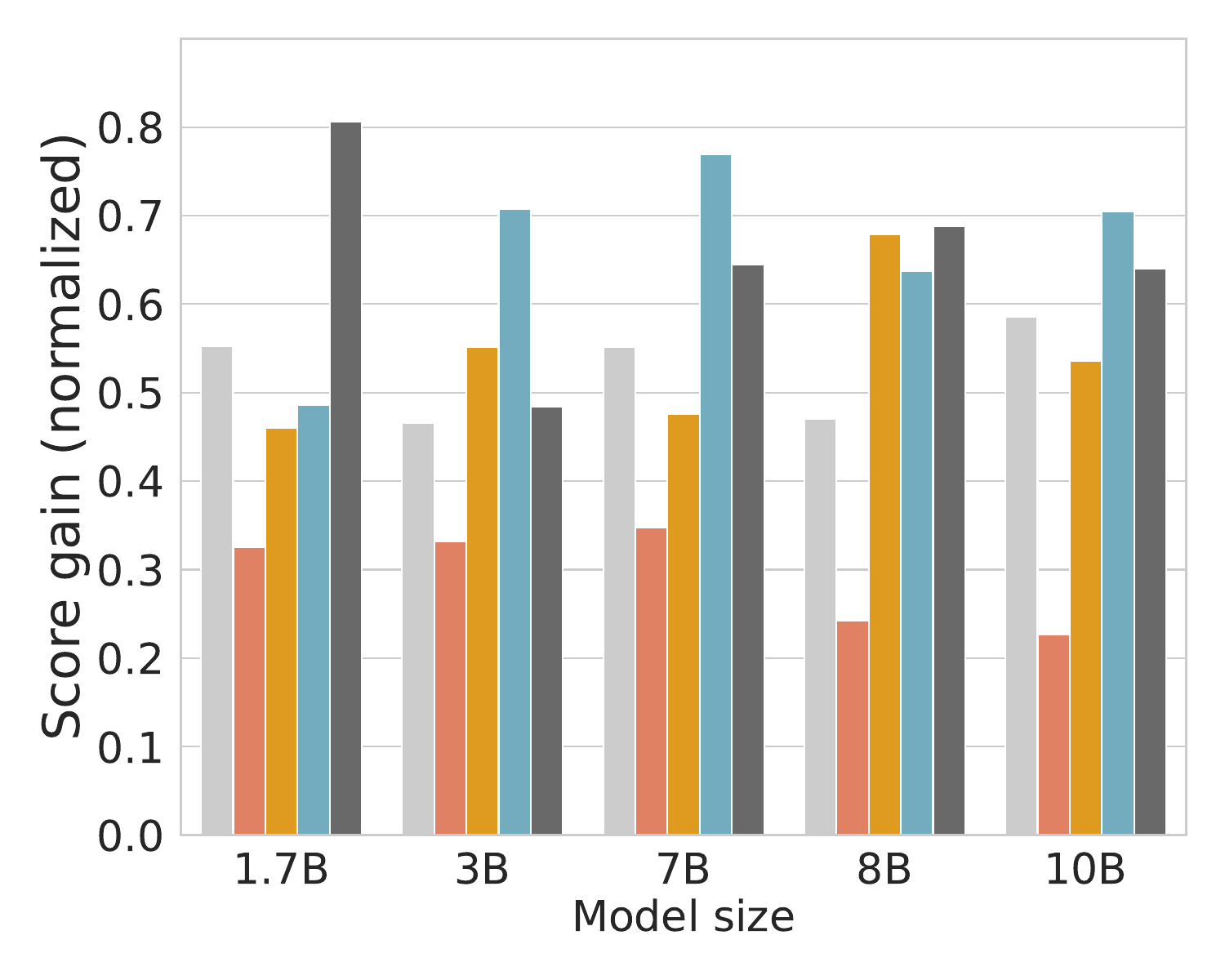}}
        \vspace{-1.7em}
        \caption{Score gains per model size}
        \label{fig:eval_by_size}
    \end{subfigure}
    
    \caption{Performance gains over the base model for different sampling strategies with 100k samples. We aggregate the results (a) across all benchmarks separated by models; (b) across all models separated by benchmark categories.}
    \label{fig:results_main}
\end{figure*}

\vspace{-5pt}
\paragraph{Finetuning.} We fine-tune each base model using the \emph{SFTTrainer} from the Transformer Reinforcement Learning (TRL) library\footnote{\url{https://huggingface.co/docs/trl/sft_trainer}} \citep{vonwerra2022trl}. 
Model-specific hyperparameters (e.g., learning rate) are selected for each model family and detailed in Appendix~\ref{app:finetuning}. 
We also employ NEFTune \citep{jain2024neftune}, a technique that improves the performance of chat models by injecting noise into embedding vectors during training. 

\vspace{-5pt}
\paragraph{Baselines.} We compare against the following alternative methods for constructing $D' \subset D$: 
\begin{enumerate}[label=\emph{\roman*)},leftmargin=*,topsep=0pt,noitemsep]
\item \emph{random}---sampling uniformly at random;
\item \emph{longest}---including samples having the longest responses;\footnote{In the case of multi-turn samples, we only consider the last turn's responses.}
\item \emph{deita} \citep{liu2024what}---ranking data points according to complexity and quality scores ($o_{deita}$ and $q_{deita}$, resp.), and iteratively builds $D'$ by adding \emph{dissimilar} samples based on embedding distances.
\end{enumerate}

\begin{table}[!t]
\centering
\small
\begin{adjustbox}{width=0.49\textwidth}
\begin{tabular}{@{}llll@{}}
\toprule
\textbf{Category ($\mathcal{L}$)} & \textbf{Benchmark} & \textbf{Evaluation details} & \texttt{lm\_eval} \\
\midrule
Math & GSM8k$_{\textrm{cot}}$ \tinycitep{cobbe2021gsm8k} & 8-shot; chain-of-thought & \texttt{gsm8k\_cot\_llama} \\
& GSM8k$_{\textrm{cot-0-shot}}$ \tinycitep{cobbe2021gsm8k} & 0-shot; chain-of-thought & \texttt{gsm8k\_cot\_zeroshot} \\
& Math \tinycitep{hendrycksmath2021}  & 4-shot & \texttt{leaderboard\_math\_hard}\\
\midrule
Coding & HumanEval \tinycitep{chen2021codex} & 0-shot; pass@1 & \texttt{humaneval\_instruct}\\
& MBPP \tinycitep{austin2021program} & 3-shot; pass@1 & \texttt{mbpp\_instruct}\\
\midrule
Generation/ & AlpacaEval \tinycitep{dubois2024lengthcontrolled} & length-controlled win-rate & - \\
Brainstorming & IFeval \tinycitep{zhou2023instructionfollowingevaluationlargelanguage} & 0-shot; prompt-level strict-acc & \texttt{leaderboard\_ifeval}\\
\midrule
Reasoning & ARC-C \tinycitep{allenai_arc} & 0-shot; multiple-choice & \texttt{arc\_challenge} \\
& BBH \tinycitep{suzgun-etal-2023-challenging} & 3-shot; multiple-choice & \texttt{leaderboard\_bbh}\\
& GPQA \tinycitep{rein2024gpqa} & 0-shot & \texttt{leaderboard\_gpqa} \\
& Hellaswag \tinycitep{zellers2019hellaswag} & 0-shot; multiple-choice & \texttt{hellaswag} \\
& MuSR \tinycitep{sprague2024musr} & 0-shot; multiple-choice & \texttt{leaderboard\_musr} \\
& Winogrande \tinycitep{ai2_winogrande} & 0-shot; multiple-choice & \texttt{winogrande} \\
\midrule
Factual QA & AGIEval \tinycitep{zhong-etal-2024-agieval} & 0-shot; multiple-choice & \texttt{agieval\_nous} \\
& MMLU \tinycitep{hendryckstest2021} & 0-shot; multiple-choice & \texttt{mmlu} \\
& TruthfulQA \tinycitep{lin-etal-2022-truthfulqa} & 0-shot; multiple-choice & \texttt{truthfulqa\_mc2} \\
\midrule
Extraction & OpenBookQA \tinycitep{mihaylov-etal-2018-suit} & 0-shot; multiple-choice & \texttt{openbookqa} \\
\arrayrulecolor{black}\bottomrule
\end{tabular}
\end{adjustbox}
\caption{Evaluation benchmarks with associated category and evaluation details.}
\label{tab:benchmark-details}
\end{table}

\vspace{-5pt}
\paragraph{Evaluation.} We evaluate the instruction-tuned models on a suite of benchmarks listed in Table~\ref{tab:benchmark-details}, leveraging popular evaluation frameworks \emph{LM-evaluation harness} \citep{eval-harness} and \emph{AlpacaEval}\footnote{We use \texttt{openai/gpt-oss-120b} as the LLM-judge.} \citep{dubois2024lengthcontrolled}.
We report \emph{normalized score gain} as the main evaluation metric, computed as the performance improvement over base models, scaled with min-max normalization.




\subsection{Results}
\label{sec:exp_results}

\subsubsection{Main results}
\label{subsubsec:main_results}

For each selection strategy, we construct a subset $\mathcal{D}'={d}_{i=1}^{m}\subset\mathcal{D}$ of size $m=100,000$, and subsequently fine‑tune each of the five base models on $\mathcal{D}'$.
Figure~\ref{fig:sampling_categories} shows the resulting category distributions across sampling strategies, including the baselines.
Figure~\ref{fig:eval_aggr} reports normalized score gains over the untuned base models for all evaluation benchmarks and models.
\textsc{Laser} achieves the highest overall performance, surpassing all baselines and even the models trained on the entire source set ($|\mathcal{D}|>707\text{k}$).
We examine the robustness by breaking the aggregated results down by model size (Figure~\ref{fig:eval_by_size}).
\textsc{Laser} yields the most consistent gains of all tested conditions for all five bases and outperforms the random sampling in all but the smallest model, confirming that the proposed sampling approach transfers well across models.

\begin{figure}[h!]
    \centering
    \includegraphics[width=\linewidth]{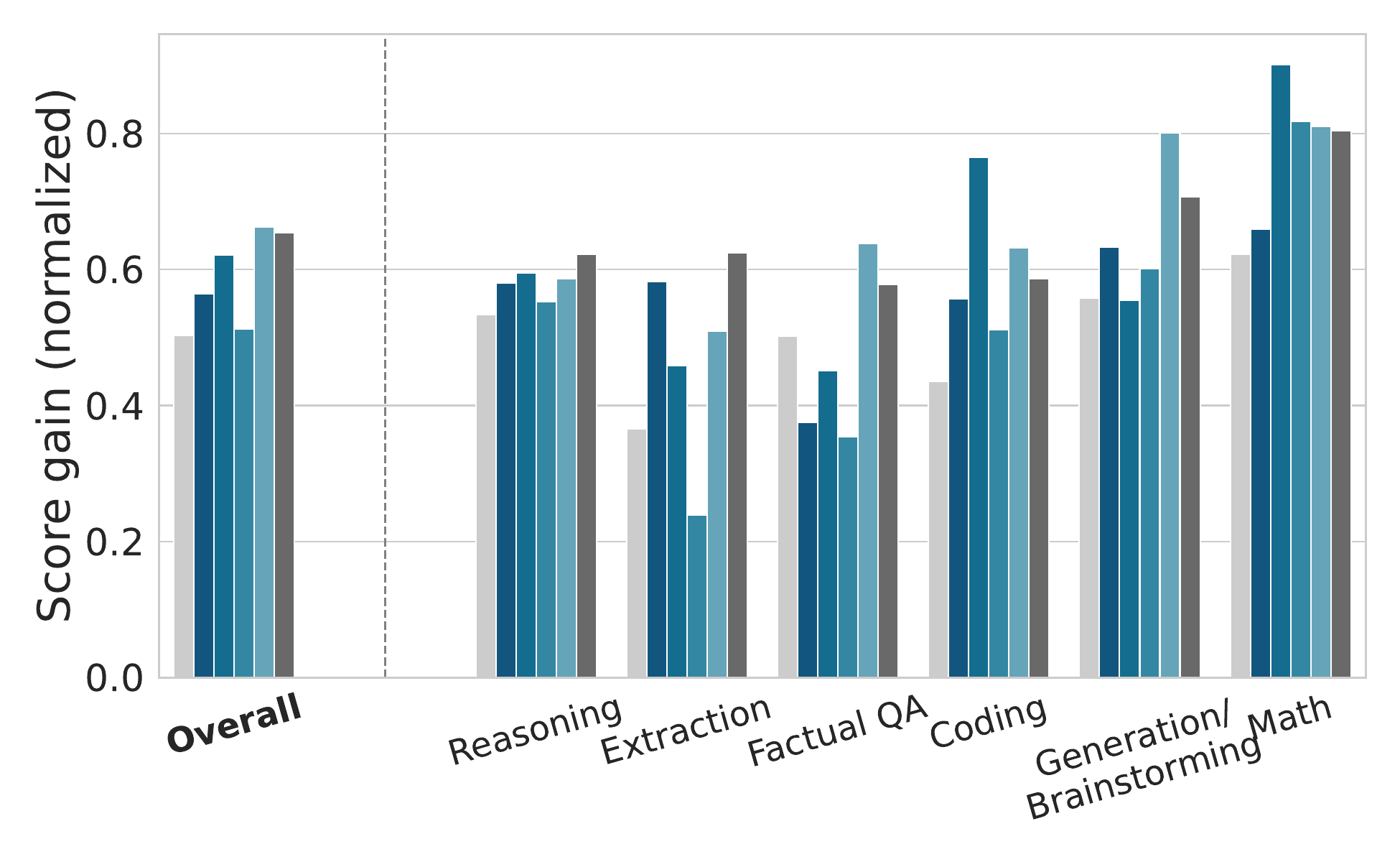}\vspace{-0.7em}
    \includegraphics[width=0.48\textwidth]{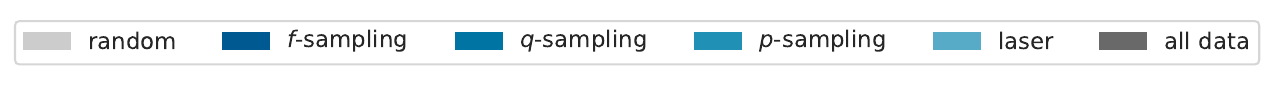}
    \caption{Average score gains over the base model when ablating \textsc{Laser} with 100k samples: with only $f$, $q$ and $p$ as ranking criteria. }
    \label{fig:eval_ablation}
\end{figure}

To better understand the contribution of each \textsc{Laser} component, we ablate sampling by ranking with only difficulty scores ($f$), quality scores ($q$), or their combination ($p$)---without clustering or stratification---and compare these variants to the full pipeline. As shown in Figure~\ref{fig:eval_ablation}, using $f$ or $q$ alone yields clear improvements over random sampling. However, their full potential is realized only when combined within the complete \textsc{Laser} pipeline. Notably, relying solely on $p$ for ranking provides only marginal gains, and even underperforms $f$- or $q$-based sampling individually, highlighting the importance of diversifying the sampling through stratification and clustering in \textsc{Laser}.

Hence, all three properties (difficulty, quality and diversity) are necessary but not sufficient for sampling data that consistently performs well. 
This is intuitive: highly challenging queries answered incorrectly provide little benefit, while accurate responses to trivial questions yield weak training signals.
Another salient result is the high performance of $q$-sampling for \emph{Coding} and \emph{Math}.
While the soundness of responses (i.e. $q$-scores) is arguably more important in these domains, we also find that $q$- and $f$-scores in our source data $\mathcal{D}$ are generally negatively correlated, with especially high negative correlations in Coding ($r=-.38$) and Math ($r=-.43$). 
Shifting emphasis onto the difficulty ($f$-scores and $p$-scores) might therefore reduce the response quality in these domains and explain the performance gaps observed in Figure~\ref{fig:eval_ablation}.
We find similar relationships between quality and difficulty for the source datasets $\mathcal{D}_{weak}$ and $\mathcal{D}_{strong}$ in the following section.

\vspace{-5pt}

\subsubsection{Additional robustness experiments}
\label{subsubsec:results_robustness}

We continue to demonstrate the robustness of the full \textsc{Laser} pipeline for different target sample sizes $m$ and variations in the source data distribution $\mathcal{D}$, such as differences in source data quality and skewed data distributions.

\begin{figure}[h!]
    \centering
    \includegraphics[width=\linewidth]{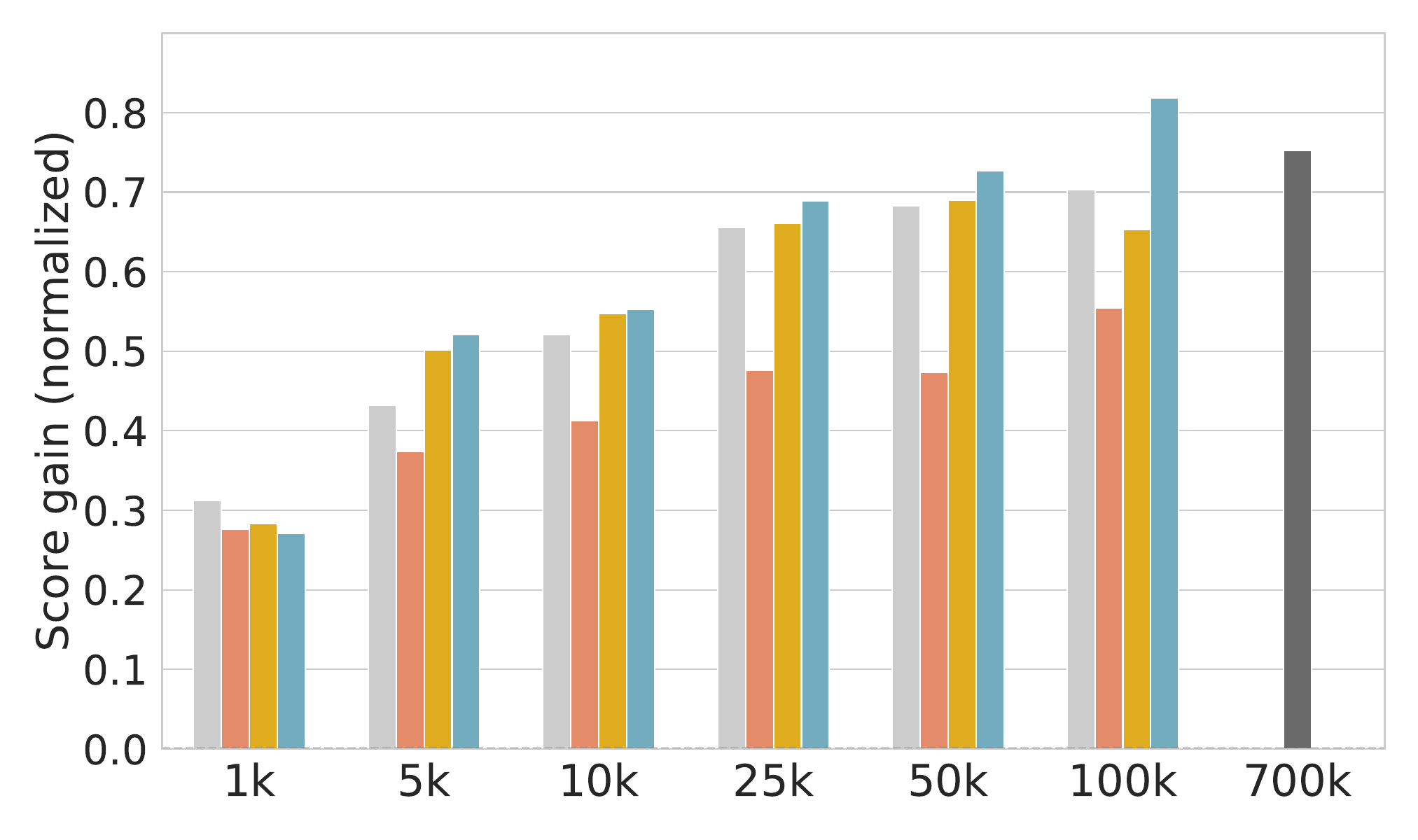}\vspace{-0.7em}
    \includegraphics[width=0.48\textwidth]{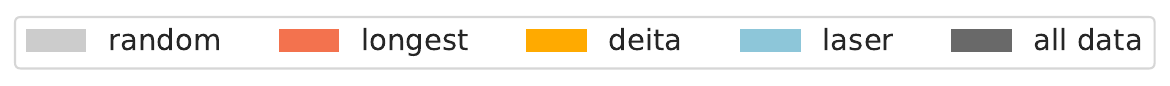}
    \caption{Average score gains over the base model, when finetuning Mistral-7B-Base using instruction datasets with varying sample sizes.}
    \label{fig:eval_scaling}
\end{figure}

\paragraph{Downscaling $m$.}
As different users may require instruction data sets of different sizes, a selection pipeline should improve sampling at every target scale. To verify this, we fix the base model to Mistral‑7B‑Base and constructs datasets $\mathcal{D}'$ of \{1k, 5k, 10k, 25k, 50k, 100k\} items using \textsc{Laser}, as well as \textit{random} and \textit{all data} as baselines. 
Figure~\ref{fig:eval_scaling} demonstrates that our method outperforms the alternatives across scales, except when $m=1,000$. Notably, with only $m=100,000$ (14\% of the source data), 
it surpasses the performance of full-data fine-tuning.
We provide benchmark-specific plots in Figure~\ref{fig:eval_scaling_all_benchmarks} (Appendix~\ref{app:results_details}), reported without normalizing \emph{score gain}.

\vspace{-5pt}
\paragraph{Effectiveness shifts in distribution of $\mathcal{D}$.}
Next, it is very probable that the average effectiveness of the source data $\mathcal{D}$ changes from scenario to scenario.
We therefore test \textsc{Laser} with two shifted source distributions: $\mathcal{D}_{\text{strong}}$ consisting of the full Tülu v3 \citep{lambert2024tulu}---a meticulously optimised dataset, which can be considered highly effective---and $\mathcal{D}_{\text{weak}}$ composed of multiple datasets that have previously proved to be less effective for instruction tuning (see Appendix~\ref{app:datasets:robustness} for details).

\begin{figure}[h!]
    \centering
    \includegraphics[width=\linewidth]{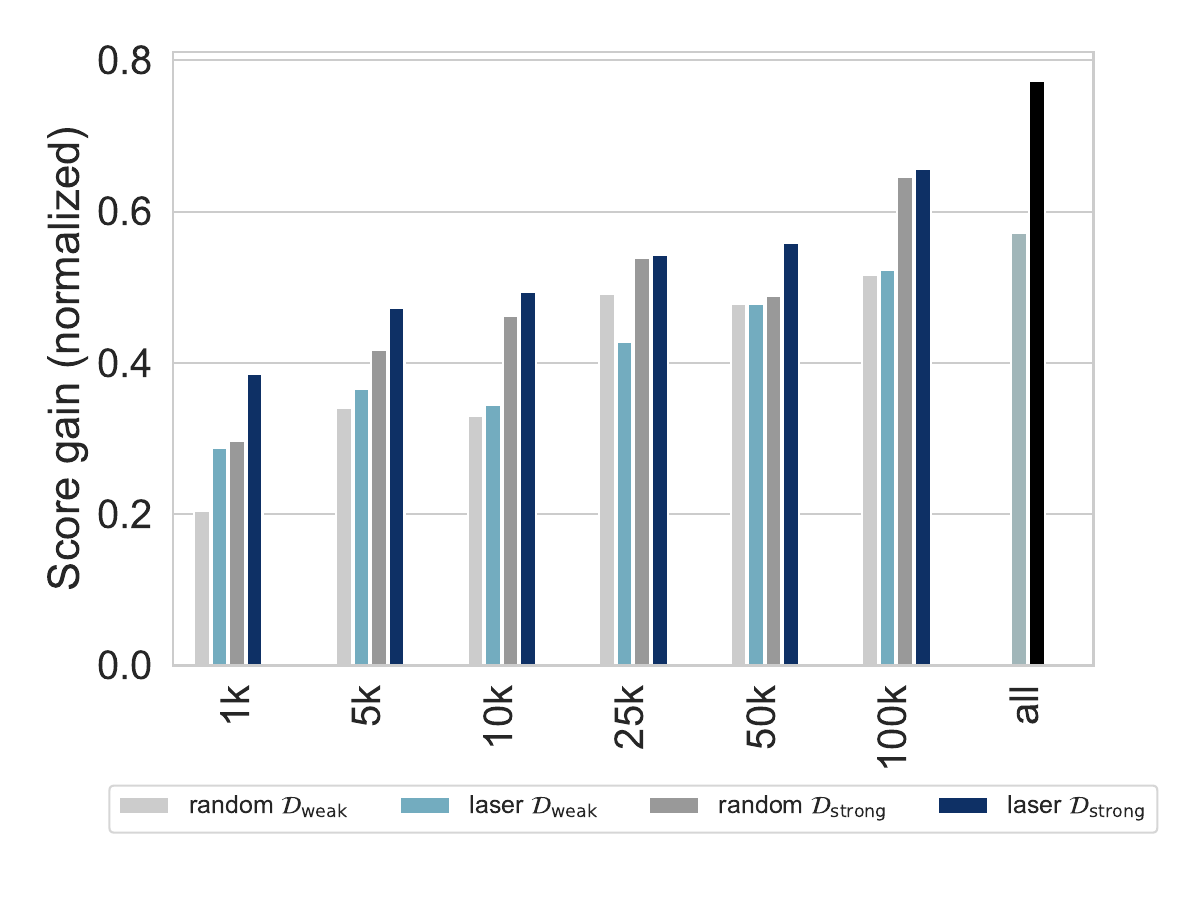}
    \caption{Performance on source datasets of different effectiveness ($\mathcal{D}_{\text{weak}}$ and $\mathcal{D}_{\text{strong}}$) for different sample sizes. For comparison, ``all'' shows performance when using full $\mathcal{D}_{\text{weak}}$ or $\mathcal{D}_{\text{strong}}$ without sampling.}
    \label{fig:eval_high_vs_lq_source} 
\end{figure}

We sample $m \in \{1\text{k}, 5\text{k}, 10\text{k}, 25\text{k}, 50\text{k}, 100\text{k}\}$ data points from both sources $\mathcal{D}_{\text{strong}}$ and $\mathcal{D}_{\text{weak}}$  using \textsc{Laser} and random sampling.
The results of finetuning Mistral-7B-v0.3 are shown in Figure~\ref{fig:eval_high_vs_lq_source}.
Generally, \textsc{Laser} performs robustly on different source datasets, with fewer fluctuations when changing the sampling size and comparably more improvements for the stronger than for the weaker source dataset. 
Interestingly, the performance gains using \textsc{Laser} over random are more pronounced for $\mathcal{D}_{\text{weak}}$ when sampling smaller target datasets, potentially distilling the few better examples from the weaker source.

\begin{figure}[th!]
    \centering
        \begin{subfigure}[l]{0.99\linewidth}
        \includegraphics[width=\linewidth]{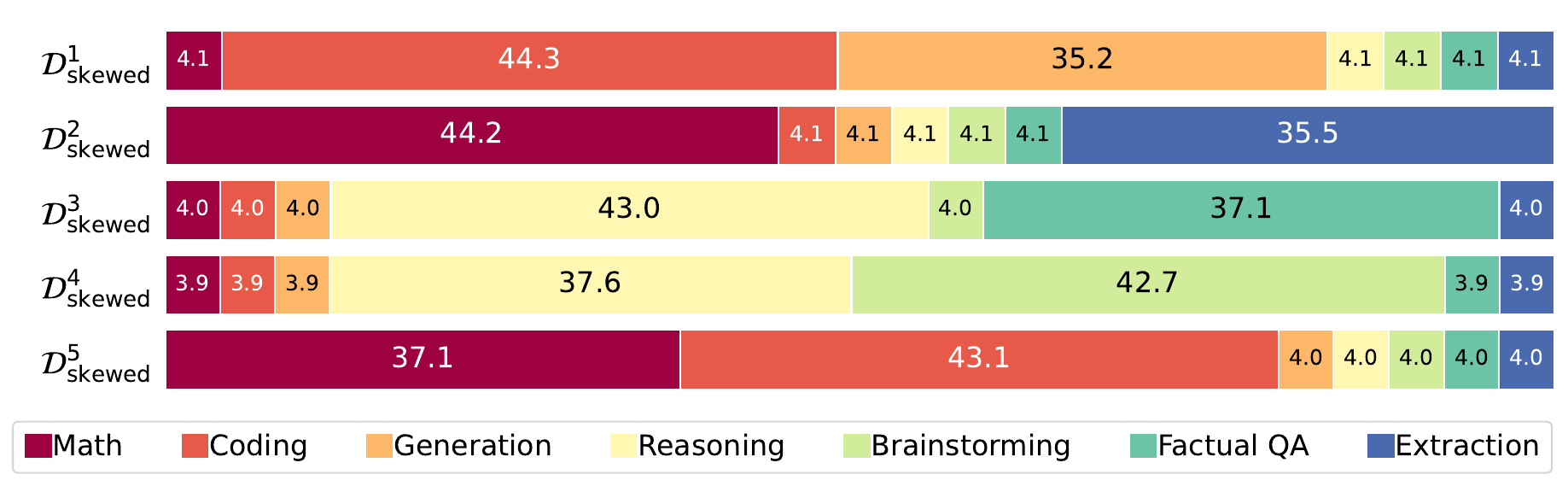}
    \caption{}
    \label{fig:porportions_skewed}
    \end{subfigure}
    \hfill
    \begin{subfigure}[l]{0.99\linewidth}
    \includegraphics[width=\linewidth]{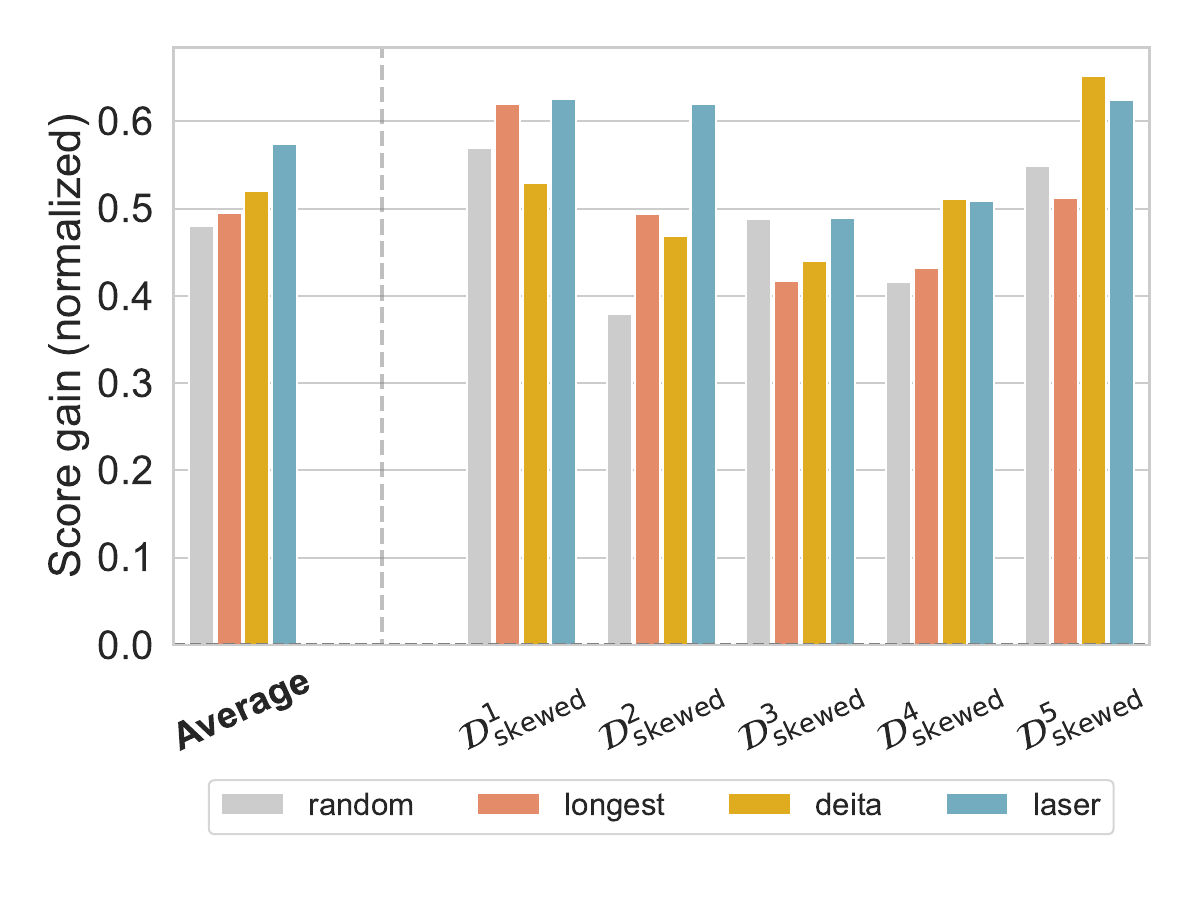}
    \caption{}
    \label{fig:results_skewed}
    \end{subfigure}
    \caption{(a) Category proportions of different $\mathcal{D}_\text{skewed}$ source distributions; (b) Average benchmark scores, when finetuning Mistral-7B-Base with data sampled from the skewed distribution using various strategies.}
    \label{fig:skewed_D}
\end{figure}

\vspace{-5pt}
\paragraph{Skewed distribution of $\mathcal{D}$.}
Ultimately, $\mathcal{D}$ might be very homogeneous (i.e. contains mostly a specific data type).
To simulate this, we create five \emph{skewed} source sets $\mathcal{D}_{\text{skewed}}$
by randomly selecting two categories $l_1,l_2\in\mathcal{L}$ for each $\mathcal{D}_{\text{skewed}}$
and retaining only items with labels $l_i \in {l_1, l_2}$, along with a small (4\%) residue from all other categories, yielding a strongly biased data distribution. 
We show the proportions of each category for all five new source datasets in Figure~\ref{fig:porportions_skewed}.

From each $\mathcal{D}_{\text{skewed}}$, we sample 25k items with the same strategies as previously and fine‑tune Mistral‑7B‑Base on each of the resulting $\mathcal{D}_{\text{skewed}}'$.
We expected diversity-aware sampling strategies like \textsc{Deita} and \textsc{Laser} to mitigate bias by enforcing either semantic spread and explicit quotas.
Indeed, the diversity-aware sampling strategies on average outperform the naive methods, with \textsc{Laser} having an edge over \textsc{Deita}, in terms of performance improvements as illustrated in Figure~\ref{fig:results_skewed}.
Upon closer inspection, we find that only \textsc{Laser} reduces the introduced bias of categories whereas \textsc{Deita} exacerbates it (see Appendix~\ref{app:datasets:robustness} for details).
Next to that, we observe a key limitation of \textsc{Deita}'s sampling method in this setup:
only adding dissimilar examples leads to quick saturation when source data is homogeneous and, thereby, hitting a ceiling of possible samples that can be sampled with \textsc{Deita}.

\section{Discussion and Conclusion}
\label{sec:discussion}
Effective data selection for instruction tuning is increasingly important as we handle the increasing amount of available data of mixed quality.
Nevertheless, the results of data selection pipelines have often been brittle, struggling to generalise across training setups \citep{diddee-ippolito-2025-chasing,zhao_long_2024}.

In this paper, we address this challenge by explicitly covering the whole spectrum of possible instruction domains and tailoring scoring strategies that are apt to judge the utility of samples in those domains.
We show the robustness of our approach by testing it across diverse settings (model families, scales, various shifts in the source data distribution) and evaluating it on a wide range of common benchmarks.
Our experiments provide further evidence for the necessity of data selection, as our sampling not only significantly reduces the amount of data required for training, but also outperforms setups trained on the full source data.
Despite its more elaborate design---including multiple scoring methods---our pipeline remains computationally efficient due to targeted scoring.

While our pipeline outperforms baselines in almost all settings, it shows only little improvements when sampling large portions from weak source data (as shown in Section~\ref{subsubsec:results_robustness}). 
This may reflect a potential ceiling effect on possible performance given the weak source material, suggesting that improving the data itself---rather than relying solely on elaborate sampling techniques---might be necessary. In such cases, iterative data refinement seems promising, especially when guided by reliable scorers for assessing sample difficulty and quality.
The modular setup of \textsc{Laser} allows for further incremental improvements of different scorers in future work.

\section*{Limitations}
\label{sec:limitations}

\paragraph{Difficulty scorer.} A major issue in collecting data for our difficulty scorer is a certain unreliability in the evaluation of model responses. 
Previous research shows that evaluations oftentimes have weak robustness to e.g. the prompt formatting \citep{weber2023icl,weber2023mind,sclar2024quantifying,polo2024efficient}, bias of LLMs-as-a-judge \citep{panickssery2024llm,stureborg2024large,wataoka2024self} or errors in post-processing (such as issues in extracting the answer from the model response).
For example, GPT4o showed overall weaker performance on some of our evaluated subsets than some small open-source models. 
Upon closer inspection, we encountered that---while providing the correct answer---GPT4o generally tends not to follow the formatting of the given few-shot examples and rather responds in an open-form manner, resulting in failing the tight response search masks of the used evaluation frameworks.
While we try to mitigate this issue as much as possible, we cannot guarantee that difficulty scores exactly reflect a model's capacity to solve a given data point.

\vspace{-5pt}
\paragraph{Quality scorer.} We did not develop nor have dedicated quality scorers for samples belonging to \emph{Reasoning}, \emph{Factual QA} and \emph{Extraction} categories. 
However, we hypothesize that process reward models (PRMs) could be adapted to \emph{Reasoning} tasks beyond math, such as spatial \citep[e.g.,][]{wu2024minds} and deductive reasoning \cite[e.g.,][]{seals-shalin-2024-evaluating}, given appropriate training data. 
\emph{Factuality} assessment is a long-standing research problem that has become more relevant in the era of LLMs. \citet{wei2024longform} introduced \emph{SAFE (Search-Augmented Factuality Evaluator)}, an LLM-agent-based method for automatically assessing long-form factuality in model response. Although significantly cheaper than human annotators (up to $20\times$), SAFE still incurs costs of \$20--\$40 per 100 prompt-response pairs.
For \emph{Extraction} tasks, future work might draw inspiration from recent advances in RAG evaluation \citep{Yu_2025}. These directions, however, are beyond the scope of this paper.

\vspace{-5pt}
\paragraph{Performance with weak source data.} Our results indicate modest performance gains in setups with very weak source data. However, there might be a ceiling for possible performance in such settings, where data selection alone might not be sufficient to achieve good performance. In such cases, iterative data refinement appears to be a promising approach, particularly when supported by reliable scorers for assessing sample difficulty and quality.


\section*{Acknowledgments}
We thank Rishiraj Saha Roy, Lucas Druart and Christian Kroos for their valuable feedback on earlier versions of this manuscript.
This work was supported in part by the German Federal Ministry for Economic
Affairs and Climate Action (BMWK) through the OpenGPT-X project (no. 68GX21007D) and in part by the Free State of Bavaria in the DSgenAI project (Grant Nr.: RMF-SG20-3410-2-18-4). 
We also gratefully acknowledge the Centre for Information Services and High Performance Computing [Zentrum für Informationsdienste und Hochleistungsrechnen (ZIH)] and ScaDS.AI at the Technical University of Dresden for providing its facilities for experimental work.

\bibliography{custom_rebibbed,InstructionTuning_rebibbed}
\appendix
\clearpage
\section{Appendix}
\label{app}

\subsection{Instruction Categorization -- details}
\label{app:instr_category}

We present the zero-shot prompt for classifying instructions in Figure~\ref{fig:zero-shot-prompt}. As for training the SetFit classifier, we leverage the training data detailed in Table~\ref{tab:setfit-training-data}, and we employ the following hyperparameters. Note that training with SetFit consists of two phases behind the scenes: finetuning embeddings and training a differentiable classification head. As a result, some of the training arguments can be tuples, where the two values are used for each of the two phases, respectively.
\setlist{nolistsep}
\begin{itemize}[leftmargin=*,topsep=0pt,noitemsep]
    \item \texttt{batch\_size}=(16, 2)
    \item \texttt{num\_epochs}=(1, 15)
    \item \texttt{end\_to\_end}=True (train the entire model end-to-end during the classifier training phase)
    \item \texttt{body\_learning\_rate}=(2e-5, 1e-5), the second value is the learning rate of the Sentence Transformer body during the classifier training phase
    \item \texttt{head\_learning\_rate}=1e-4
    \item \texttt{max\_steps}=500
\end{itemize}

\subsection{Difficulty Scorer -- details}
\label{app:diff_scorer}

\subsubsection{Generating $\mathcal{D}_{\text{diff}}$}
\label{app:difficulty_scorer_train_data}

\begin{figure}[h!]
    \centering
    \includegraphics[width=\linewidth]{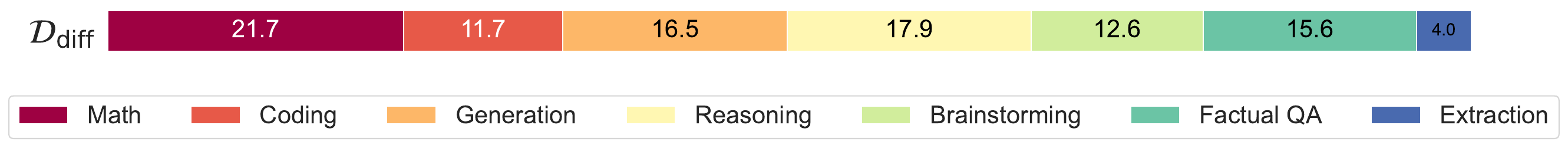}
    \caption{Proportions of different categories in the difficulty scorer training set $\mathcal{D}_{\text{diff}}$ after cleaning.}
    \label{fig:diff_scorer_setfit_proportions}
\end{figure}

Figure~\ref{fig:diff_scorer_setfit_proportions} shows the proportions of different instruction-response pairs that we use as the training data for the difficulty scorer.
In Table~\ref{tab:diff_scorer_training_datasets} we report the benchmark data sources from which we obtained the data, as well as the type of evaluation metric that we use to evaluate the 18 LLMs from Table~\ref{tab:diff_scorer_evaluated_models}.

\begin{table}[!t] 
\centering
\small
\begin{tabular}{@{\hskip 2pt}l@{\hskip 4pt}l@{\hskip 4pt}r@{}} 
\toprule
\textbf{Model} & \textbf{Type} & \textbf{Size} \\
\midrule
SmolLM2-135M-Instruct \tiny\citep{allal2025smollm2smolgoesbig}\normalsize & Univ. & 135M \\
SmolLM2-360M-Instruct \tiny\citep{allal2025smollm2smolgoesbig}\normalsize & Univ. & 360M \\ \arrayrulecolor{black!30}\midrule
Qwen2.5-0.5B-Instruct \tiny\citep{qwen2.5}\normalsize & Univ. & 0.5B \\ \midrule
Qwen2.5-Math-1.5B-Instruct \tiny\citep{yang2024qwen25mathtechnicalreportmathematical}\normalsize & Math & 1.5B \\
Qwen2.5-Coder-1.5B-Instruct \tiny\citep{hui2024qwen2}\normalsize & Code & 1.5B \\
Qwen2.5-1.5B-Instruct \tiny\citep{qwen2.5}\normalsize & Univ. & 1.5B \\
SmolLM2-1.7B-Instruct \tiny\citep{allal2025smollm2smolgoesbig}\normalsize & Univ. & 1.7B \\ \midrule
Qwen2.5-Math-7B-Instruct \tiny\citep{yang2024qwen25mathtechnicalreportmathematical}\normalsize & Math & 7B \\
Qwen2.5-Coder-7B-Instruct \tiny\citep{hui2024qwen2}\normalsize & Code & 7B \\
Qwen2.5-7B-Instruct \tiny\citep{qwen2.5}\normalsize & Univ. & 7B \\
Mistral-7B-Instruct-v0.3 \tiny\citep{jiang2023mistral7b}\normalsize & Univ. & 7B \\ \midrule
Qwen2.5-14B-Instruct \tiny\citep{qwen2.5}\normalsize & Univ. & 14B \\ \midrule
Mistral-Small-24B-Instruct-2501 \tiny\citep{mistral_small_3}\normalsize & Univ. & 24B \\ \midrule
Qwen2.5-32B-Instruct \tiny\citep{qwen2.5}\normalsize & Univ. & 32B \\
Qwen2.5-Coder-32B-Instruct \tiny\citep{hui2024qwen2}\normalsize & Code & 32B \\
Qwen3-32B \tiny\citep{qwen3}\normalsize & Univ. & 32B  \\ \midrule
gpt-4o-mini-2024-07-18 \tiny\citep{hurst2024gpt}\normalsize & Univ. & unk \\
gpt-4o-2024-08-06 \tiny\citep{hurst2024gpt}\normalsize & Univ. & unk \\
\arrayrulecolor{black}\bottomrule
\end{tabular}
\caption{Models that were evaluated to obtain difficulty scores for our difficulty scorer training set.}
\label{tab:diff_scorer_evaluated_models}
\end{table}

During the data collection for the difficulty scorer, we collect training sets from different benchmarks as well as data points from OpenAssistent \citep{kopf2023openassistant} as samples of open-ended generation. 
We evaluate these samples using the existing evaluation frameworks \emph{LM-evaluation harness} \citep{eval-harness}, \emph{big-code evaluation harness} \citep{bigcode-evaluation-harness} and \emph{FastChat} \citep{zheng2023judging}. 
For the LLM-as-a-judge evaluation, we use GPT4o as the judge and evaluate only a subset of 4 LLMs from Table~\ref{tab:diff_scorer_evaluated_models} that we deem representative in terms of capabilities (gpt-4o-2024-08-06, Mistral-Small-24B-Instruct-2501, Mistral-7B-Instruct-v0.3 and SmolLM2-1.7B-Instruct).

\begin{figure}
    \centering
    \includegraphics[width=\linewidth]{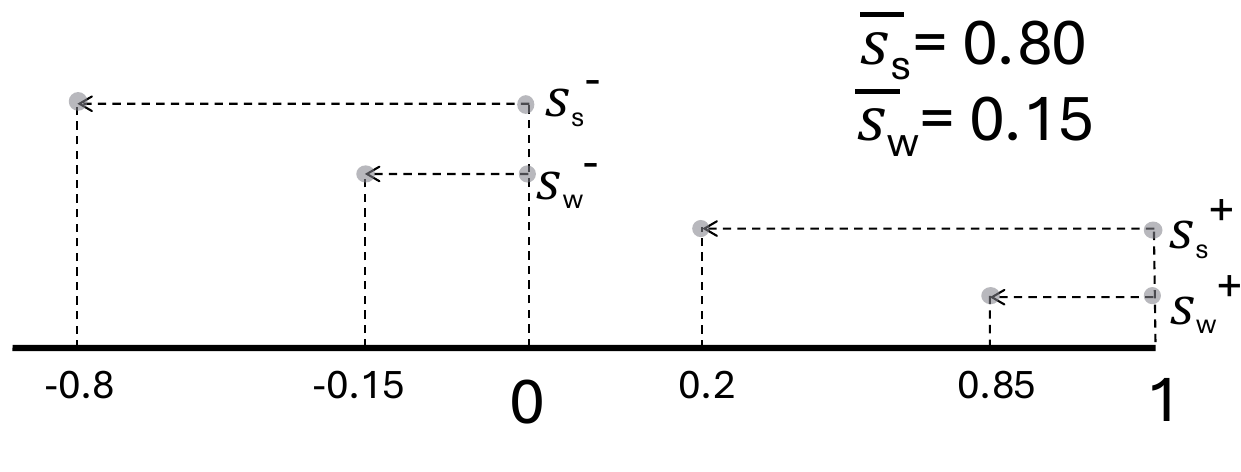}
    \caption{An illustration of calculating the \emph{deviation of the average} per dataset: Assuming we have a strong and a weak model with the average scores of $\bar{s}_{\text{s}} = 0.8$ and $\bar{s}_{\text{w}} = 0.15$ respectively. We subtract these averages from the respective item scores (here two examples with $s^+ = 1$ and $s^- = 0$). We can see how an incorrect response of the strong model produces a much more negative score than an incorrect response from the weak model (compare $s_{\text{s}}^-$ and $s_{\text{w}}^-$). In other words, when a strong model gets an item wrong, this is much more meaningful for its difficulty compared to the weak model. Similarly, when a weak model correctly solves a datapoint, it is a much more meaningful signal for the data point's ease than when a strong model is correct.}   
    \label{fig:relative_performance_illustration}
\end{figure}

Figure~\ref{fig:relative_performance_illustration} demonstrates the preprocessing step of calculating the deviation from the average as mentioned in Section~\ref{subsubsec:difficulty_scoring}.

\subsubsection{Difficulty scorer training}
\label{app:diff_scorer:training_details}

We equip regular CausalLLMs with a regression head by pooling the final hidden states and adding a linear projection to a scalar output, then finetune these models on the difficulty scoring task using the hyperparameters detailed in Table~\ref{tab:diff_scorer_hyperparams}.
We evaluate four different base models as difficulty scorer---Llama-3.1-8B \citep{grattafiori_llama_2024_reduced}, Qwen-2.5-7B \citep{qwen2.5}, Qwen3-4B and Qwen3-8B \citep{qwen3})---and select Qwen3-8B, which achieves the best performance on in-distribution evaluation data.

\begin{table}[!t]
\centering
\small
\begin{tabular}{@{}lc@{}}
\toprule
\textbf{Hyperparameter} & \textbf{Value} \\
\midrule
batch\_size & 1  \\
gradient\_accumulation & 16  \\
learning\_rate & 1e-5  \\
lr\_scheduler\_type & linear \\
num\_train\_epochs & 8 \\
warmup\_steps & 100 \\
max\_seq\_length & 2048 \\
weight\_decay & 0.01 \\
neftune\_noise\_alpha & 10 \\
\arrayrulecolor{black}\bottomrule
\end{tabular}
\caption{Hyperparameter details for training the difficulty scorer}
\label{tab:diff_scorer_hyperparams}
\end{table}

\subsubsection{Effectiveness}
\label{app:diff_scorer:effectiveness}
\begin{figure}[h!]
\centering
    \begin{subfigure}[b]{\linewidth}
        \includegraphics[width=\linewidth]{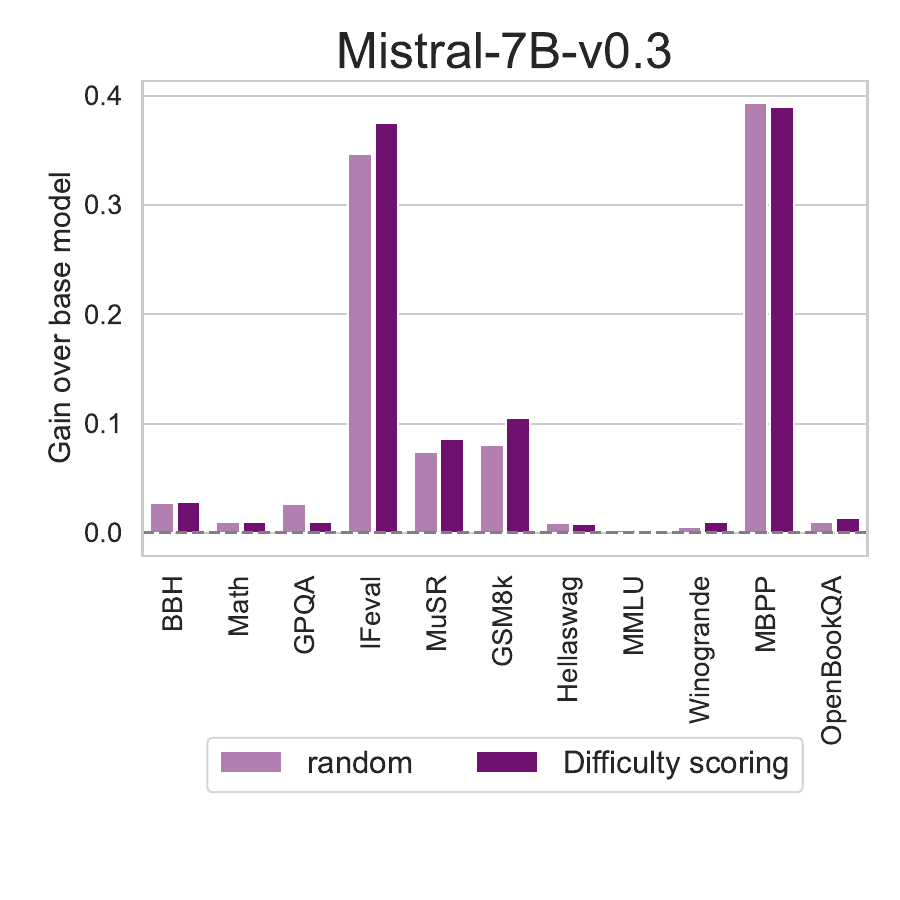}
        \caption{}
        \label{fig:scorer_eval_difficulty}
    \end{subfigure}
\hfill
    \begin{subfigure}[b]{\linewidth}
        \includegraphics[width=\linewidth]{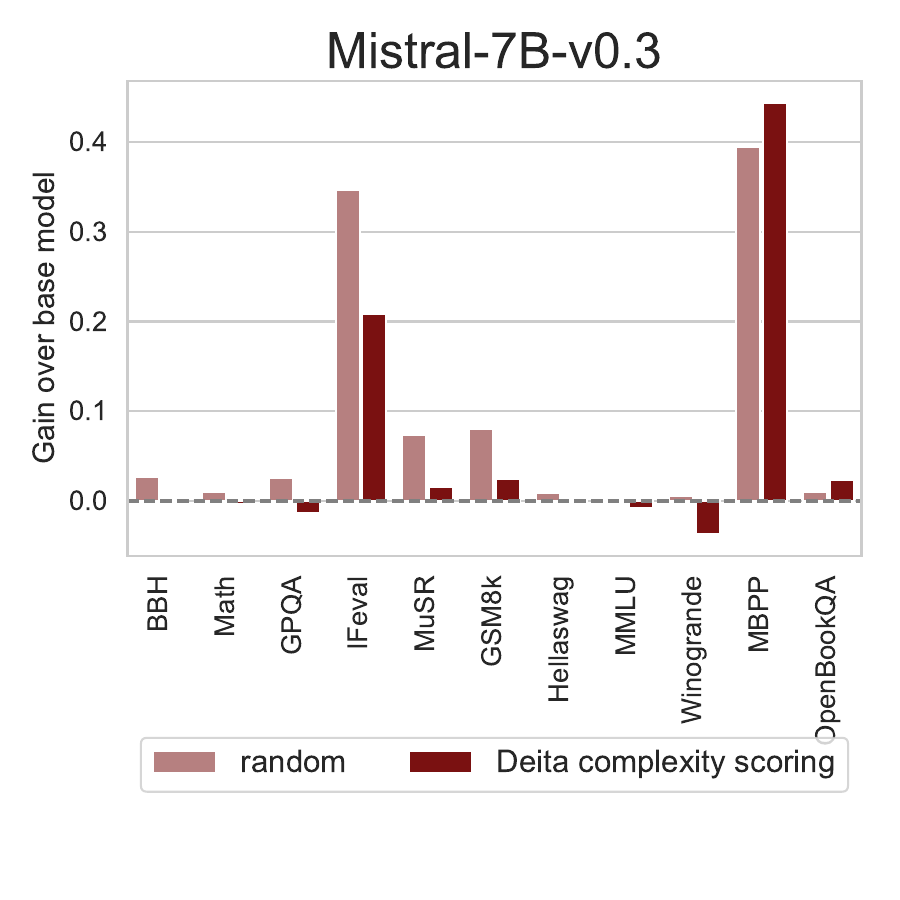}
        \caption{}
        \label{fig:scorer_eval_complexity}
    \end{subfigure}
\caption{(a) Performance comparison of difficulty scorer vs. random sampling; (b) Performance comparison of complexity scorer vs. random sampling}
\label{fig:scorer_eval_difficulty_complexity}
\end{figure}

To evaluate the effectiveness of the difficulty scorer for data selection across benchmarks, we sample 25k data points from the datasets in Table~\ref{tab:datasets}, either at random or based on \textsc{Laser} \emph{difficulty score} ranking. The results in Figure~\ref{fig:scorer_eval_difficulty} show that difficulty-based sampling outperforms or is comparable to random sampling on all benchmarks except GPQA. For comparison, Figure~\ref{fig:scorer_eval_complexity} reports results using the \textsc{Deita} \emph{complexity scorer} under the same setup.

\begin{figure}[h!]
    \centering
    \begin{subfigure}[b]{0.49\linewidth}
    \includegraphics[width=\linewidth]{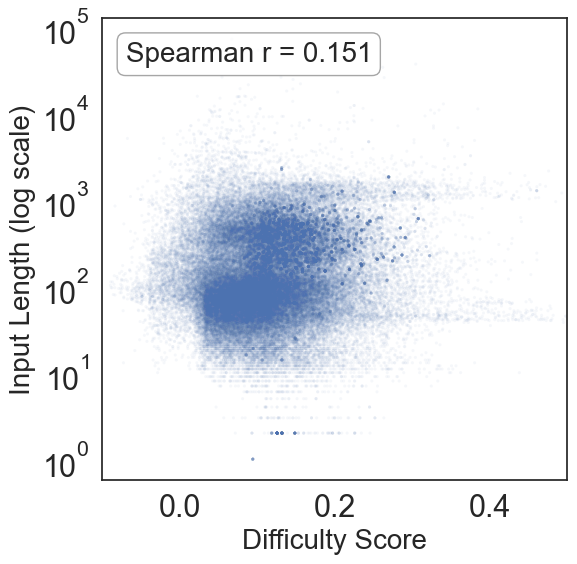}
    \end{subfigure}
    \hfill
    \begin{subfigure}[b]{0.49\linewidth}
    \includegraphics[width=\linewidth]{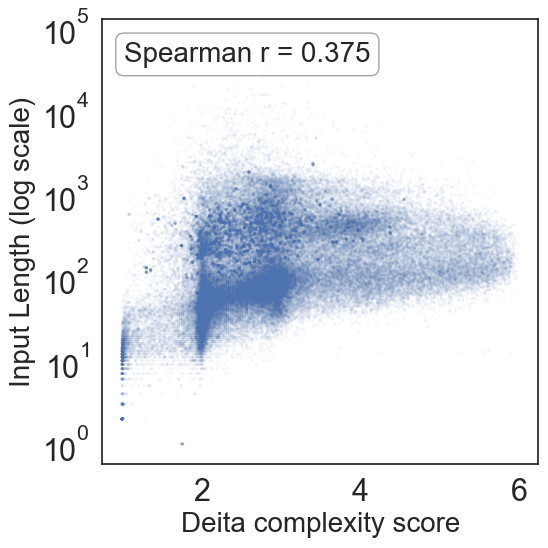}
    \end{subfigure}
    \caption{Relation between complexity/difficulty scores and length of the input sequence. The length of the input sequence should, in most cases, be unrelated to its difficulty.}
\label{fig:corr_length_x_complexity}
\end{figure}

\begin{figure}[h!]
    \centering
    \begin{subfigure}[b]{0.49\linewidth}
    \includegraphics[width=\linewidth]{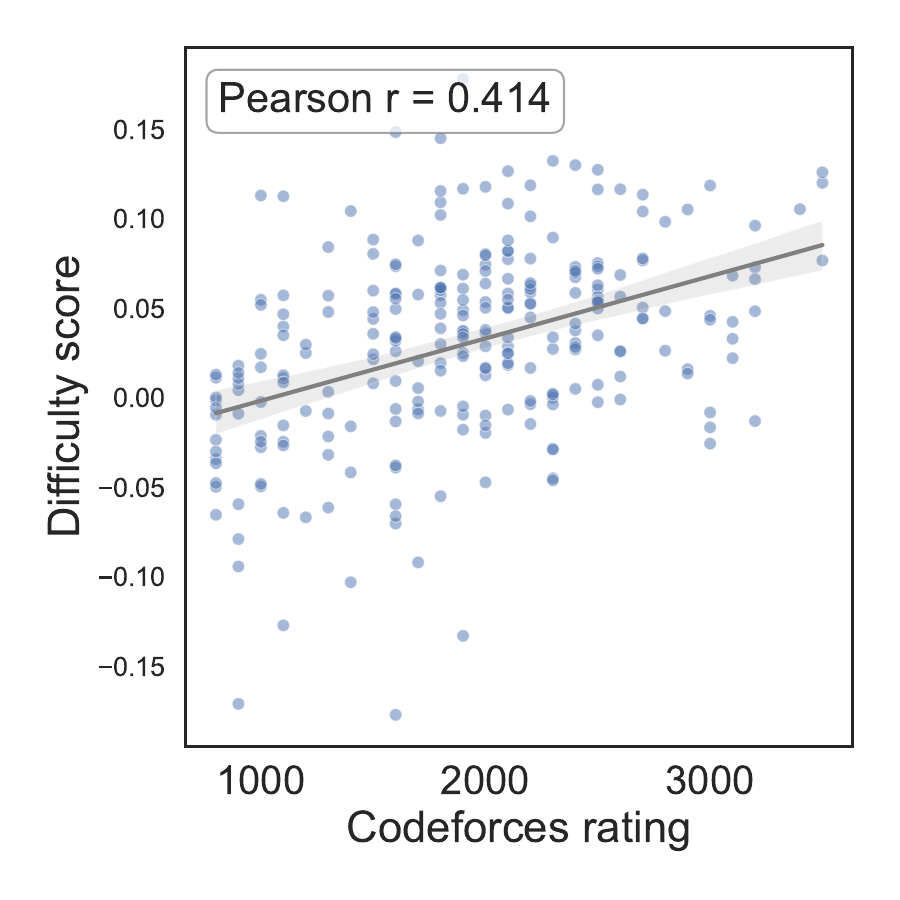}
    \end{subfigure}
    \hfill
    \begin{subfigure}[b]{0.49\linewidth}
    \includegraphics[width=\linewidth]{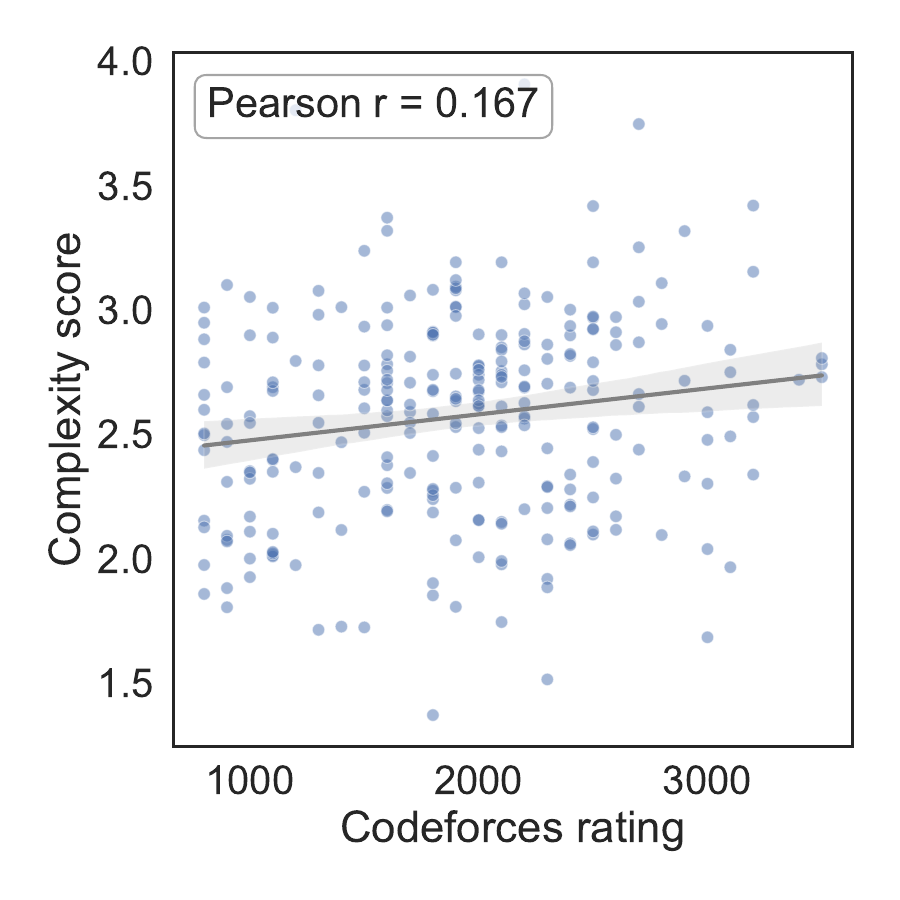}
    \end{subfigure}
    \caption{Relation between complexity/difficulty scores and the human code-forces difficulty scores. Difficulty scores show higher correlation to human scores than complexity scores.}
\label{fig:cc_x_difficulty}
\end{figure}

\begin{figure}[h!]
    \centering
    \includegraphics[width=0.5\linewidth]{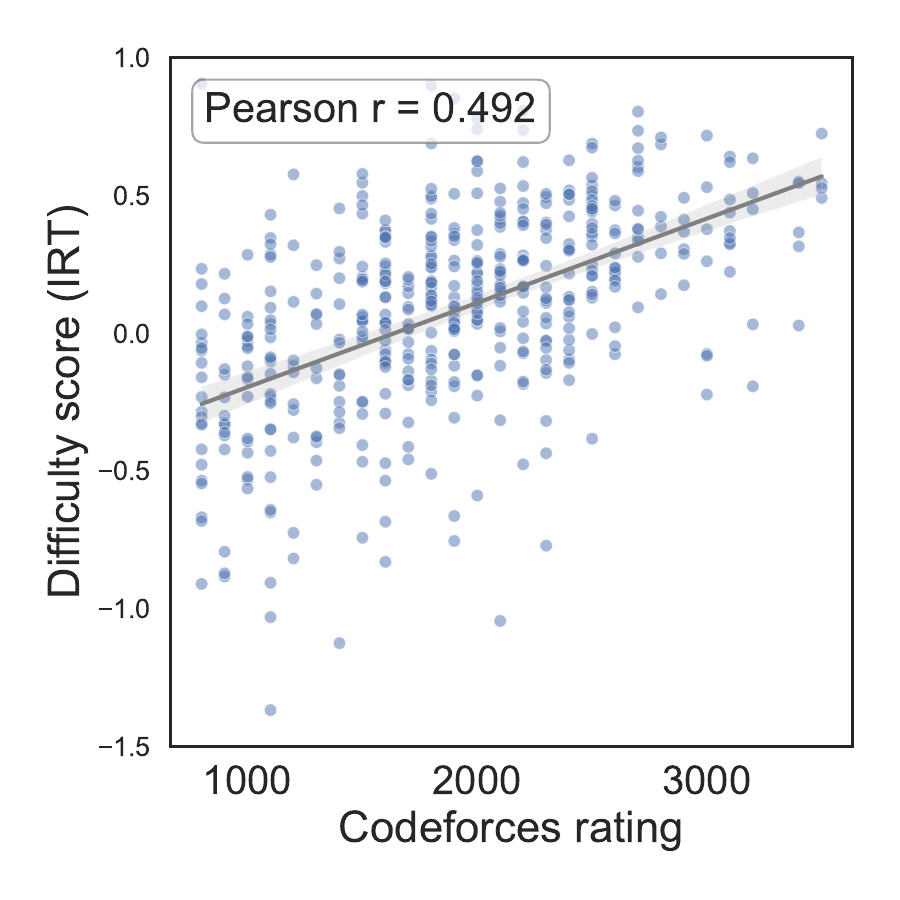}
    \caption{Correlation of difficulty scores with Codeforce ratings when the difficulty scorer is trained on IRT $\beta$s as targets (as described in Appendix~\ref{app:difficulty_scorer_train_data}). 
    As we can see here, IRT $\beta$s yields higher validity with the external criterion. However, we find it not to benefit data selection.}   
    \label{fig:cc_x_difficulty_IRT}
\end{figure}

\subsubsection{Validity}
\label{app:diff_scorer:validity}

We further evaluate whether the difficulty scores are akin to the notion of difficulty in humans or whether our scorer fits a spurious feature of our training data. 

We first test it against instruction length, a common spurious feature of difficulty/complexity scorers.
While difficult instructions are plausibly often longer than easier instructions, there is no causal relationship between the length and difficulty. 
A scorer that is trained to predict difficulty should therefore not rely on input length as a feature to rely its prediction upon.
We show in Figure~\ref{fig:corr_length_x_complexity} how our difficulty scorer is weakly related to the spurious feature of instruction length, while the correlation of \textsc{Deita} complexity is much higher.

Besides showing how the difficulty scorer does not rely on instruction length, we also use an external criterion to evaluate whether it measures difficulty in the human sense. For this end, we score the CodeForces section of DeepMind's code-contests dataset \citep{li2022competition} with our difficulty scorer as well as the \textsc{Deita} complexity scorer and correlate the scores with the given cf\_rating.
Figure~\ref{fig:cc_x_difficulty} shows how the difficulty scorer is much better at predicting human difficulty scores than the \textsc{Deita} complexity scorer.

\subsubsection{Experimentation with IRT-based targets}

We further experimented with difficulty targets grounded in psychometric methods called item response theory \citep[IRT; see][]{lord1968statistical,brzezinska2020item,van2018handbook,cai2016item}. 
IRT provides a framework for estimating latent traits from observed responses and has recently been applied to the evaluation of large language models \citep[see e.g.][]{lalor2016building,rodriguez2021evaluation,vania2021comparing,zhuang2023efficiently,polo2024tinybenchmarks}.
In our setting, an IRT model estimates item-level parameters from correctness scores across the LLMs from Table~\ref{tab:diff_scorer_evaluated_models}. 
We fit such a model separately on each dataset in $\mathcal{D}_{\text{diff}}$ and use the resulting item difficulty parameters $\beta$ as targets for training our difficulty scorer.
We find that the IRT-based difficulty scorer yields improved results on the correlations with the Codeforces ratings (the validity criterion; see Figure~\ref{fig:cc_x_difficulty_IRT}).
At the same time, this IRT-based difficulty scorer does not yield better results when used for data selection.
We see the use of IRT-based difficulty scorers as an exciting future research direction, as the possibilities to gain more detailed insights into data point features are great (e.g. by using 2PL or 3PL instead of 1PL models or by extending from unidimensional to multidimensional models).

\subsection{Quality Scorers -- details}
\label{app:quality_scorer}

\subsubsection{Instruction-following Scorer}
\label{app:quality_scorer:if}

Prompts used by LLM-annotator/judge for deriving instruction-following scores can be found in Figure \ref{fig:few-shot-prompt-constraint-identification}, \ref{fig:prompt-constraint-verification} and \ref{fig:prompt-response-evaluation}. As our test dataset, we collected responses from 10 LLMs on the IFEval benchmark (see Table~\ref{tab:models-ifeval}), available on 
open-llm-leaderboard's dataset collection of evaluation details.\footnote{\url{https://huggingface.co/open-llm-leaderboard}}
 Table~\ref{tab:if-scoring-evaluation} presents the evaluation results of various LLMs as the annotator/judge.

\begin{table}[t]
\centering
\small
\adjustbox{max width=.48\textwidth}{%
    \begin{tabular}{@{}lc@{}}
        \toprule
        \textbf{Model} & \textbf{IFEval score} \\
        \midrule
        meta-llama/Llama-3.3-70B-Instruct & 0.90 \\
        Qwen/Qwen2.5-14B-Instruct-1M & 0.84 \\
        allenai/Llama-3.1-Tulu-3-70B-SFT & 0.81 \\
        tiiuae/Falcon3-7B-Instruct & 0.76 \\
        ibm-granite/granite-3.1-8b-instruct & 0.72 \\
        microsoft/Phi-3-medium-128k-instruct & 0.60 \\
        abacusai/Smaug-34B-v0.1 & 0.50 \\
        Qwen/Qwen2.5-32B & 0.41 \\
        google/gemma-1.1-2b-it & 0.31 \\
        databricks/dolly-v2-7b & 0.20 \\
        \arrayrulecolor{black}\bottomrule
        \end{tabular}}
    \caption{Performance of 10 considered LLMs on IFEval.}
    \label{tab:models-ifeval}
\end{table}

 \begin{table}[!t] 
\centering 
\scriptsize
\begin{adjustbox}{width=0.48\textwidth}
\begin{tabular}{@{}l@{}c@{}c@{~~~}c@{}}
\toprule
\multirow{2}{*}{\textbf{LLM annotator/judge}} & \textbf{macro} & \multicolumn{2}{c}{\textbf{Pearson's $r$}} \\
 & \textbf{F1} & \textbf{instance-level} & \textbf{model-level} \\
\midrule
Qwen/Qwen2.5-7B-Instruct & 0.80 & 0.503 & 0.986 \\
meta-llama/Llama-3.1-8B-Instruct & 0.83 & 0.454 & 0.982 \\
tiiuae/Falcon3-10B-Instruct & 0.84 & 0.515 & 0.969 \\
\arrayrulecolor{black!30}
Qwen/Qwen3-14B $\dagger$ & \textbf{0.86} & \textbf{0.523} & \textbf{0.995} \\
\arrayrulecolor{black}\bottomrule
\end{tabular}
\end{adjustbox}
\caption{Evaluation results of various LLMs as annotator/judge on identifying expressed constraints (macro-F1), and Pearson correlations coefficient ($r$) of resulting instruction-following scores with IFEval benchmarks scores at both instance-level and model-level. $\dagger$ denotes the chosen LLM annotator/judge for our instruction-following scorer.}
\label{tab:if-scoring-evaluation}
\end{table}

\begin{table}[!t] 
\centering 
\small
\begin{adjustbox}{width=0.48\textwidth}
\begin{tabular}{@{}lcc@{}}
\toprule
\textbf{LLM reviewer} & \textbf{acc} & \textbf{Pearson's $r$} \\
\midrule
deepseek-ai/DeepSeek-Coder-V2-Lite-Instruct & 0.64 & 0.278 \\
deepseek-ai/DeepSeek-R1-Distill-Qwen-14B & 0.66 & 0.389 \\
Qwen/Qwen2.5-Coder-14B-Instruct $\dagger$ & \textbf{0.70} & \textbf{0.412} \\
\arrayrulecolor{black!30}\midrule
Qwen/Qwen3-14B \emph{(non-coding LLM)} & \textbf{0.70} & 0.411 \\
\arrayrulecolor{black}\bottomrule
\end{tabular}
\end{adjustbox}
\caption{Evaluation results of various LLMs as code reviewer (acc), and Pearson correlation coefficient ($r$) of resulting code quality scores with LiveCodeBench benchmarks binary correctness labels. $\dagger$ denotes the chosen LLM annotator/judge for our instruction-following scorer.}
\label{tab:code-scoring-evaluation}
\end{table}


\subsubsection{Code-quality Scorer}
\label{app:quality_scorer:code}
The prompt used by LLM-annotator/judge for deriving code-quality scores can be found in Figure~\ref{fig:zero-shot-prompt-code-score}. 
Table~\ref{tab:code-scoring-evaluation} presents the evaluation results of various LLMs as the code reviewer.

\begin{figure*}[th!]
    \centering
    \includegraphics[width=\linewidth]{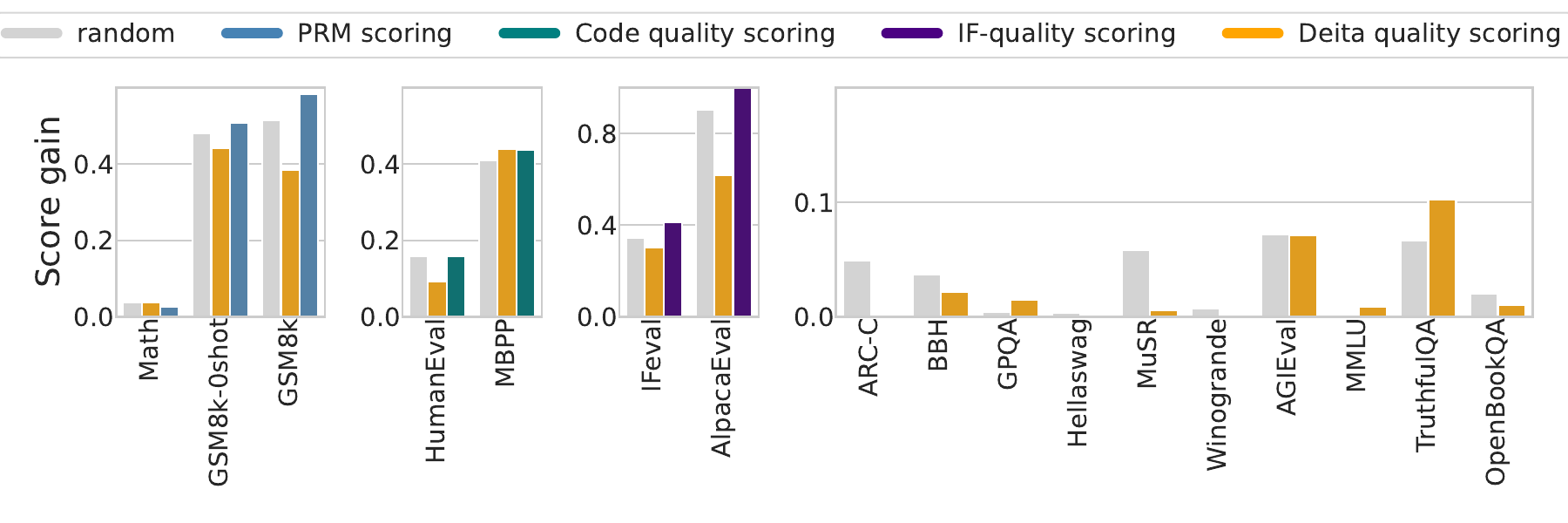}
    \caption{Performance of quality scoring-based sampling on various benchmarks.}
    \label{fig:scorer_eval_deita_quality}
\end{figure*}

\subsubsection{Effectiveness}
\label{app:quality_scorer:effectiveness}

We test the efficacy of \textsc{Laser} \emph{task-specific quality scorers} against the general-purpose \textsc{Deita} \emph{quality scorer}, by comparing the performance of Mistral-7B-v0.3 finetuned on 25k data points, either randomly selected from data in Table~\ref{tab:datasets}, or based on ranking by considered quality scorers. For each considered benchmark, we selected only samples from the relevant category as defined in Table~\ref{tab:benchmark-details}. Figure~\ref{fig:scorer_eval_deita_quality} demonstrates the clear gain of process reward model (PRM) scoring for \emph{Math} benchmarks, particularly GSM8K. For \emph{Coding} benchmarks, code-quality scoring outperforms \textsc{Deita} quality scoring on HumanEval, but achieves similar results as random. IF-quality scoring surpasses both \textsc{Deita} quality scoring and random on \emph{Generation/Brainstorming} benchmarks. For other categories (\textit{Reasoning}, \textit{Factual QA} and \textit{Extraction}), \textsc{Deita} quality scoring provides noticeable benefits only on TruthfulQA, with minimal impact on other benchmarks.

\begin{figure}[h!]
    \centering
    \includegraphics[width=1\linewidth]{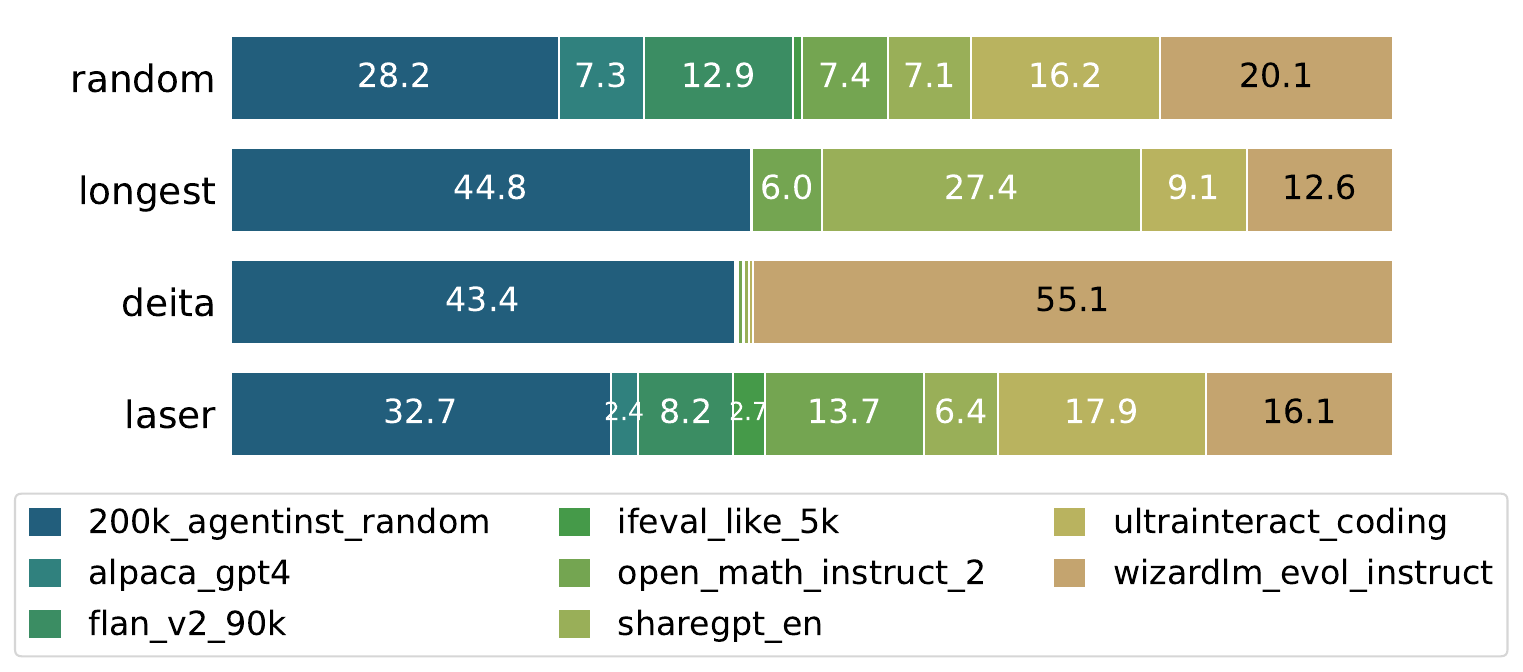}
    \caption{Source dataset proportions of different sampling strategies for 100k samples.}
    \label{fig:sampling_origin}
\end{figure}

\subsection{Sampling -- details}
\label{app:sampling}

\subsubsection{Effectiveness}
\label{app:sampling:effectiveness}

\begin{figure}[h!]
    \centering
    \includegraphics[width=1\linewidth]{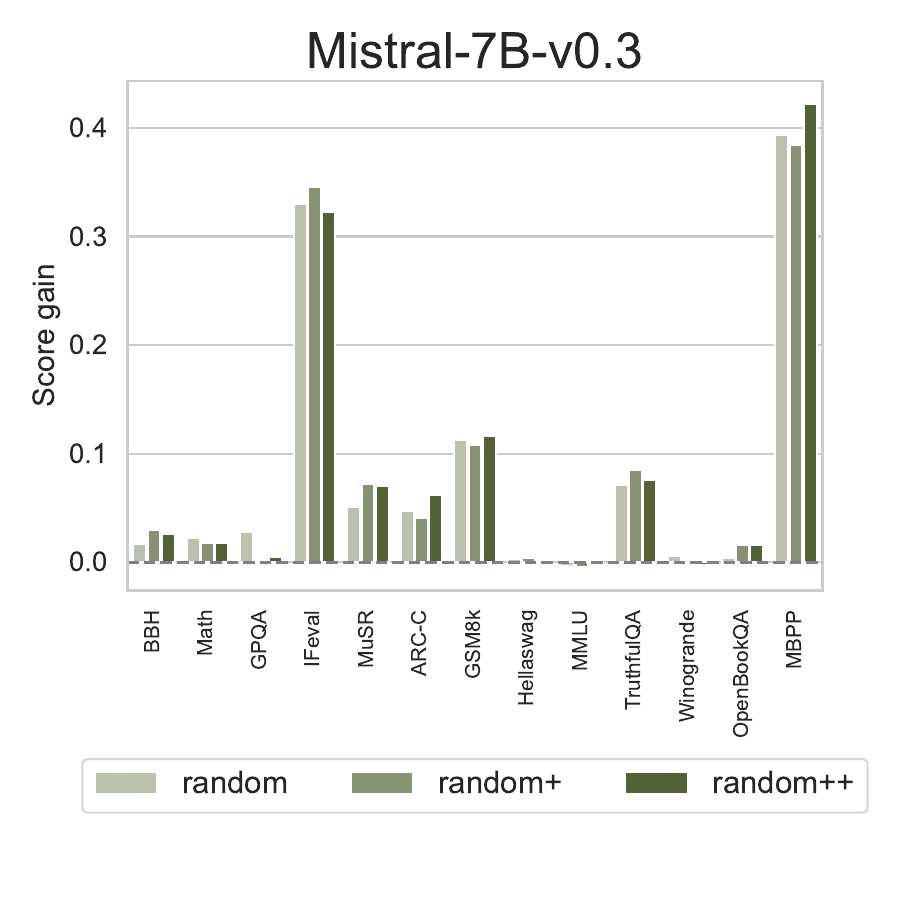}
    \caption{Effect of different sampling strategies on training outcomes. While \emph{random} selects $m$ = 25k arbitrary samples, \emph{random{\small+}} samples by stratifying the categories in $D'$ according to the proportions shown in Figure~\ref{fig:sampling_categories} (\textbf{laser}). Ultimately, \emph{random{\small++}} also includes clustering (as described in Section~\ref{subsec:Sampling}, while randomly selecting a datapoint from each cluster)}
    \label{fig:sampling_eval}
\end{figure}

In Section~\ref{subsec:Sampling}, we introduce a sampling technique that aims to improve the diversity of the target dataset $\mathcal{D'}$. 
The sampling technique consists of two components: firstly, we stratify the sampling by the categories that we assigned using the instruction classifier $\pi_c$ (according to the proportions shown in Figure~\ref{fig:sampling_categories} [\textbf{laser}]) and, secondly, we cluster the datapoints in embedding space and restrict the sampling to one datapoint per cluster.
We sample here once 25k data points with the first technique (which we will call \emph{random{\small+}}) and once 25k data points with both methods (correspondingly, \emph{random{\small++}}) from $\mathcal{D}$. 
The \emph{random} in the method name indicates that we select samples otherwise randomly (hence, not using any scoring during sampling).
We then train a Mistral-7B-v0.3-base model on the resulting datasets and compare the results with entirely random sampling in Figure~\ref{fig:sampling_eval}.

\subsubsection{Efficiency}
\label{app:sampling efficiency}

In Table~\ref{tab:runtime}, we compare the sampling runtime of \textsc{Laser} with \textsc{Deita}, which iteratively builds $\mathcal{D'}$ by adding dissimilar samples based on embedding distance, as well as \textsc{CaR}, which runs clustering on the entire dataset $\mathcal{D}$. While \textsc{Deita} is highly efficient given small $m$ (e.g., 1k), its runtime grows rapidly with larger $m$. Meanwhile, \textsc{CaR} suffers from out-of-memory issues at large $m$. In contrast, \textsc{Laser} remains stable across different $m$ and is generally more efficient than \textsc{CaR}.

\begin{table}[h]
\centering
\small
\begin{tabular}{@{}crrr@{}}
\toprule
\textbf{$m$} & \textsc{Deita} & \textsc{CaR} & \textsc{Laser} \\
\midrule
1k   &   92.3 s   &  5515.7 s  & 1505.1 s \\
10k  &  970.2 s   &  5750.9 s  & 1555.9 s \\
100k & 41055.3 s  & OOM        & 2818.1 s \\
\bottomrule
\end{tabular}
\caption{Runtime comparison for different $m$.}
\label{tab:runtime}
\end{table}

\subsection{Datasets -- details}
\label{app:datasets}

\subsubsection{Datasets main experiments}
\label{app:datasets:source}
\begin{figure*}[th!]
    \centering
    \includegraphics[width=1\linewidth]{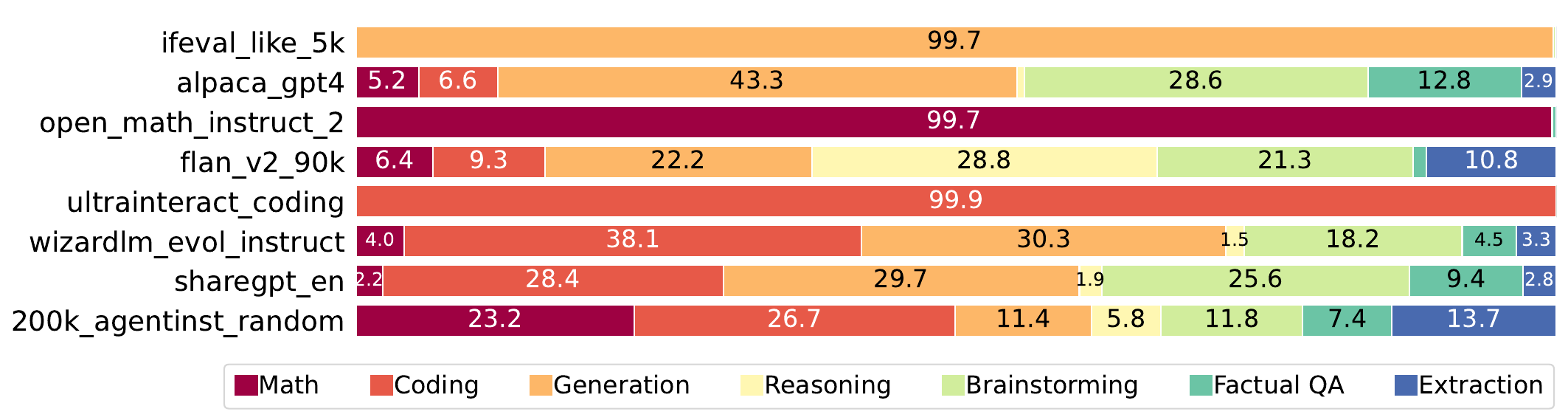}
    \caption{Category distribution in IT datasets used in the experiments.}
    \label{fig:origin_proportion}
\end{figure*}

Figure~\ref{fig:origin_proportion} shows SetFit classification results for IT datasets from which we composed $\mathcal{D}$.
On the other hand, Figure~\ref{fig:sampling_origin} shows which of the source datasets are sampled when we use various sampling strategies. 
Interestingly, while \textsc{Laser} maintains relative diversity across different source datasets, \textsc{Deita} is completely dominated by only two datasets: microsoft/orca-agentinstruct-1M-v1 (200k sampled) and WizardLMTeam/WizardLM\_evol\_instruct\_V2\_196K.

\subsubsection{Datasets robustness experiments}
\label{app:datasets:robustness}

\paragraph{Dataset details $\mathcal{D}_{\text{strong}}$ and $\mathcal{D}_{\text{weak}}$}
In Section~\ref{subsubsec:results_robustness}, we introduce $\mathcal{D}_{\text{strong}}$ and $\mathcal{D}_{\text{weak}}$.
While $\mathcal{D}_{\text{strong}}$ simply consists of the Tülu v3 dataset, $\mathcal{D}_{\text{weak}}$ is a aggregation of different datasets.
The single components exhibited either in our own pilot experiments or in prior related work comparably weak performance to other datasets. 
This weaker performance also shows, when we compare the finetuning using full $\mathcal{D}_{\text{weak}}$ with the finetuning using full $\mathcal{D}_{\text{strong}}$ in Figure~\ref{fig:eval_high_vs_lq_source}, with the latter significantly outperforming the former.

\begin{table}[!t] 
\centering 
\scriptsize
\begin{adjustbox}{width=0.48\textwidth}
\begin{tabular}{@{}l@{}rr@{}}
        \toprule
        \textbf{Dataset} & \textbf{\#samples} \\
         & (\emph{\#turns}) \\
        \midrule
        ConiferLM/Conifer \tinycitep{coniferlm} & 13k\\
        databricks/databricks-dolly-15k \tinycitep{DatabricksBlog2023DollyV2} & 15k \\
        akoksal/LongForm \tinycitep{koksal2023longform} & 23k\\
        alpaca \tinycitep{taori2023stanford} & 52K \\ 
        vicgalle/alpaca-gpt4 & 52K \\
        ai2-adapt-dev/flan\_v2\_converted \tinycitep{sanh2022multitask} & 90K \\
        nvidia/Daring-Anteater \tinycitep{wang2024helpsteer2} & 100k  \\
        AI-MO/NuminaMath-CoT\footnote{random subset} \tinycitep{numina_math_datasets} & 250k \\
        \arrayrulecolor{black!30}\midrule
        \emph{all} & 595K  \\ 
        \arrayrulecolor{black}\bottomrule
\end{tabular}
\end{adjustbox}
\caption{Composition of $\mathcal{D}_{\text{weak}}$}
\label{tab:low-quality-data-composition}
\end{table}

\begin{figure}[h!]
    \centering
    \begin{subfigure}[b]{0.99\linewidth}
        \includegraphics[width=\linewidth]{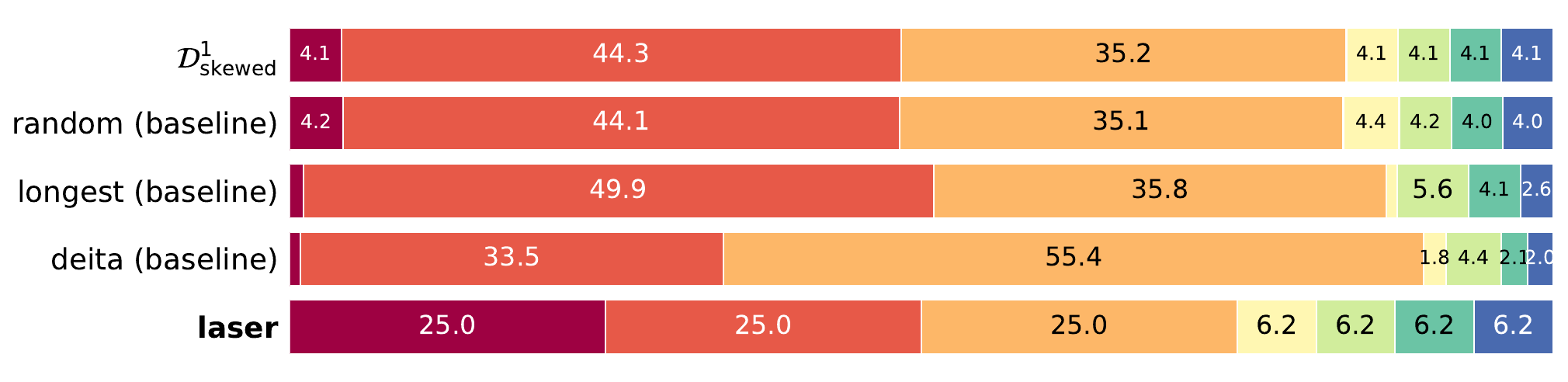}
    \end{subfigure}
    \begin{subfigure}[b]{0.99\linewidth}
        \includegraphics[width=\linewidth]{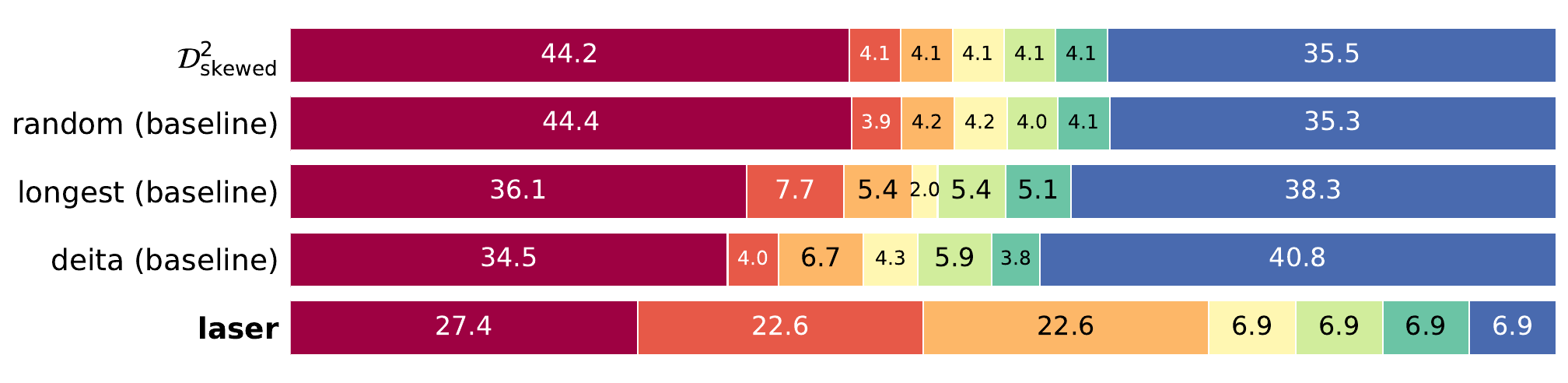}
    \end{subfigure}
    \begin{subfigure}[b]{0.99\linewidth}
        \includegraphics[width=\linewidth]{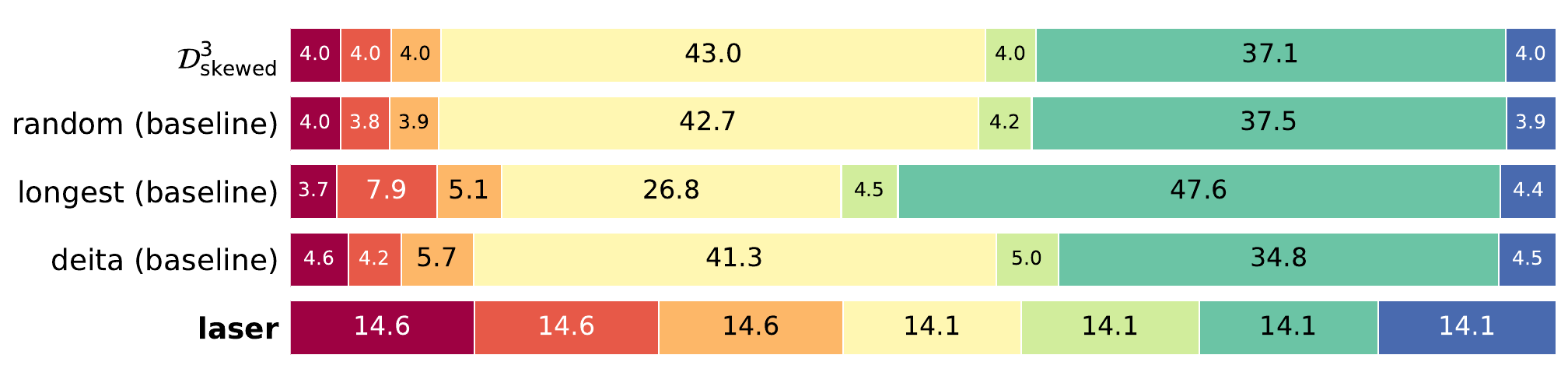}
    \end{subfigure}
    \begin{subfigure}[b]{0.99\linewidth}
        \includegraphics[width=\linewidth]{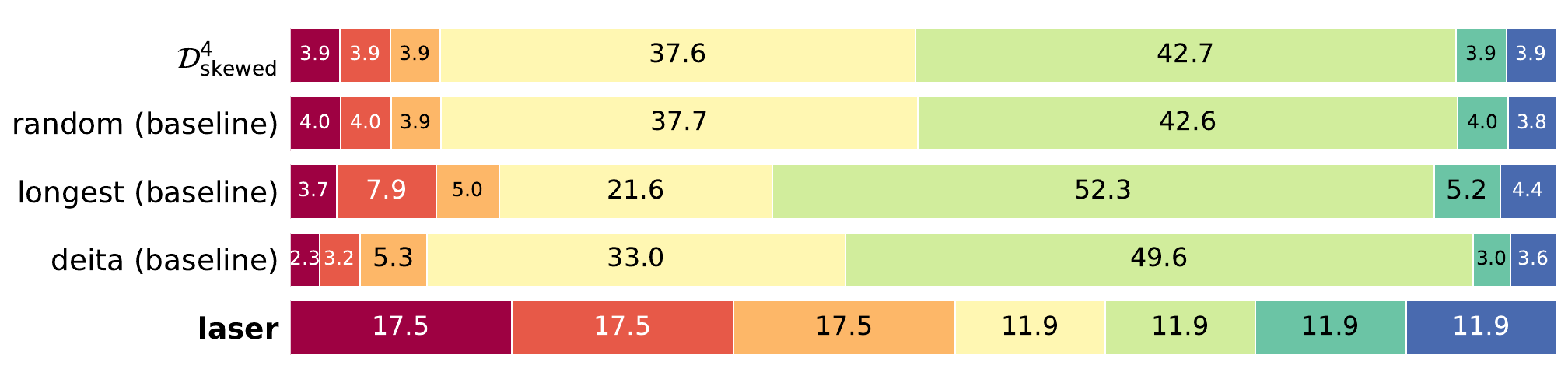}
    \end{subfigure}
    \begin{subfigure}[b]{0.99\linewidth}
        \includegraphics[width=\linewidth]{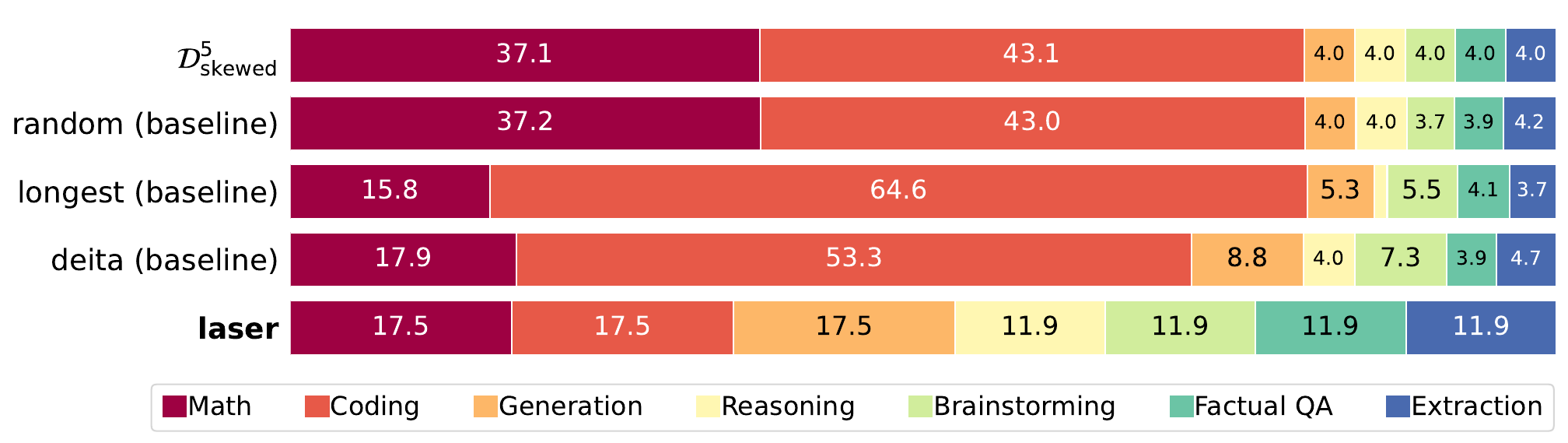}
    \end{subfigure}
    \caption{Category distribution for datasets sampled from the 5 bias source datasets $\mathcal{D}_{\text{skewed}}$ using different sampling strategies.}
    \label{fig:porportions_skewed_sampled}
\end{figure}

\paragraph{Results of experiments with $\mathcal{D}_{\text{skewed}}$}
In Section~\ref{subsubsec:results_robustness}, we experiment with skewed data distributions. In Figure~\ref{fig:porportions_skewed_sampled}, we show the resulting category distributions when sampling with different sampling approaches from the 5 skewed source distributions. 
As we can see, only \textsc{Laser} reduces the introduced bias in the skewed source data, while \textsc{Deita} is not effective in reducing the bias, but in many cases exacerbates it.

\subsection{Finetuning -- details}
\label{app:finetuning}

We finetune all models in our experiments on one node of 4 x NVIDIA H100-SXM5 Tensor Core-GPUs (94 GB HBM2e). Table~\ref{tab:finetuning-hyperparameters} details the hyperparameters used for finetuning. 

\begin{table*}[!t]
\centering
\small
\begin{tabular}{@{}lccccc@{}}
\toprule
\textbf{Hyperparameter} & \textbf{falcon-10B} & \textbf{llama-8B} & \textbf{mistral-7B} & \textbf{qwen-3B} & \textbf{smollm-1.7B} \\
\midrule
batch\_size & 4 & 8 & 8 & 8 & 32 \\ 
gradient\_accumulation & 16 & 8 & 16 & 8 & 16 \\ \arrayrulecolor{black!30}\midrule
learning\_rate & 2.0e-05 & 5.0e-06 & 5.0e-06 & 2.0e-05 & 1.0e-04 \\ \arrayrulecolor{black!30}\midrule
num\_train\_epochs & 2 & 2 & 2 & 2 & 3 \\
weight\_decay & 0.1 & 0.01 & 0.01 & 0.01 & 0.01 \\
\arrayrulecolor{black!30}\midrule
warmup\_ratio & 0.03 & 0.03 & 0.1 & 0.03 & 0.03 \\
lr\_scheduler\_type & \multicolumn{5}{c}{cosine} \\
attn\_implementation & \multicolumn{5}{c}{flash\_attention\_2} \\
\arrayrulecolor{black!30}\midrule
max\_seq\_length & \multicolumn{5}{c}{2048} \\
neftune\_noise\_alpha & \multicolumn{5}{c}{5} \\
use\_liger & \multicolumn{5}{c}{True} \\
\arrayrulecolor{black}\bottomrule
\end{tabular}
\caption{Hyperparameter details for finetuning.}
\label{tab:finetuning-hyperparameters}
\end{table*}


\subsection{Results -- details}
\label{app:results_details}
\begin{figure*}[h!]
    \centering
    \begin{subfigure}[b]{0.8\linewidth}
        \includegraphics[width=\linewidth]{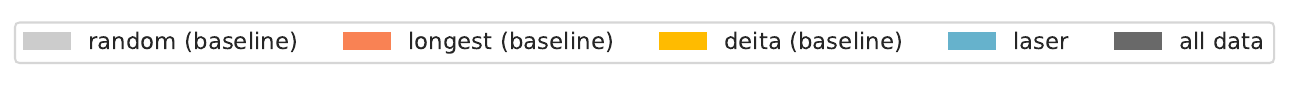}
    \end{subfigure}
    
    \begin{subfigure}[b]{0.3\linewidth}
        \includegraphics[width=\linewidth]{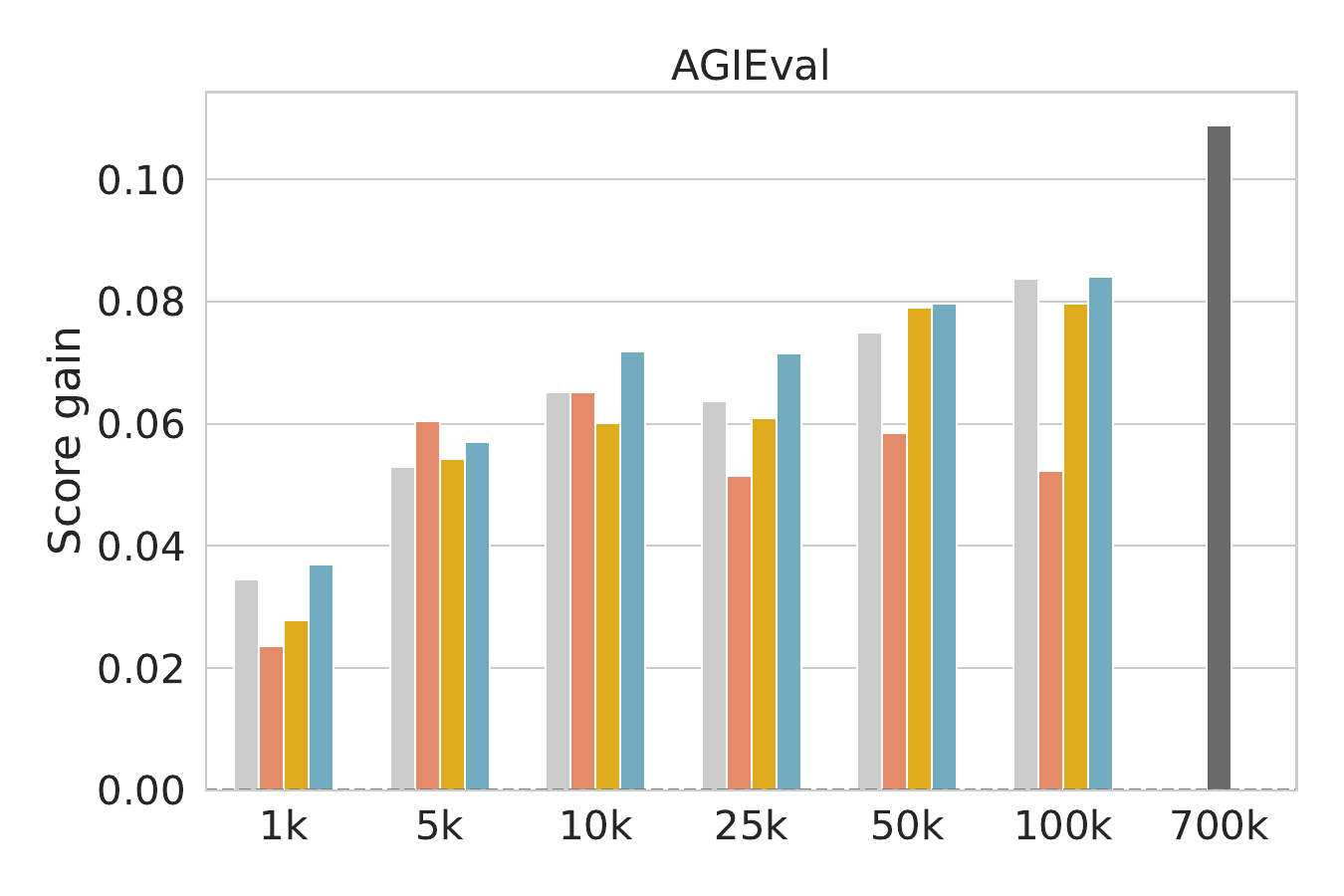}
        \caption{AGIEval \citep{zhong-etal-2024-agieval}}
        \label{fig:eval_scaling_agieval}
    \end{subfigure}
    \begin{subfigure}[b]{0.3\linewidth}
        \includegraphics[width=\linewidth]{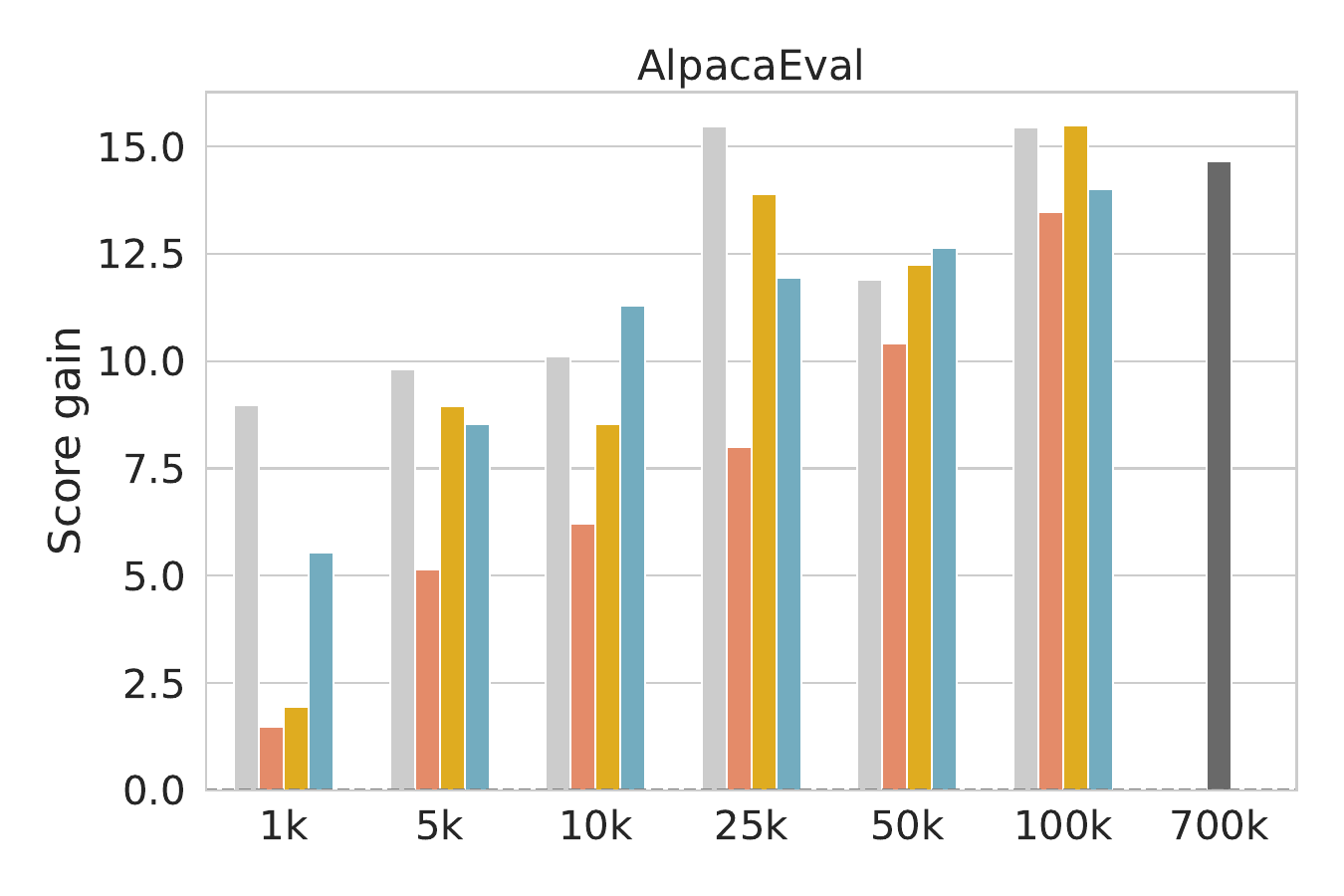}
        \caption{AlpacaEval \citep{dubois2024lengthcontrolled}}
        \label{fig:eval_scaling_alpacaeval}
    \end{subfigure}
    \begin{subfigure}[b]{0.3\linewidth}
        \includegraphics[width=\linewidth]{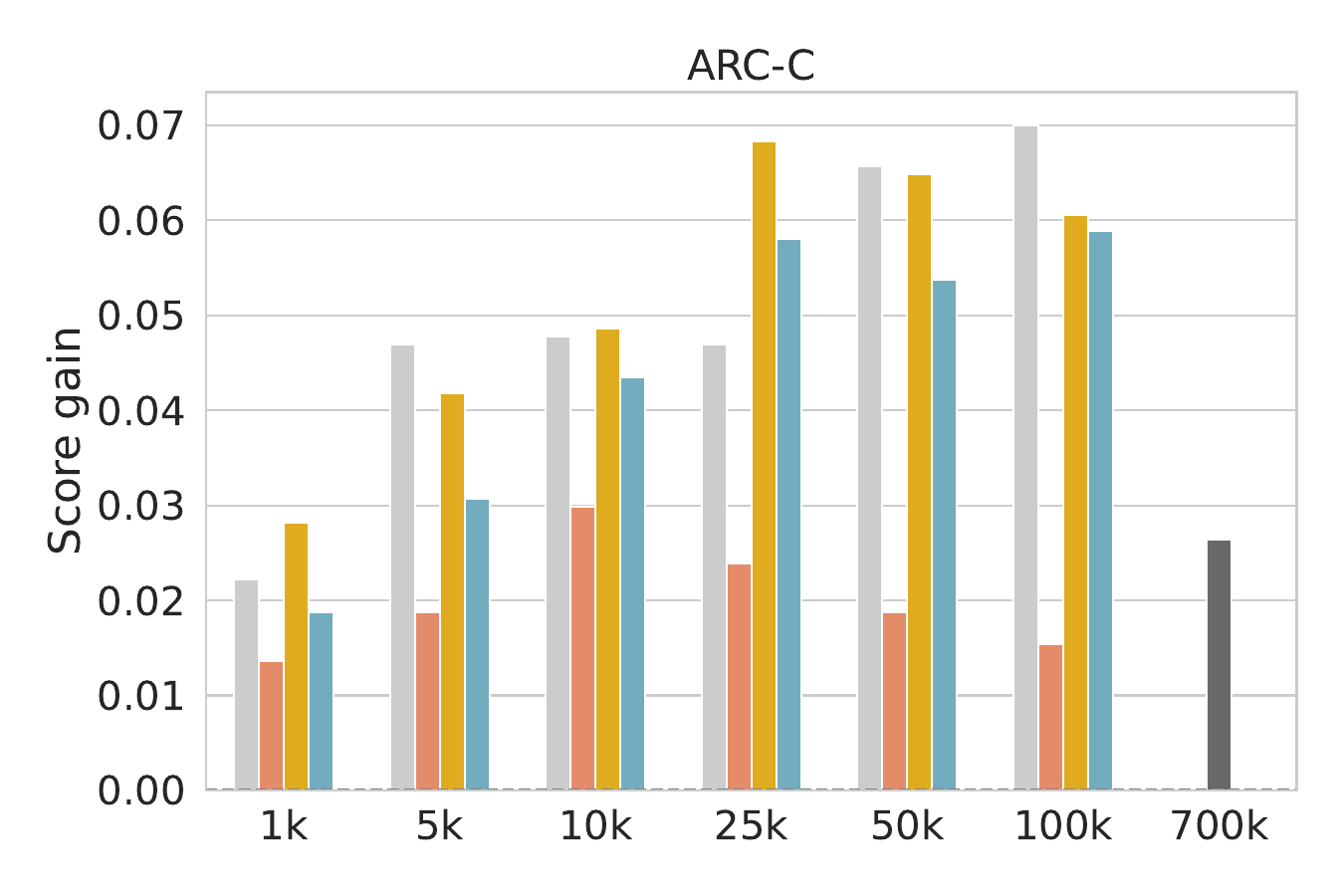}
        \caption{ARC-Challenge \citep{allenai_arc}}
        \label{fig:eval_scaling_arc}
    \end{subfigure}

    \begin{subfigure}[b]{0.3\linewidth}
        \includegraphics[width=\linewidth]{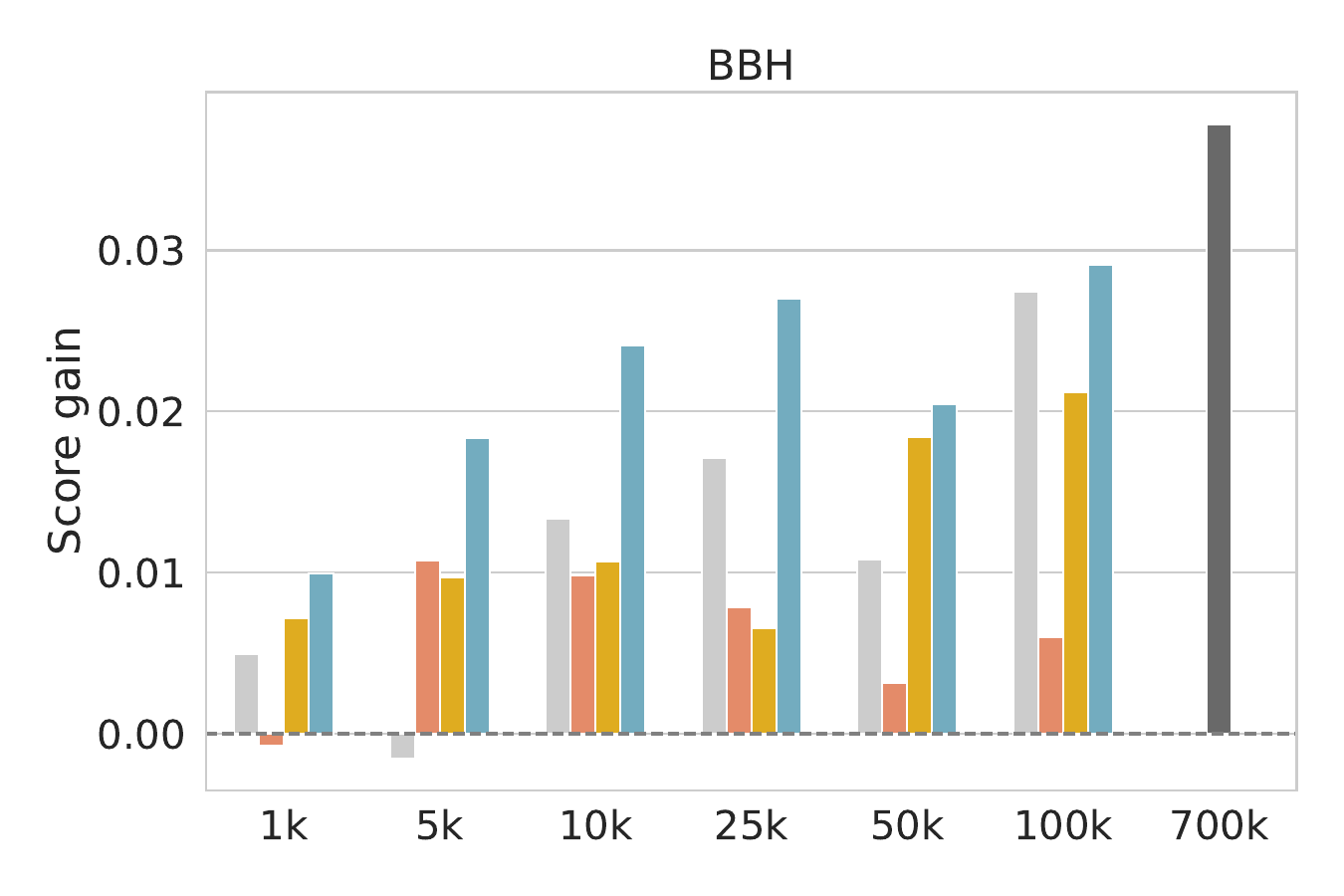}
        \caption{BBH \citep{suzgun-etal-2023-challenging}}
        \label{fig:eval_scaling_bbh}
    \end{subfigure}
    \begin{subfigure}[b]{0.3\linewidth}
        \includegraphics[width=\linewidth]{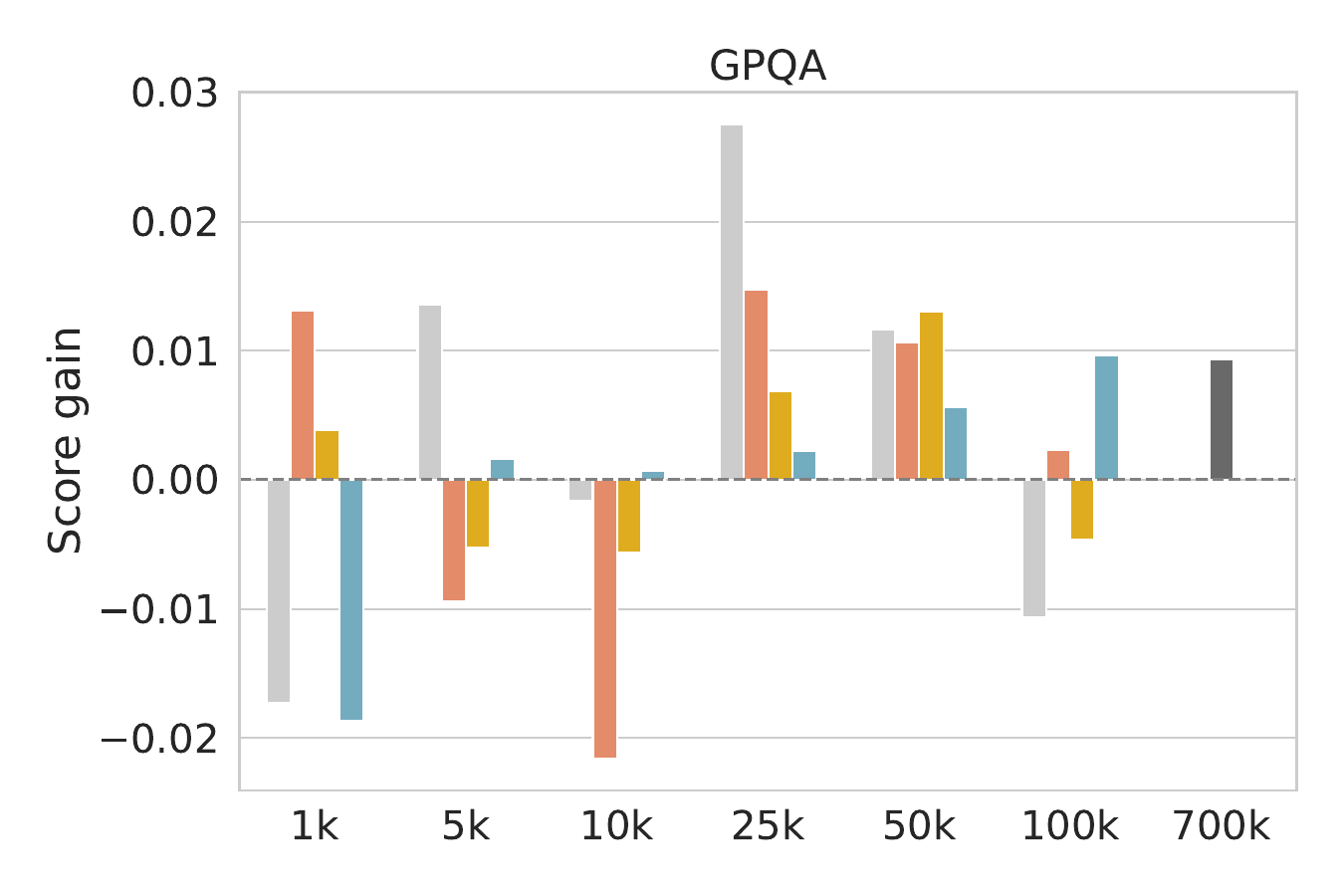}
        \caption{GPQA \citep{rein2024gpqa}}
        \label{fig:eval_scaling_gpqa}
    \end{subfigure}
    \begin{subfigure}[b]{0.3\linewidth}
        \includegraphics[width=\linewidth]{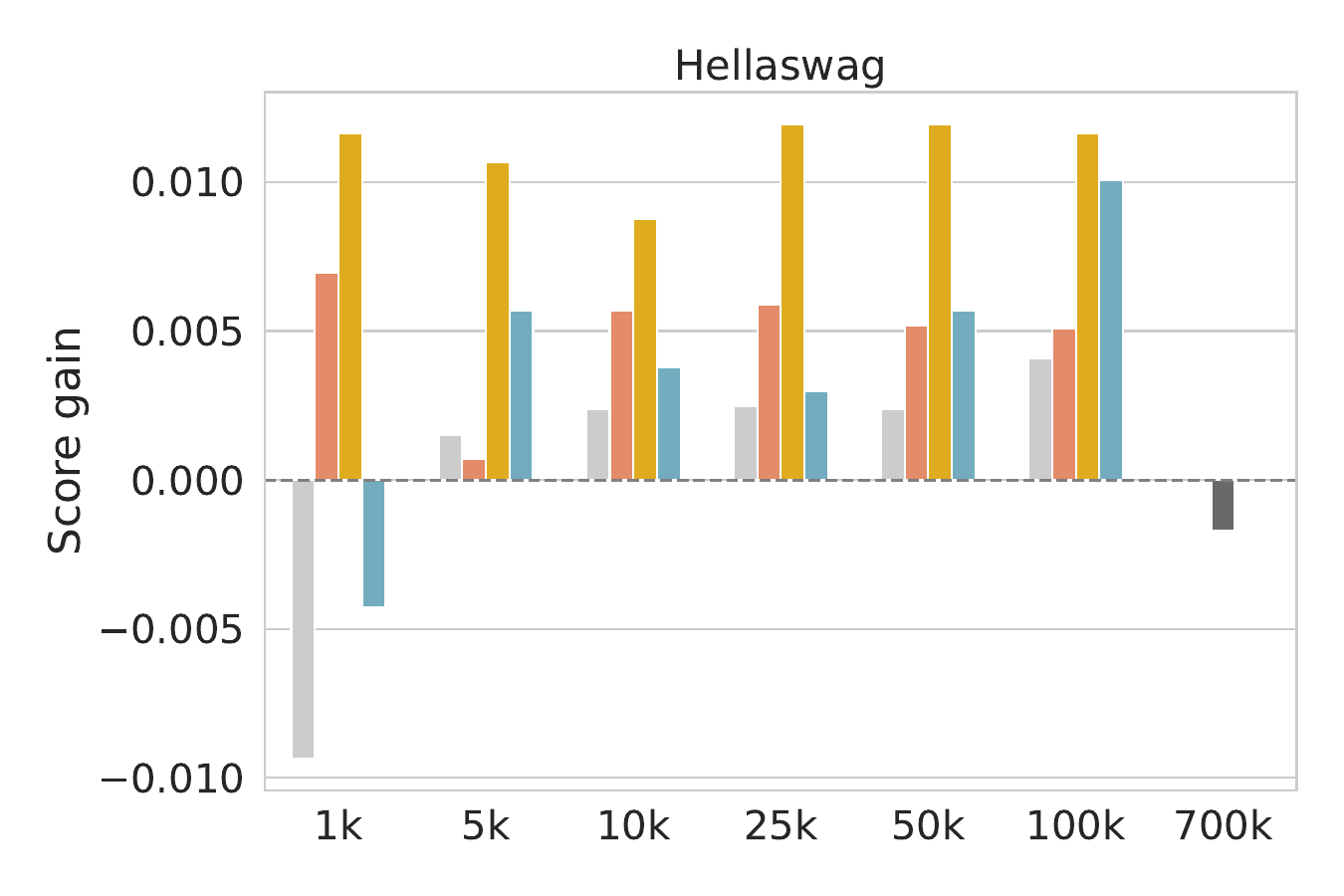}
        \caption{HellaSwag \citep{zellers2019hellaswag}}
        \label{fig:eval_scaling_hellaswag}
    \end{subfigure}
    
    \begin{subfigure}[b]{0.3\linewidth}
        \includegraphics[width=\linewidth]{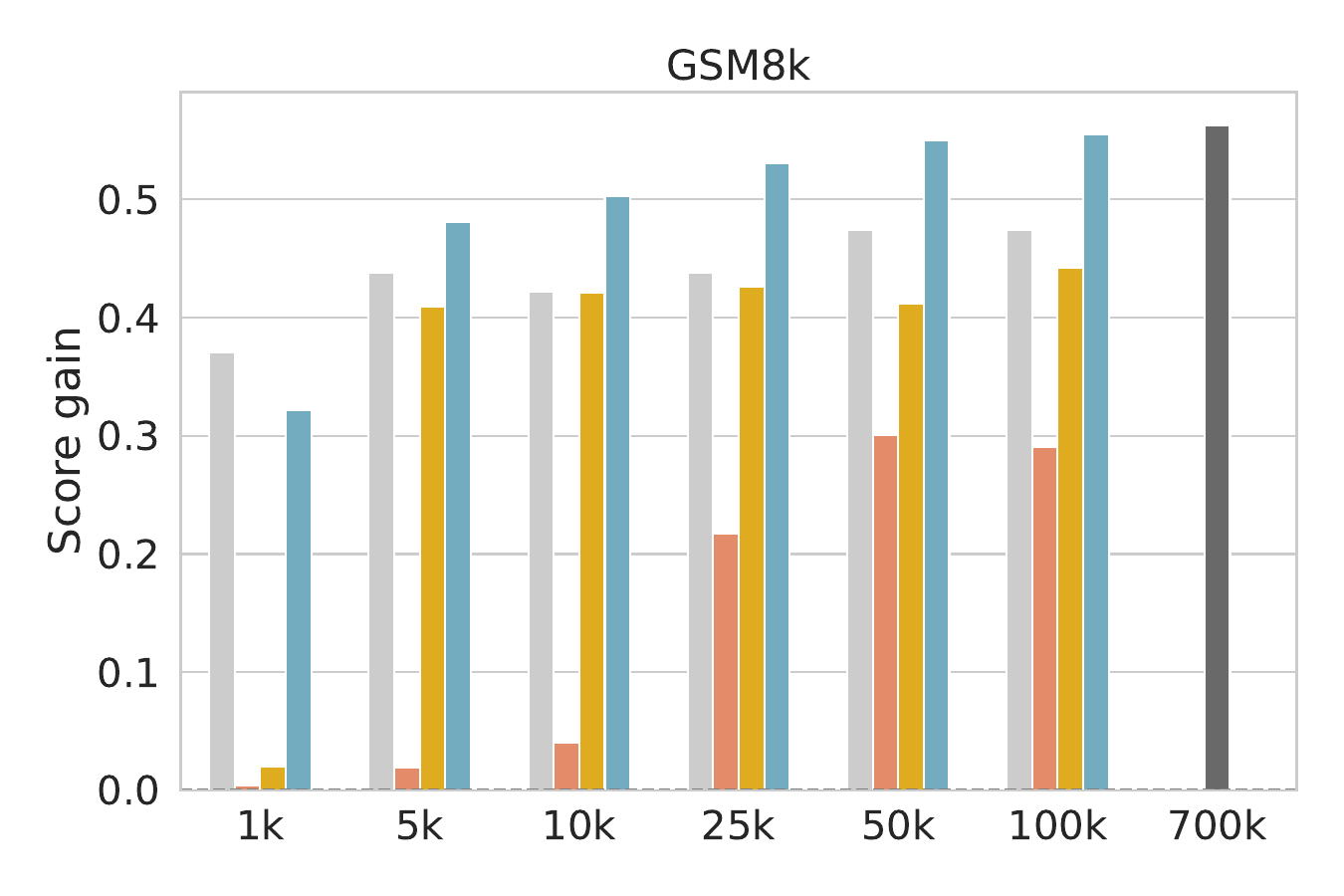}
        \caption{GSM8k$_{\textrm{cot}}$ \citep{cobbe2021gsm8k}}
        \label{fig:eval_scaling_gsm8k_cot}
    \end{subfigure}
    \begin{subfigure}[b]{0.3\linewidth}
        \includegraphics[width=\linewidth]{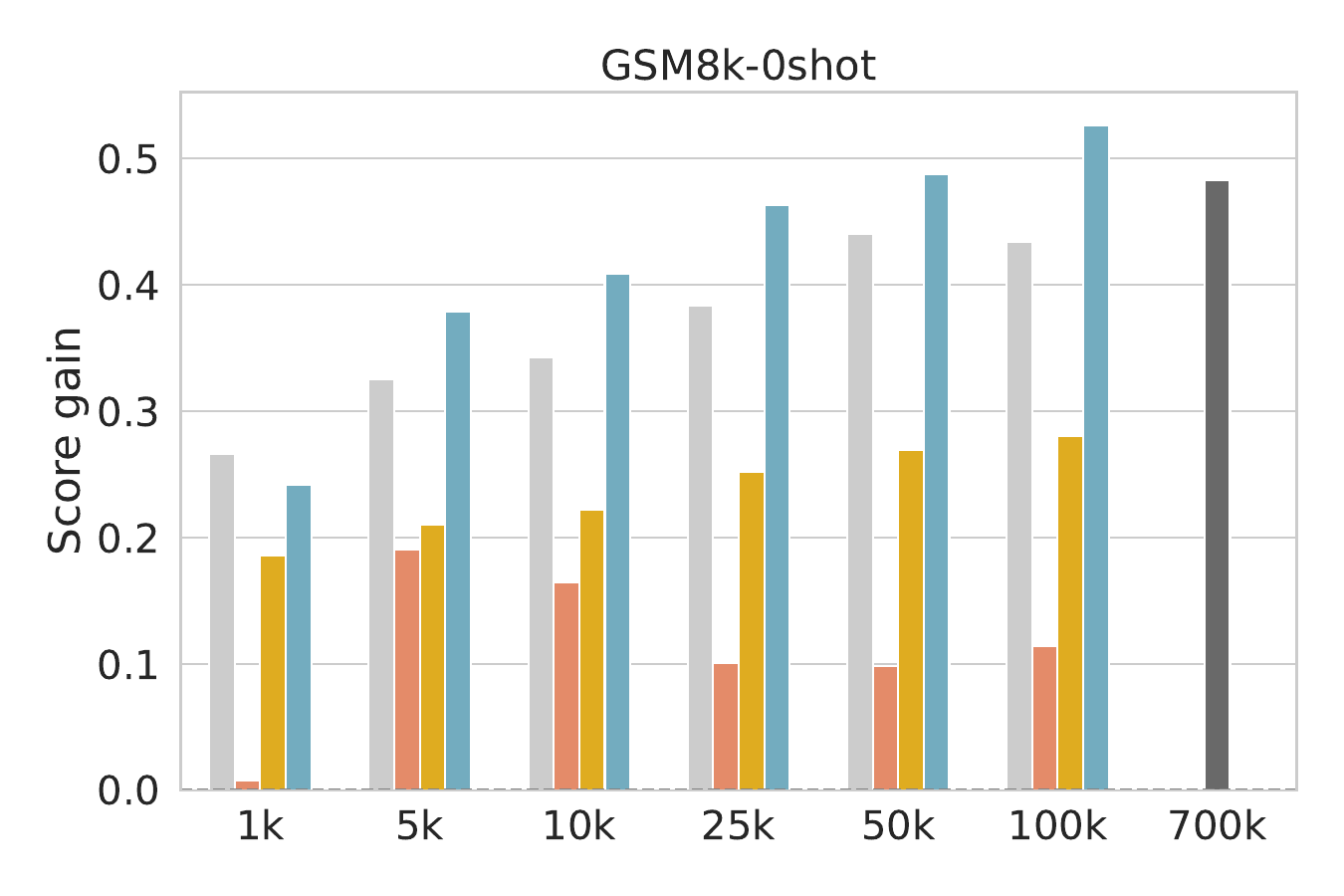}
        \caption{GSM8k$_{\textrm{cot-0-shot}}$ \citep{cobbe2021gsm8k}}
        \label{fig:eval_scaling_gsm8k_cot_0_shot}
    \end{subfigure}
    \begin{subfigure}[b]{0.3\linewidth}
        \includegraphics[width=\linewidth]{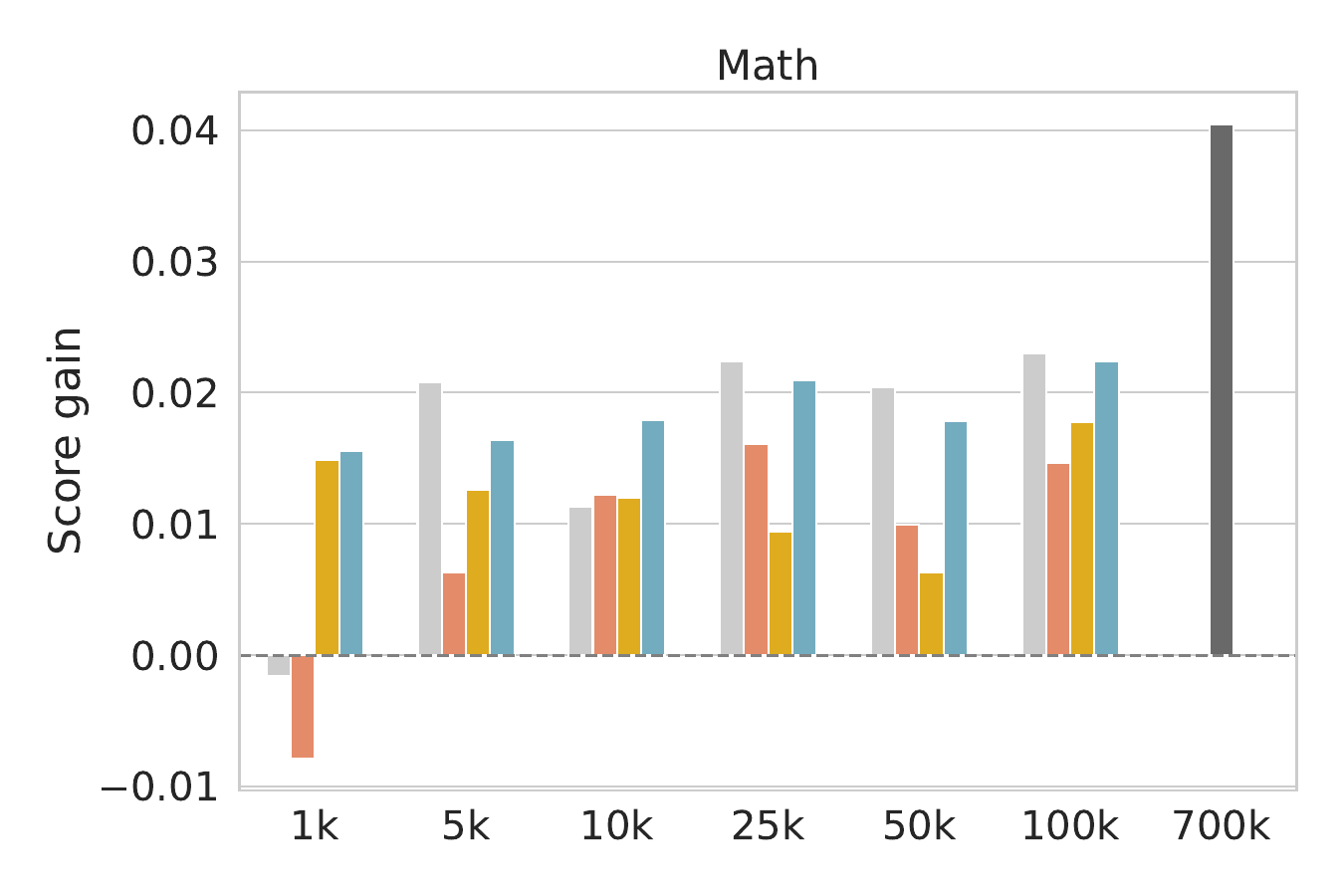}
        \caption{Math \citep{hendrycksmath2021}}
        \label{fig:eval_scaling_math}
    \end{subfigure}
    
    \begin{subfigure}[b]{0.3\linewidth}
        \includegraphics[width=\linewidth]{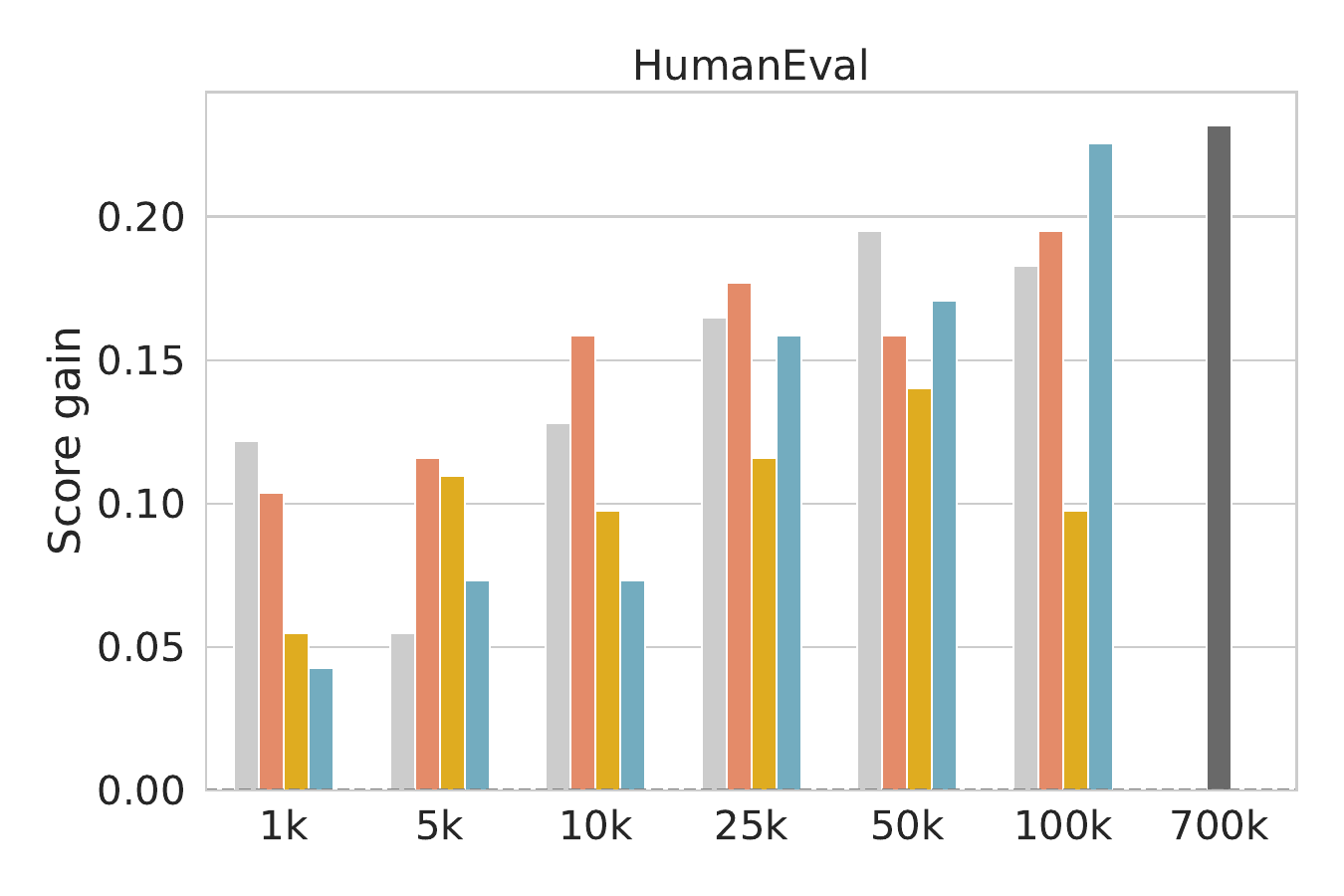}
        \caption{HumanEval \citep{chen2021codex}}
        \label{fig:eval_scaling_humaneval}
    \end{subfigure}
    \begin{subfigure}[b]{0.3\linewidth}
        \includegraphics[width=\linewidth]{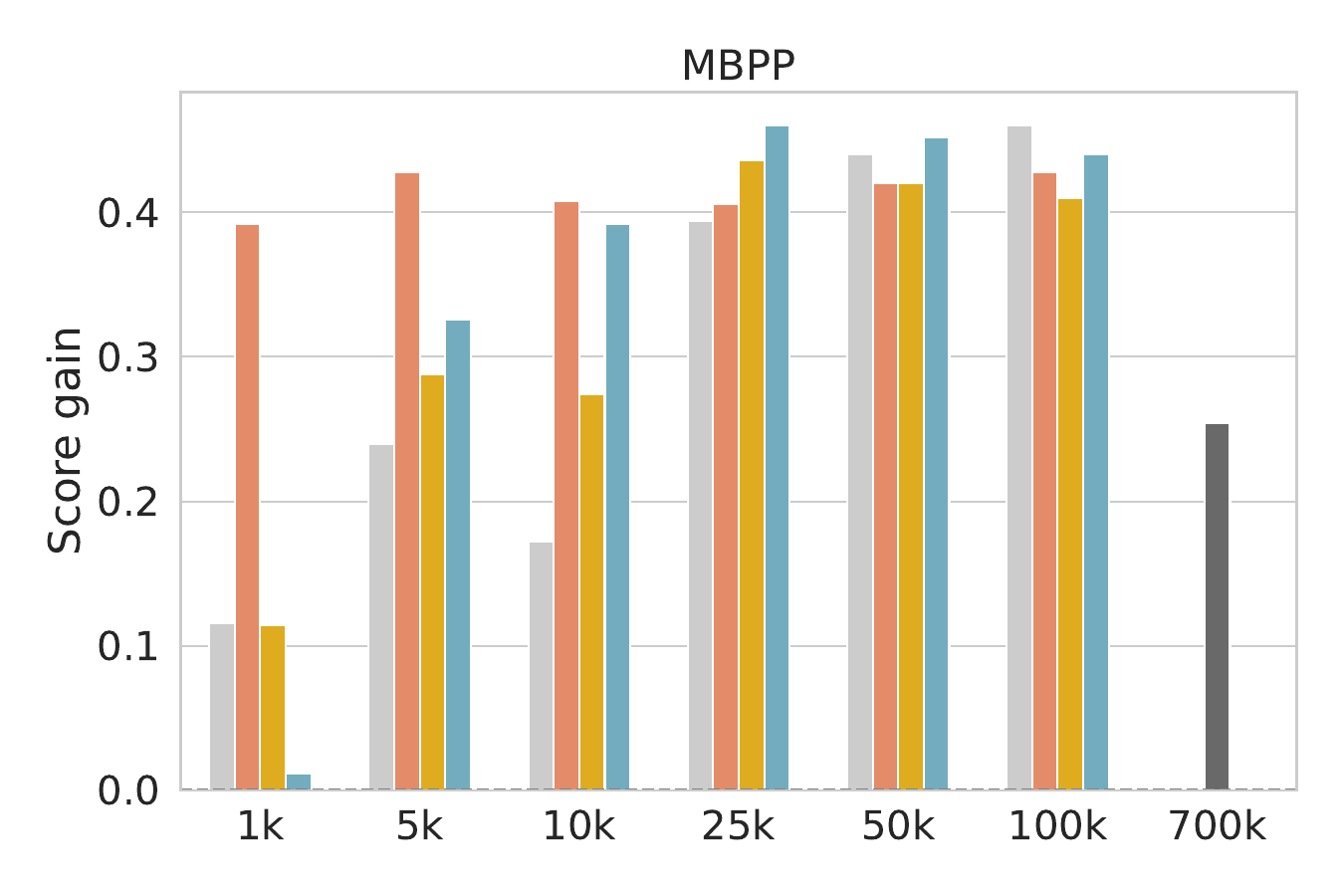}
        \caption{MBPP \citep{austin2021program}}
        \label{fig:eval_scaling_mbpp}
    \end{subfigure}
    \begin{subfigure}[b]{0.3\linewidth}
        \includegraphics[width=\linewidth]{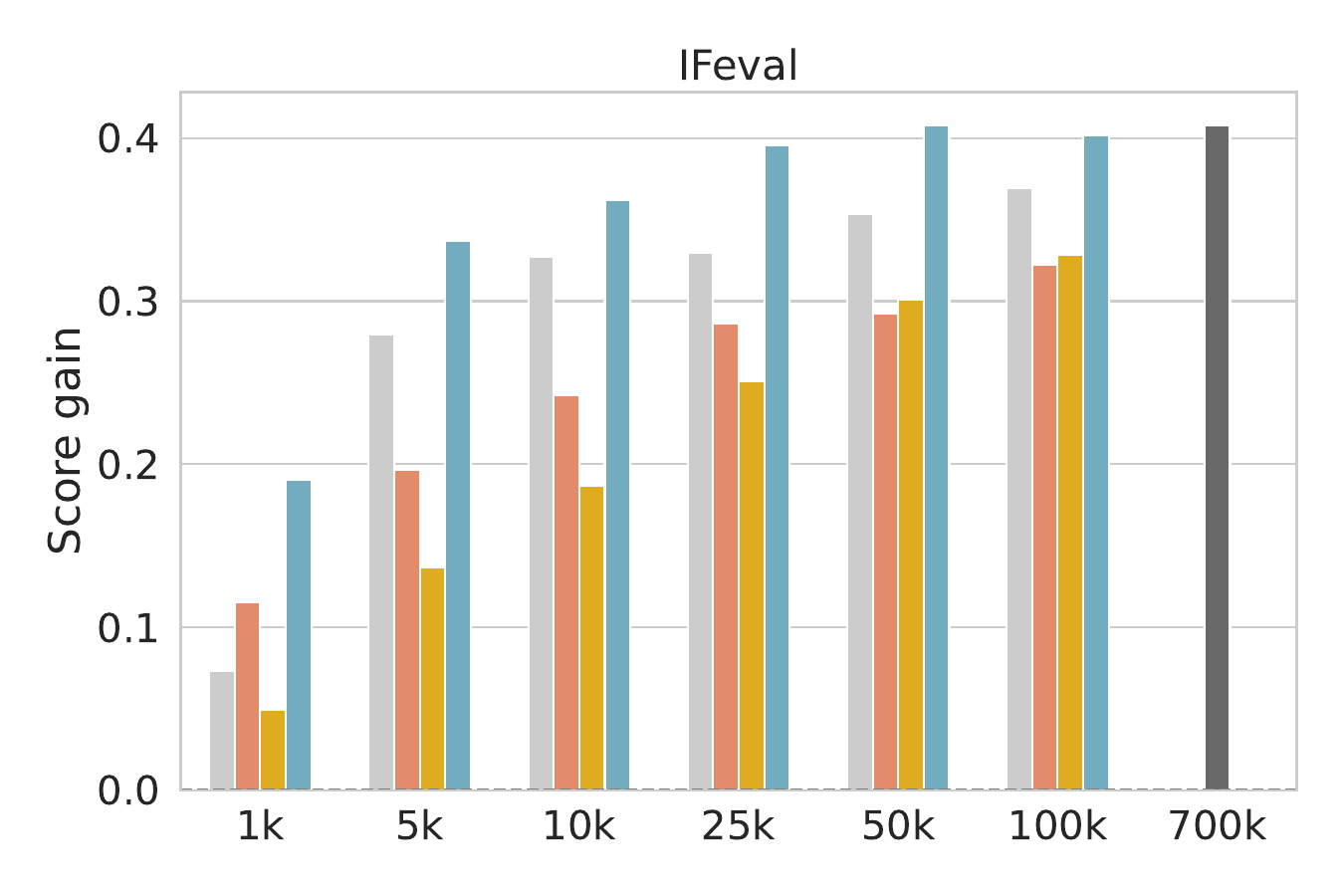}
        \caption{IFeval \citep{zhou2023instructionfollowingevaluationlargelanguage}}
        \label{fig:eval_scaling_ifeval}
    \end{subfigure}

    \begin{subfigure}[b]{0.32\linewidth}
        \includegraphics[width=\linewidth]{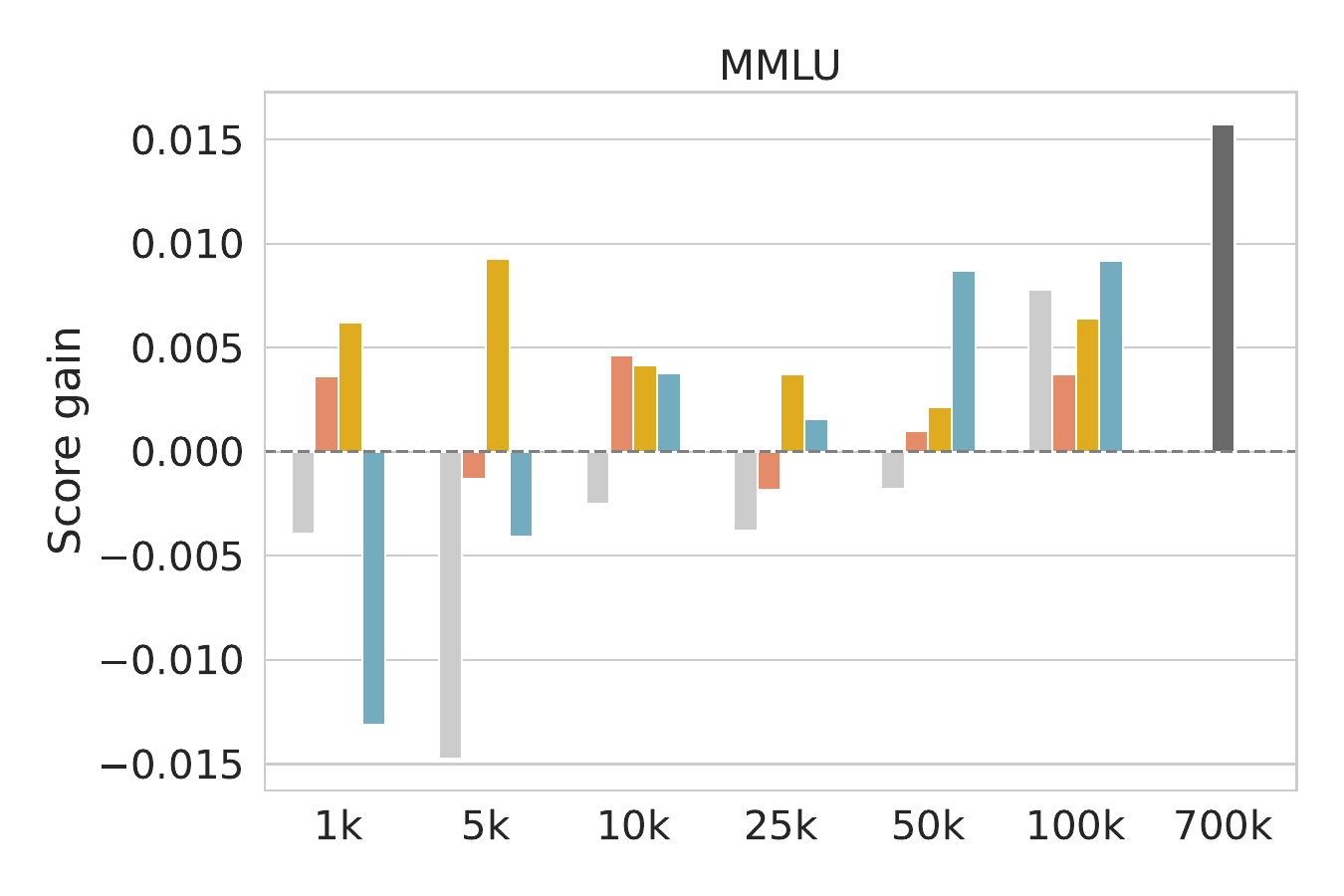}
        \caption{MMLU \citep{hendryckstest2021}}
        \label{fig:eval_scaling_mmlu}
    \end{subfigure}
    \begin{subfigure}[b]{0.32\linewidth}
        \includegraphics[width=\linewidth]{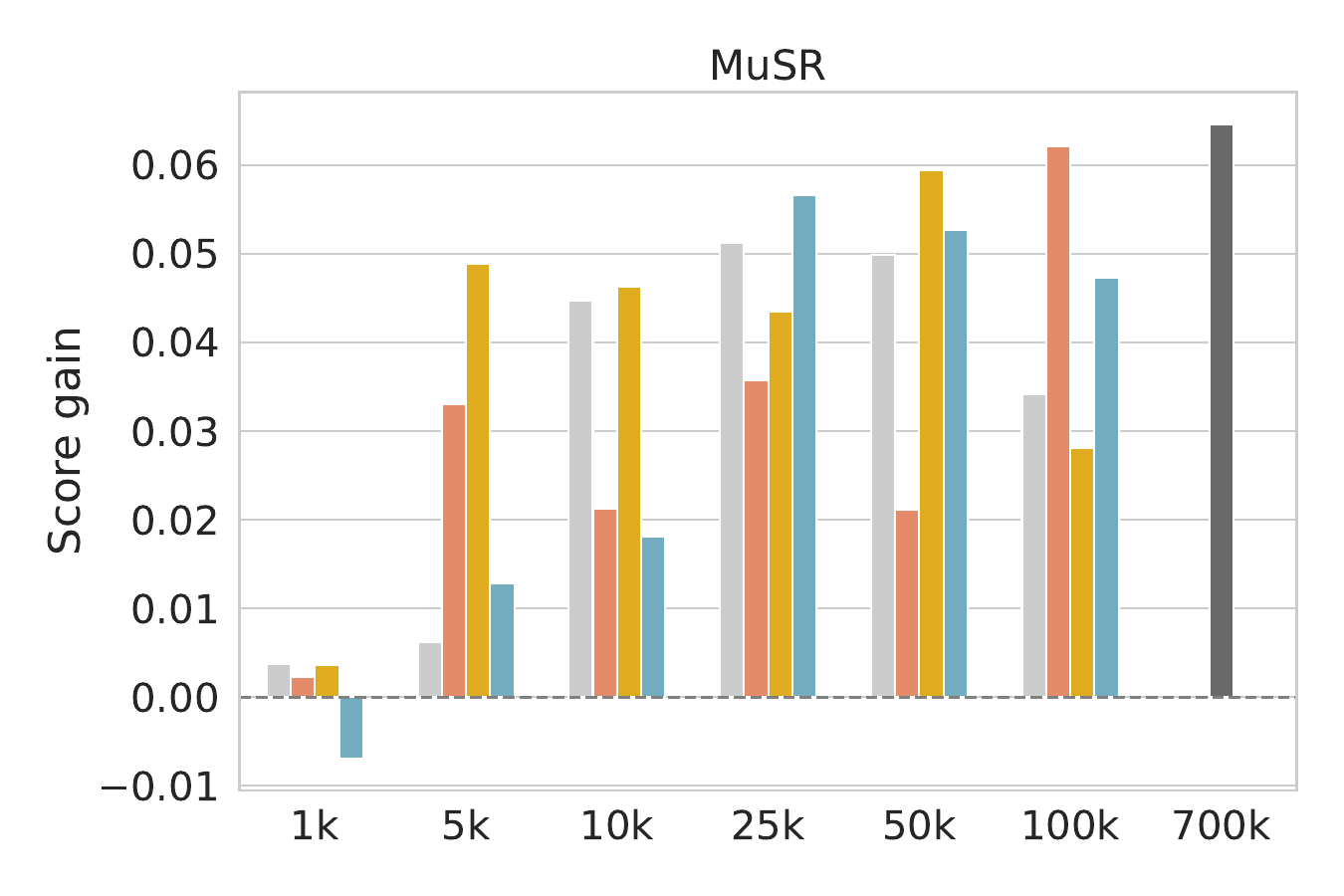}
        \caption{MuSR \citep{sprague2024musr}}
        \label{fig:eval_scaling_musr}
    \end{subfigure}
    \begin{subfigure}[b]{0.32\linewidth}
        \includegraphics[width=\linewidth]{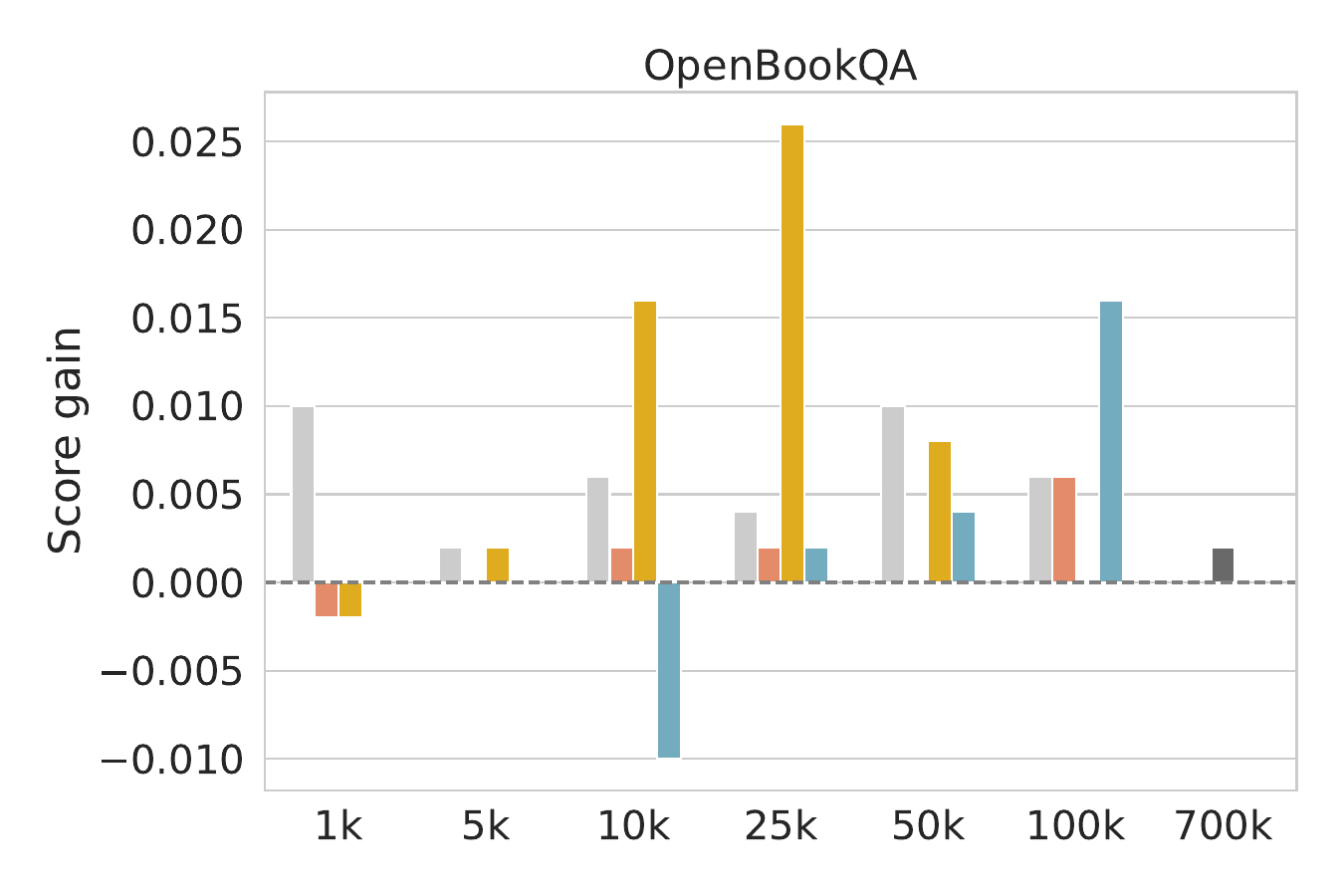}
        \caption{OpenBookQA \citep{mihaylov-etal-2018-suit}}
        \label{fig:eval_scaling_openbookqa}
    \end{subfigure}

    \begin{subfigure}[b]{0.32\linewidth}
        \includegraphics[width=\linewidth]{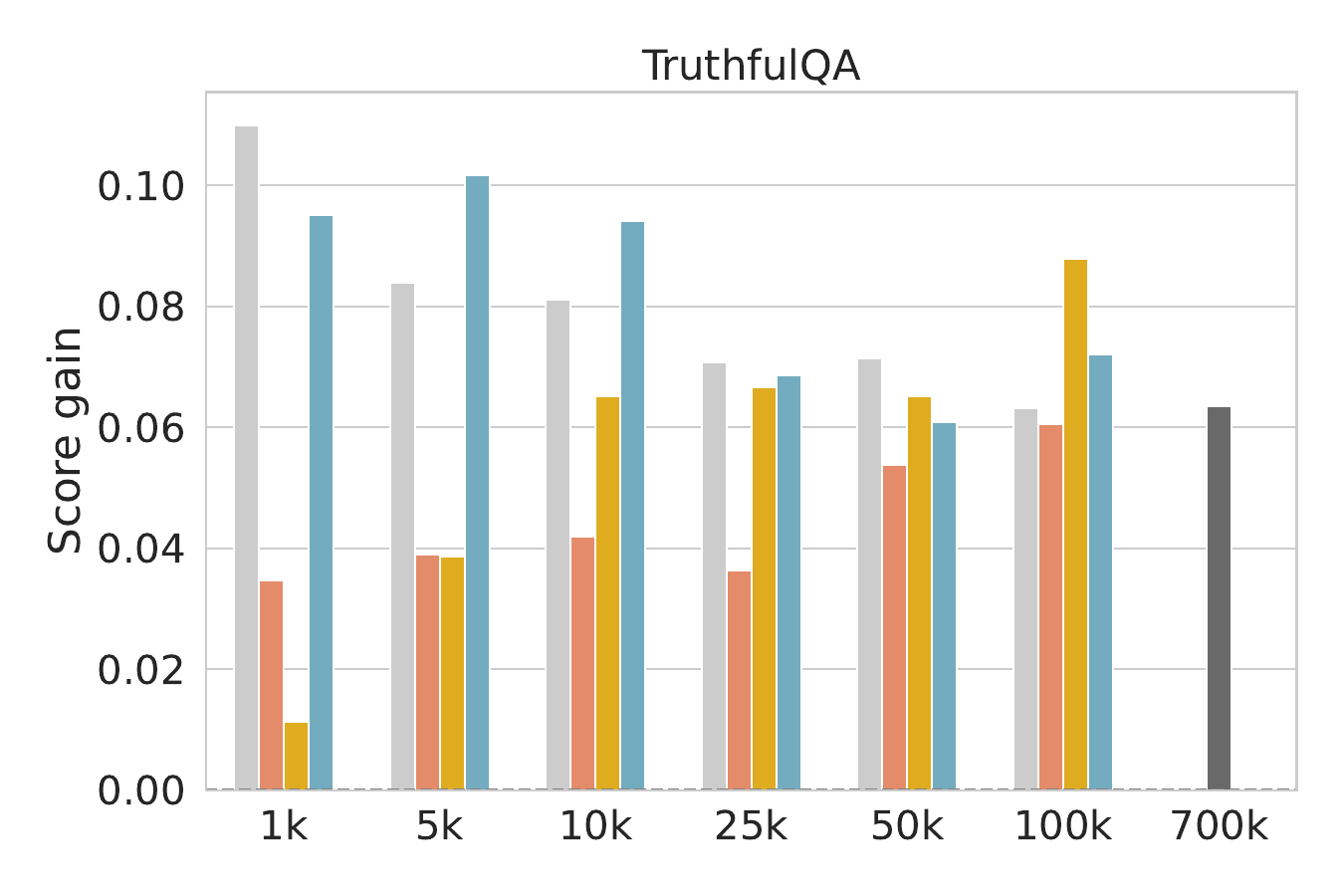}
        \caption{TruthfulQA \citep{lin-etal-2022-truthfulqa}}
        \label{fig:eval_scaling_truthfulqa}
    \end{subfigure}
    \begin{subfigure}[b]{0.32\linewidth}
        \includegraphics[width=\linewidth]{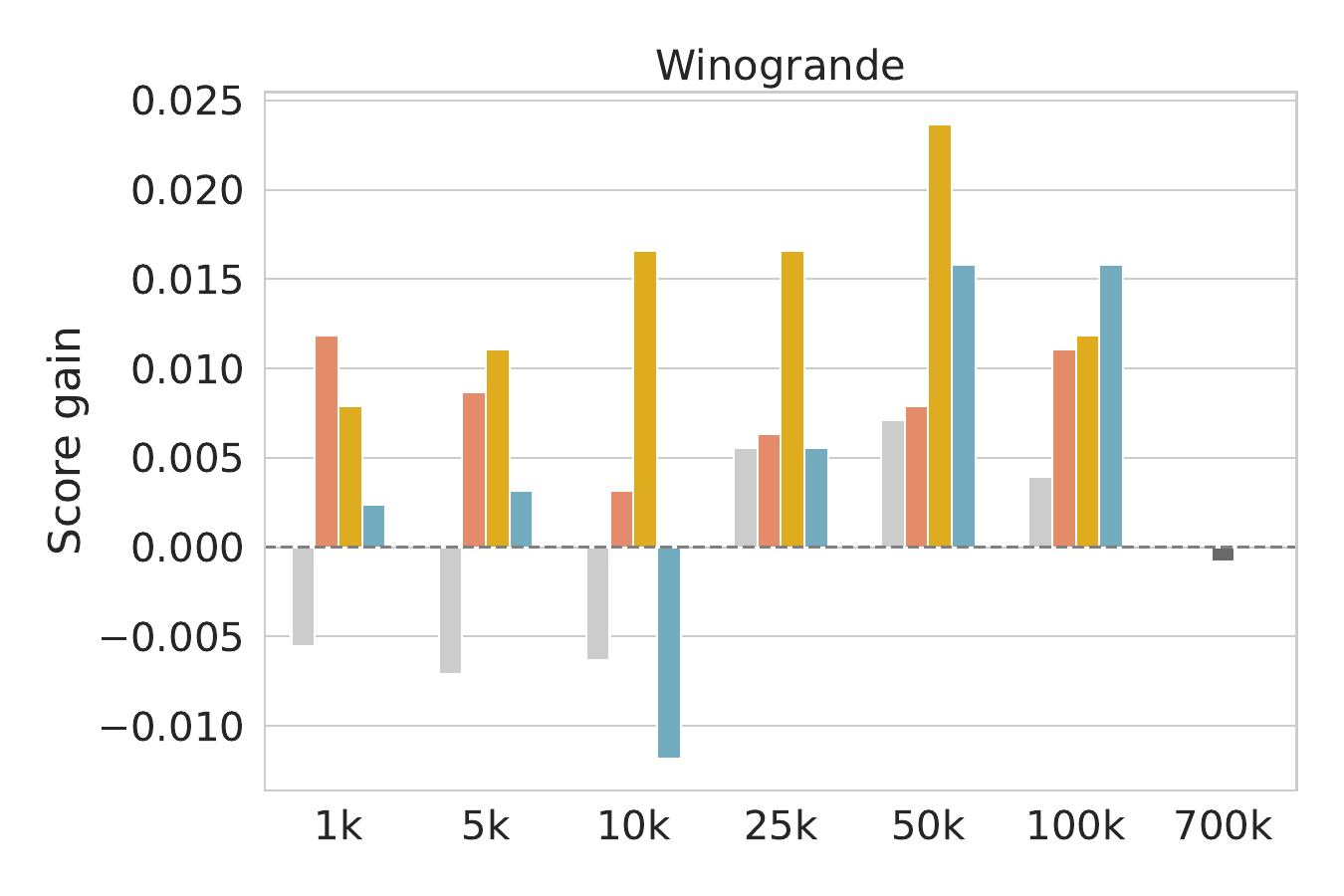}
        \caption{Winogrande \citep{ai2_winogrande}}
        \label{fig:eval_scaling_winogrande}
    \end{subfigure}

    \caption{Results scaling across all benchmark datasets}
    \label{fig:eval_scaling_all_benchmarks}
\end{figure*}

Figure~\ref{fig:eval_scaling_all_benchmarks} shows the scaling experiment results across benchmarks, reported without normalizing \emph{score gain}.

\FloatBarrier

\begin{table*}[!th] 
\centering 
\small
\begin{adjustbox}{width=1.0\textwidth}
\begin{tabular}{@{}llr@{}}
\toprule
\textbf{Category} & \textbf{Training data} (\emph{subset}) & \textbf{\#Samples} \\
\midrule
Math & nvidia/OpenMathInstruct-2, AI-MO/NuminaMath-CoT & 250 \\ \arrayrulecolor{black!30}\midrule
Coding & openbmb/UltraInteract\_sft (\emph{Coding}), microsoft/orca-agentinstruct-1M-v1 (\emph{code}), & 253 \\
 & HuggingFaceH4/no\_robots (\emph{Coding}), lissadesu/codeqa\_v3 \\ \midrule
Generation & HuggingFaceH4/no\_robots (\emph{Generation, Rewrite, Summarize}), HuggingFaceH4/ifeval-like-data,  & 250 \\
 & declare-lab/InstructEvalImpact (\emph{Creative, Professional}), iamketan25/roleplay-instructions-dataset \\ \midrule
Extraction & HuggingFaceH4/no\_robots (\emph{Closed QA, Extract}) & 250 \\ \midrule
Factual QA & HuggingFaceH4/no\_robots (\emph{Open QA}), basicv8vc/SimpleQA & 255 \\ \midrule
Brainstorming & declare-lab/InstructEvalImpact (\emph{Informative, Argumentative}), HuggingFaceH4/no\_robots (\emph{Brainstorming, Classify}), & 250 \\ 
 & matt-seb-ho/WikiWhy \\ \midrule
Reasoning & renma/ProofWriter, hitachi-nlp/ruletaker, lucasmccabe/logiqa, lucasmccabe/logiqa, tasksource/strategy-qa & 250 \\
\arrayrulecolor{black}\bottomrule
\end{tabular}
\end{adjustbox}
\caption{Training data overview for SetFit classifier.}
\label{tab:setfit-training-data}
\end{table*}

\begin{table*}[!t] 
\centering
\small
\begin{tabular}{@{\hskip 2pt}l@{\hskip 6pt}l@{\hskip 6pt}l@{\hskip 6pt}r@{\hskip 6pt}r@{}} 
\toprule
\textbf{Dataset} & \textbf{Subset} & \textbf{Split} & \textbf{$n$ samples} & \textbf{Eval. metric} \\
\midrule
GSM8K \tiny\citep{cobbe2021gsm8k}\normalsize & Default & train & 1192 & exact match \\
Math \tinycitep{hendrycksmath2021} & Algebra & train & 322 & exact match \\
  & Counting \& Probability & train & 264 & exact match \\
  & Geometry & train & 220 & exact match \\
  & Intermediate Algebra & train & 233 & exact match \\
  & Number Theory & train & 314 & exact match \\
  & Prealgebra & train & 348 & exact match \\
  & Precalculus & train & 238 & exact match \\
IFEval-like & Default & - & 1990 & instance level loose acc \\
MBPP \tinycitep{austin2021program}& Default & train \& val & 448 & pass@1 \\
OpenBookQA \tinycitep{mihaylov-etal-2018-suit} & Default & train & 299 & acc \\
ARC \tinycitep{allenai_arc} & Challenge & train & 464 & acc \\
bAbI \tinycitep{dodge2016evaluating} & Default & train & 224 & exact match \\
CommonsenseQA \tinycitep{talmor-etal-2019-commonsenseqa} & Default & train & 248 & acc \\
CoQA \tinycitep{reddy-etal-2019-coqa} & Default & train & 380 & F1 \\
DROP \tinycitep{Dua2019DROP} & Default & train & 383 & F1 \\
FLD \tinycitep{pmlr-v202-morishita23a}& Default & train & 739 & exact match \\
 & Logical Formula Default & train & 750 & exact match \\
HeadQA \tinycitep{vilares-gomez-rodriguez-2019-head} & English & train & 689 & acc \\
 & Spanish & train & 703 & acc \\
JSONSchemaBench \tinycitep{geng2025jsonschemabench} & Easy & train & 200 & schema compliance \& json validity \\
 & Medium & train & 198 & schema compliance \& json validity \\
 & Hard & train & 179 & schema compliance \& json validity \\
LogiQA  \tinycitep{liu2020logiqa} & LogiEval & train & 490 & exact match \\
 & LogiQA2 & train & 416 & acc \\
MLQA \tinycitep{lewis2019mlqa} & 49 lang. combinations & val & 715 & F1 \\
TriviaQA \tinycitep{2017arXivtriviaqa} & Default & train & 1389 & exact match \\
OpenAssistent \tinycitep{kopf2023openassistant} & Default & train & 5318 & LLM-as-a-judge \\
APPS \tinycitep{hendrycksapps2021} & introductory & train & 422 & pass@1 \\
  & interview & train & 303 & pass@1 \\
  & competition & train & 289 & pass@1 \\
CONALA \tinycitep{yin2018learning} & Default & train & 97 & Bleu \\
\bottomrule
\end{tabular}
\caption{Datasets used in difficulty scorer training.}
\label{tab:diff_scorer_training_datasets}
\end{table*}

\FloatBarrier

\begin{figure*}[!t]
\centering
\scriptsize
\ttfamily
\begin{tabular}{@{}l@{}}
\toprule
Given the following categories:\\
- Math (math questions and math reasoning problems)\\
- Coding (programming tasks or coding questions)\\
- Generation (creative generation tasks with constraints, including roleplaying)\\
- Reasoning (logical deductive reasoning tasks that are neither math nor coding)\\
- Brainstorming (information-seeking or recommendation questions requiring explanation, or classification tasks)\\
- Factual QA (simple factual questions, without any context)\\
- Extraction (extraction tasks, including QA, from a given textual passage)\\
\\
What is the category of the following task? Please respond only in JSON format (e.g., {{"answer": "Generation"}})\\
\\
\#\#\# Task \#\#\#\\
\{input\}\\
\bottomrule
\end{tabular}
\caption{Zero-shot prompt for instruction categorization.}
\label{fig:zero-shot-prompt}
\end{figure*}

\begin{figure*}[!h]
\centering
\scriptsize
\ttfamily
\begin{tabular}{@{}l@{}}
\toprule
\#\#\# Task \#\#\#\\
1. Given the following user's PROMPT and system's RESPONSE, please review the code snippet in the RESPONSE.\\
2. Focusing on functional correctness, give the final verdict: 'correct' vs 'incorrect'.\\
3. Extract the original code snippet, write "no code" if there's no code snippet.\\
4. If the original code is correct, simply write "no revision", otherwise propose a code revision to improve the code.\\
5. Provide your answer in JSON format, with \"review", "final\_verdict", "code\_original" and "code\_revision" as keys.\\
\\
\#\#\# User's PROMPT \#\#\#\\
\{instruction\}\\
\\
\#\#\# System's RESPONSE \#\#\#\\
\{output\}\\
\bottomrule
\end{tabular}
\caption{Zero-shot prompt for code review and code revision.}
\label{fig:zero-shot-prompt-code-score}
\end{figure*}

\begin{figure*}[!h]
    \centering
    \scriptsize
    \ttfamily
    \begin{tabular}{@{}l@{}}
    \toprule
    \#\#\# Task \#\#\#\\
    1. Given a USER's prompt, decide whether the constraints from the list below are expressed in the USER's prompt (yes/no).\\
    2. Provide the expressed constraints in JSON format with the expressed constraint as the key\\
       and the constraint type as the value if the respective constraint is expressed.\\
    \\
    \#\#\# List of Constraints \#\#\#\\
    - letter\_case, e.g., lowercase, all capitals\\
    - placeholder\_and\_postscript\\
    - repeat\_prompt, e.g., repeat the request\\
    - output\_combination, e.g., multiple responses, separate the response\\
    - choose\_output, e.g., choose answer from given options\\
    - output\_format, e.g., json format, markdown format, bulleted list, formatted title, highlighted sections\\
    - keyword\_included, e.g., included words\\
    - keyword\_avoided, e.g., avoided words\\
    - keyword\_frequency, e.g., five hashtags, 'but' two times, letter 'r' at least 3 times\\
    - language, e.g., english, two languages\\
    - length, e.g., number of words, number of sentences, number of paragraphs\\
    - punctuation, e.g., no commas, quotation\\
    - start\_and\_ending, e.g., start with 'Hello', end with 'Thank you!'\\
    - writing\_style (e.g., shakespeare, easy-to-read, 5-year-old, persuasive)\\
    - writing\_type (e.g., letter, email, proposal, poem)\\
    - topic (e.g., love)\\
    \\
    \#\#\# Examples \#\#\#\\
    \{few-shot\_examples\}\\
    \\
    \#\#\# Question \#\#\#\\
    USER: \{instruction\}\\
    \bottomrule
    \end{tabular}
    \caption{Few-shot prompt for constraint identification.}
    \label{fig:few-shot-prompt-constraint-identification}
\end{figure*}

\begin{figure*}[ht]
    \parbox{.48\linewidth}{
        \centering
        \scriptsize
        \ttfamily
        \begin{tabular}{@{}l@{}}
        \toprule
        \#\#\# Task \#\#\#\\
        Answer the following questions. Provide the answer in JSON format \\
        with the question number as the key and the answer as the value \\
        (true/false).\\
        Questions:\\
        \{questions\}\\
        \\
        ASSISTANT:\\
        \{output\}\\
        \bottomrule
        \end{tabular}
        \caption{Zero-shot prompt for constraint verification.}
        \label{fig:prompt-constraint-verification}
    }
    \hfill
    \parbox{.48\linewidth}{
        \centering
        \scriptsize
        \ttfamily
        \begin{tabular}{@{}l@{}}
        \toprule
        \#\#\# Task \#\#\#\\
        Given the USER's prompt and ASSISTANT's response below, \\
        analyze whether the response addresses USER's intents properly, \\
        while respecting any constraints expressed in the prompt. \\
        Based on these judgments, provide the final score of response \\
        quality in the range of 1 to 10, in JSON format ('score' as key \\
        and quality score as value).\\
        \\
        USER: {instruction}\\
        ASSISTANT:\\
        \{output\}\\
        \bottomrule
        \end{tabular}
        \captionof{figure}{Zero-shot prompt for response evaluation.}
        \label{fig:prompt-response-evaluation}
        }
\end{figure*}

\end{document}